%% file: main.tex
\documentclass[10pt,journal,compsoc]{IEEEtran}

\ifCLASSOPTIONcompsoc
  \usepackage[nocompress]{cite}
\else
  \usepackage{cite}
\fi

\usepackage{gensymb}
\usepackage{comment}
\usepackage{graphicx}
\usepackage{amsmath,amssymb} 
\usepackage{color}
\usepackage{wrapfig}
\usepackage{makecell}
\usepackage{svg}
\usepackage{rotating}
\usepackage{enumitem}
\usepackage{multirow}
\usepackage{microtype}
\usepackage{xparse}
\usepackage{booktabs,tabu}
\usepackage{floatrow}
\usepackage{nicefrac}
\usepackage{subfigure}
\usepackage{float}
\usepackage{titlesec}

\usepackage[colorlinks]{hyperref}

\newcommand{\change}[1]{#1}
\newcommand{\changeagain}[1]{#1}

\newcommand{\myparagraph}[1]{\noindent\textbf{#1}\quad}

\graphicspath{{figures/}}

\begin{document}

\title{A Perceptual Measure for Deep Single Image Camera and Lens Calibration}

\author{
Yannick Hold-Geoffroy\textsuperscript{1},
Dominique Pich\'{e}-Meunier\textsuperscript{4}, 
Kalyan Sunkavalli\textsuperscript{1}, \\ 
Jean-Charles Bazin\textsuperscript{2}, 
Fran\c{c}ois Rameau\textsuperscript{3}, 
and Jean-Fran\c{c}ois Lalonde\textsuperscript{4} \\
Adobe Research\textsuperscript{1}, KAIST\textsuperscript{2}, SUNY Korea\textsuperscript{3}, Universit\'e Laval\textsuperscript{4}}

\IEEEtitleabstractindextext{%
\begin{abstract}
Image editing and compositing have become ubiquitous in entertainment, from digital art to AR and VR experiences. To produce beautiful composites, the camera needs to be geometrically calibrated, which can be tedious and requires a physical calibration target. In place of the traditional multi-image calibration process, we propose to infer the camera calibration parameters such as pitch, roll, field of view, and lens distortion directly from a single image using a deep convolutional neural network. We train this network using automatically generated samples from a large-scale panorama dataset, yielding competitive accuracy in terms of standard $\ell^2$ error. However, we argue that minimizing such standard error metrics might not be optimal for many applications. In this work, we investigate human sensitivity to inaccuracies in geometric camera calibration. To this end, we conduct a large-scale human perception study where we ask participants to judge the realism of 3D objects composited with correct and biased camera calibration parameters. Based on this study, we develop a new perceptual measure for camera calibration and demonstrate that our deep calibration network outperforms previous \change{single-image based calibration} methods both on standard metrics as well as on this novel perceptual measure. Finally, we demonstrate the use of our calibration network for several applications, including virtual object insertion, image retrieval, and compositing. 
\end{abstract}

}

\maketitle
\IEEEpeerreviewmaketitle

\input{sec_intro}

\input{sec_relatedwork}

\input{sec_image-formation}

\input{sec_proposed}

\input{sec_results}

\input{sec_applications}

\input{sec_human_perception}

\input{sec_conclusion}
\textbf{Acknowledgements} This project was partially supported by NSERC, Korea NRF, a NSERC USRA to D. Piché-Meunier and a donation from Adobe to J-C Bazin and J-F Lalonde. 



\input{sec_bios}

\bibliographystyle{IEEEtran}
\bibliography{bibliography}

\end{document}

%% file: sec_intro.tex
\IEEEraisesectionheading{\section{Introduction}\label{sec:introduction}}

Most image understanding and editing tasks---ranging from 3D scene reconstruction to image metrology to photographic editing---require some degree of camera calibration to be performed. For example, performing virtual object insertion involves constructing a virtual scene and camera that mimic the background image. To do so, estimating the scene local geometry in addition to the camera parameters---such as focal length, orientation and lens distortion---is of paramount importance. 

Many methods have been proposed in the literature to estimate the camera parameters from images. \change{In particular, methods focusing on metrology} aim to perform absolute metric estimations that typically achieve subpixel reprojection accuracy~\cite{Zhang:TPAMI:00,heikkila2000geometric,Scaramuzza:IROS:06}. However, the usefulness of these approaches comes at a high cost, as they often require multiple captures of specific calibration objects~\cite{Zhang:TPAMI:00}. Some methods relax these requirements to a single capture~\cite{boby2016single}, but still need known objects such as markers to be present in the image. These restrictions make those methods typically tedious, and impractical to perform on images obtained ``in the wild'' where access to the camera is impossible.

One key observation is that a high degree of accuracy in camera calibration may not be necessary for all applications. Consider, for example, image editing such as object copy-paste~\cite{lalonde-siggraph-07} or virtual object insertion~\cite{debevec-siggraph-98}. It is well-known that humans have a very low sensitivity to deviations from realism in some cases, for example in paintings~\cite{Cavanagh2005} and digital composites~\cite{Farid2010}. Is subpixel-precise camera calibration needed in these cases, or do we need to be accurate only up to human perception? 
This raises the important question \emph{how do humans perceive inaccuracies in geometric camera calibration?} In this work, we investigate this topic by conducting a large-scale user study on the human perception of errors in camera calibration. We asked participants to select the most plausible virtual object insertion between pairs of images, using either ground-truth calibration or perturbed parameters to perform the insertions. Using the answers from this user study, we build a human perception measure that estimates how sensitive humans are to specific combinations of geometric camera calibration errors. 


We also develop a single image calibration approach that generalizes to a large diversity of both natural and urban scenes, and which spans a large set of cameras. We parameterize the camera by its field of view, lens distortion, and horizon position in the image, from which can be derived pitch and roll angles. We evaluate our method on a challenging test set and demonstrate that it quantitatively outperforms the state of the art in terms of raw parameter error and are more pleasing to the human eye according to our novel human perception measure when used for applications such as virtual insertion. 

This work unifies and significantly extends on contributions introduced in \cite{Bogdan:CVMP:2018} \change{(field of view, distortion estimation)} and \cite{hold2017perceptual} \change{(field of view, pitch, roll estimation)}. First, we \change{perform two large-scale perceptual experiments which evaluate the sensitivity of human observers to} inaccuracies in geometric camera calibration, in terms of field of view, roll and pitch angles, as well as lens distortion. Second, from \change{these experiments} is derived a novel quantitative measure of \change{observer} sensitivity to geometric camera calibration errors. Third, we present a method for automatically estimating both intrinsic and extrinsic parameters of a camera from a single image, which works on a wide range of parameters including very high lens distortion. Finally, we present an extensive set of experiments to benchmark our method, both in terms of absolute error and human perception.

%% file: sec_relatedwork.tex
\section{Related work}
\label{sec:related_work}

\myparagraph{Geometric camera calibration} is a widely studied topic that has a significant impact on a variety of applications including metrology~\cite{Criminisi2000}, 3D inference~\cite{Criminisi00,Fouhey2013} and augmented reality, both indoor~\cite{hedau-iccv-09,izadinia-cvpr-17} and outdoor~\cite{hoiem-cvpr-06}. As such, many techniques have been developed to perform precise geometric calibration using a calibration target inserted in the image~\cite{Hartley2004,Sturm1999,Zhang:TPAMI:00,remondino2006digital,Heikkila1997,Chen2004,Scaramuzza:IROS:06,mei2007single,Gasparini:ICCV:09,Shah:ICRA:94,Ying:IJCV:08,larsson_revisiting_2019}. Despite their great accuracy, these methods require a cumbersome acquisition process, precluding them from being applied on images in the wild. To alleviate this limitation, work has been done to directly use the features present in urban scenes to perform calibration, typically using either lines~\cite{zhang2015line,Workman2016,Barreto:TPAMI:05,Lee2014}, vanishing points~\cite{hughes2010equidistant,Antunes:CVPR:2017}, higher-level geometric cues~\cite{xian2019uprightnet}, or specific objects typically present in human-made environments~\cite{Rother:BMVC:2000,Melo:ICCV:13}. 
While greatly simplifying the calibration process, these methods are limited to structured man-made scenes, and thus cannot deal with general environments such as landscapes or natural scenes. 

Other work has proposed to take advantage of lighting cues for camera calibration~\cite{Lalonde2010,Workman2014} from the sky and from rainbows, circumventing the need to detect vanishing lines. However, these techniques often fail on complex scenes where semantic reasoning is required to discard misleading textures and visual cues. To solve the need for high-level reasoning, convolutional neural networks (CNNs) were used to bring camera calibration on single images to a wider variety of scenes. One of such first attempts is the approach of Mendon{\c{c}}a et al.~\cite{mendoncca2002camera}, which uses neural network to compute the camera parameters given 3D point locations from a calibration target and their respective 2D observations. 
More recently, deep learning-based methods were proposed to estimate individual camera parameters from a single image without calibration target. For example, DeepFocal~\cite{workman2015deepfocal} predicts the camera's focal length, DeepHorizon ~\cite{Workman2016} estimates the camera's pose using the horizon line, and
Rong et al.~\cite{rong2016radial} estimate the radial distortion. Similar techniques were explored for 360 panorama upright alignment~\cite{Jung:VR:upright:2019,jung2017robust}. 

Hold-Geoffroy et al.~\cite{hold2017perceptual} estimate the camera's focal length and orientation. They train a CNN on images generated from panoramas using standard pinhole model, and thus, are limited to perspective cameras. Latest work on geometric calibration estimation propose optimization-based methods that simultaneously solves for lens distortion and camera pose~\cite{Lochman_minimal_2021_WACV}, combining circular vanishing lines estimation~\cite{Antunes:CVPR:2017,wildenauer2013closed} with the detection of repeated patterns~\cite{pritts2018radially}. Closer to our work,~\cite{Lopez:CVPR:2019} presents an end-to-end learned approach to estimate camera parameters of large field-of-view images trained on pixel ray direction error. In our work, we draw inspiration from~\cite{Bogdan:CVMP:2018} and use a unified spherical model~\cite{mei2007single,barreto2006unifying} to represent the lens distortion.

Our method holistically estimates the camera parameters, providing both camera intrinsics and its pose at the same time. Its output is readily usable for image undistortion and 3D reconstruction and supports a large diversity of lens types, from perspective (pinhole) to fisheye.

\myparagraph{Image undistortion} A common task when using wide FOV cameras is image undistortion---also called rectification---which undoes the warping caused by the lens, effectively straightening the lines in an image. To do this, we need to calibrate the camera's lens distortion. For this purpose, various distortion models have been developed. One of the most popular representation is the Brown-Conrady model~\cite{duane1971close} which models the radial and tangential lens distortion via a polynomial function. While this model approximates reasonably low distortions, it is not suitable for thicker lenses with larger fields of view which exhibit large geometric distortions (e.g. fisheye cameras)~\cite{sturm2011camera}. To cope with this limitation, appropriate distortion models for large field-of-view cameras have been developed. A representative example is the \emph{division model} proposed by Fitzgibbon~\cite{fitzgibbon2001simultaneous}, which also represents non-linear distortion via a radially-symmetric polynomial equation. These models have proven effective for self-calibration~\cite{fitzgibbon2001simultaneous} and robotics navigation with wide FoV cameras~\cite{pizarro2003toward,lin2020infrastructure}. Unfortunately, they require multiple unbounded parameters and they are difficult to invert, which make them both less suited for training a neural network and generating training data, respectively. 
%
Once geometrically calibrated, many image processing libraries such as OpenCV~\cite{opencv_library}, ImageMagick~\cite{imagemagick}, Adobe Photoshop, PTLens~\cite{ptlens} and Hugin~\cite{hugin} propose algorithms to project captured images to a flat surface.


%

For large field of view capture setups, the Fisheye-Hemi plug-in for Photoshop~\cite{FisheyeHemi}, Carroll et al.~\cite{CarrollAA09} and the FOVO projection~\cite{pepperell2019fovo} aim to produce aesthetically pleasing visualizations, used notably in architecture model display.

\myparagraph{Perception}
Understanding the limits of the human visual system has also received significant attention, with studies quantifying color sensitivity~\cite{fairchild2013color}, how reliably we can detect photo manipulations artifacts~\cite{Farid2010} and how people perceive distortion in street-level image-based rendering~\cite{Vangorp2013}. More recently, perceptual studies were performed to assess human appreciation on tasks like super-resolution~\cite{ledig-cvpr-17}, image caption generation~\cite{vinyals-cvpr-15} and video temporal alignment~\cite{papazoglou-accv-16}.

In this work, we go one step further by studying human sensitivity to camera calibration errors (pitch, roll, field of view and distortion) by using virtual object insertion as a test scenario.

%% file: sec_image-formation.tex
\begin{figure}[tb]
\floatbox[{\capbeside\thisfloatsetup{capbesideposition={left,top},capbesidewidth=4cm}}]{figure}[\FBwidth]
{\caption{The unified spherical model~\cite{barreto2006unifying,mei2007single}. A 3D point $\mathbf{p}_\mathrm{c} = \mathbf{R}\mathbf{p}_\mathrm{w}$ (see sec.~\ref{sec:extrinsics}) is first projected onto the unit sphere with $\mathbf{p}_\mathrm{s} = \mathbf{p}_\mathrm{c}/\|\mathbf{p}_\mathrm{c}\|$, and then onto the image plane $\mathbf{p}_\mathrm{im}$ using a line starting from a point $\mathbf{O}_\mathrm{c}$ located at $\left[ 0,0,\xi \right]$ below the sphere center $\mathbf{O}$. Distances not to scale to simplify visualization. 
\label{fig:unified_spherical_model}}}{
\includegraphics[width=0.85\linewidth]{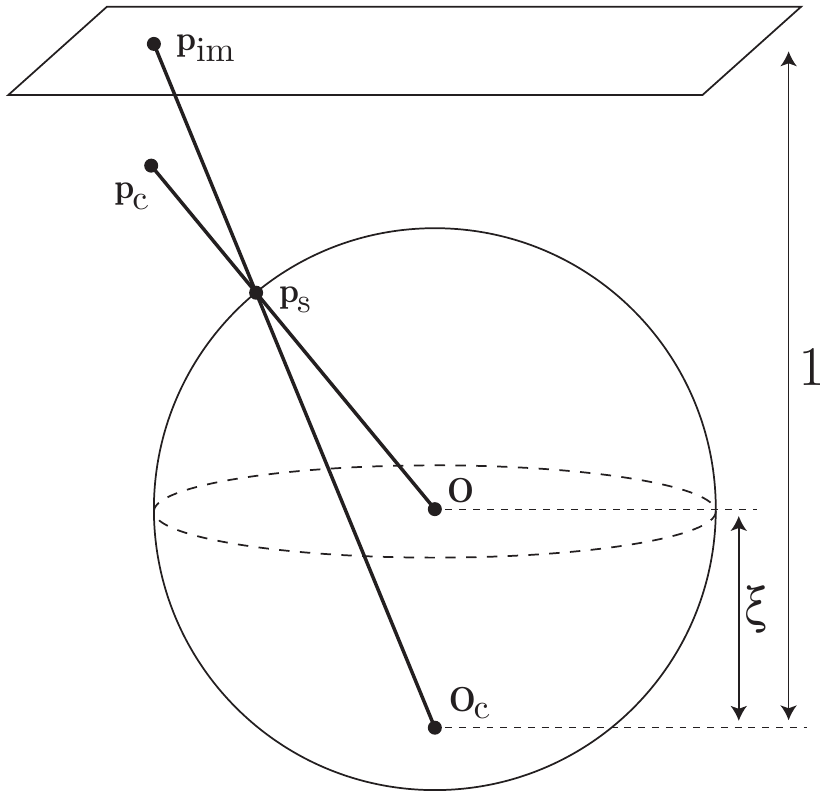}
}
\end{figure}

\addtolength{\tabcolsep}{-3pt}
\begin{figure*}
    \centering
    \subfigure[]{
        \begin{tabular}{cc}
            \includegraphics[width=0.23\linewidth]{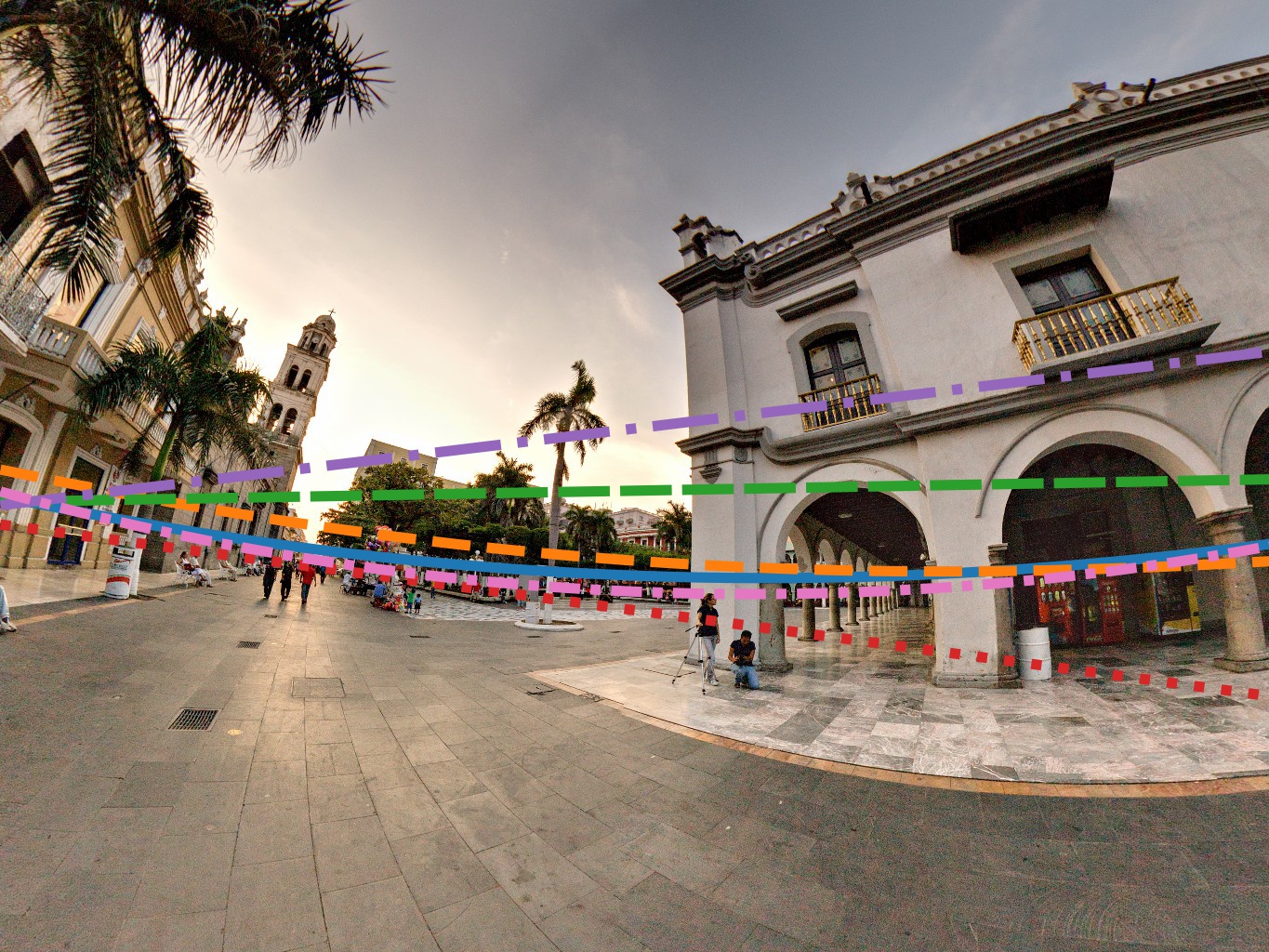} &  \includegraphics[width=0.23\linewidth]{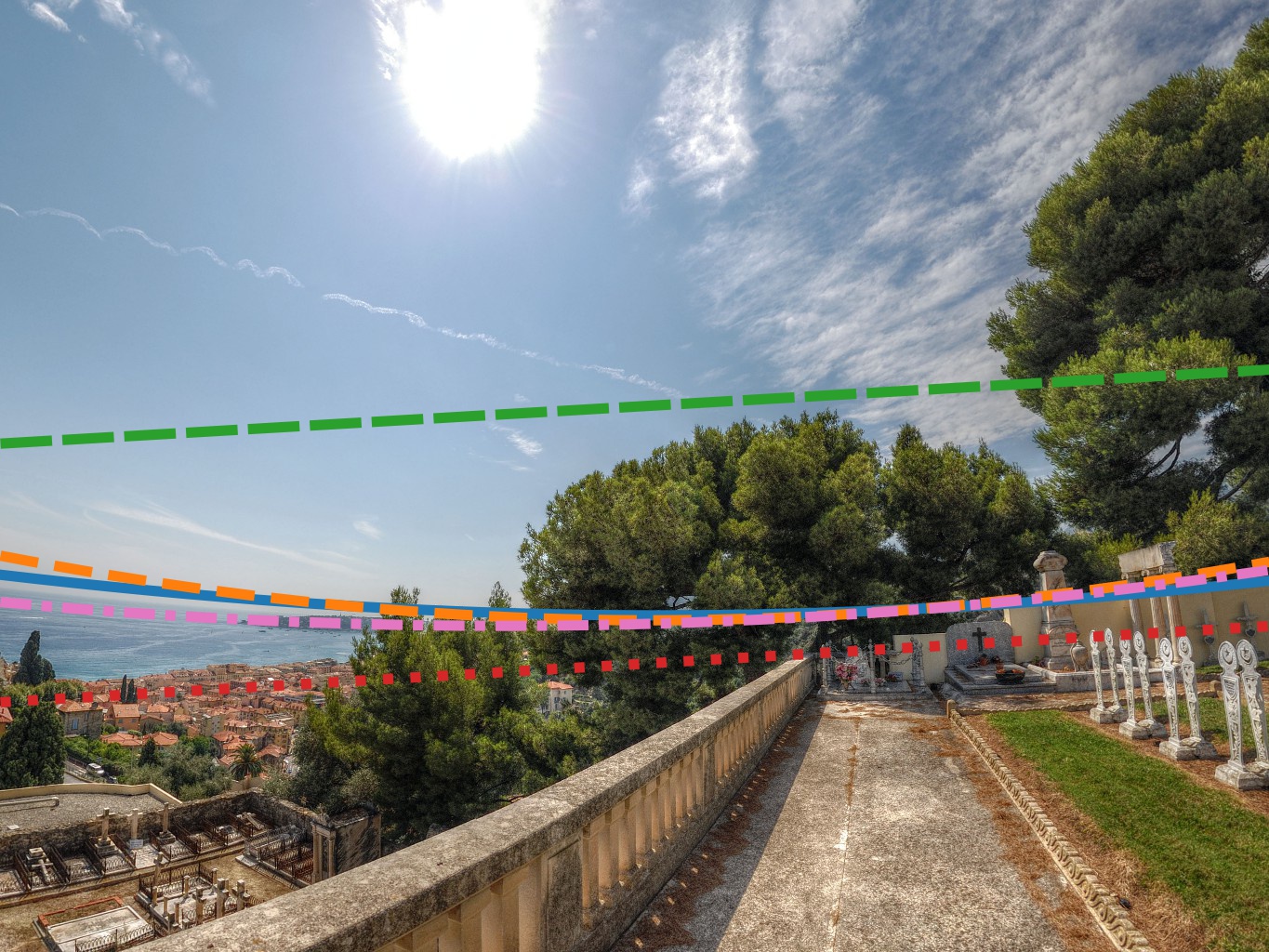} \\
            \includegraphics[width=0.23\linewidth]{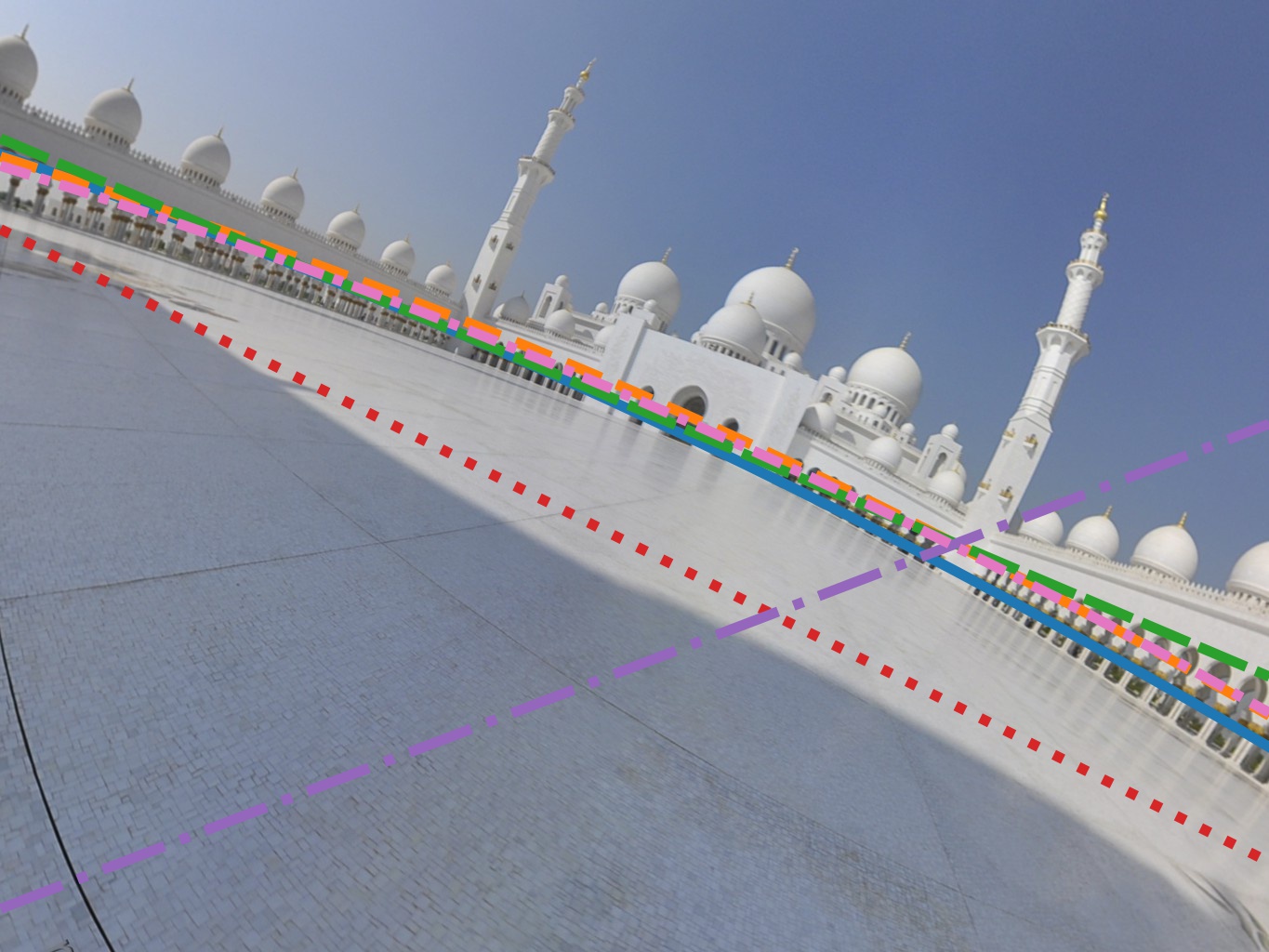} &
            \includegraphics[width=0.23\linewidth]{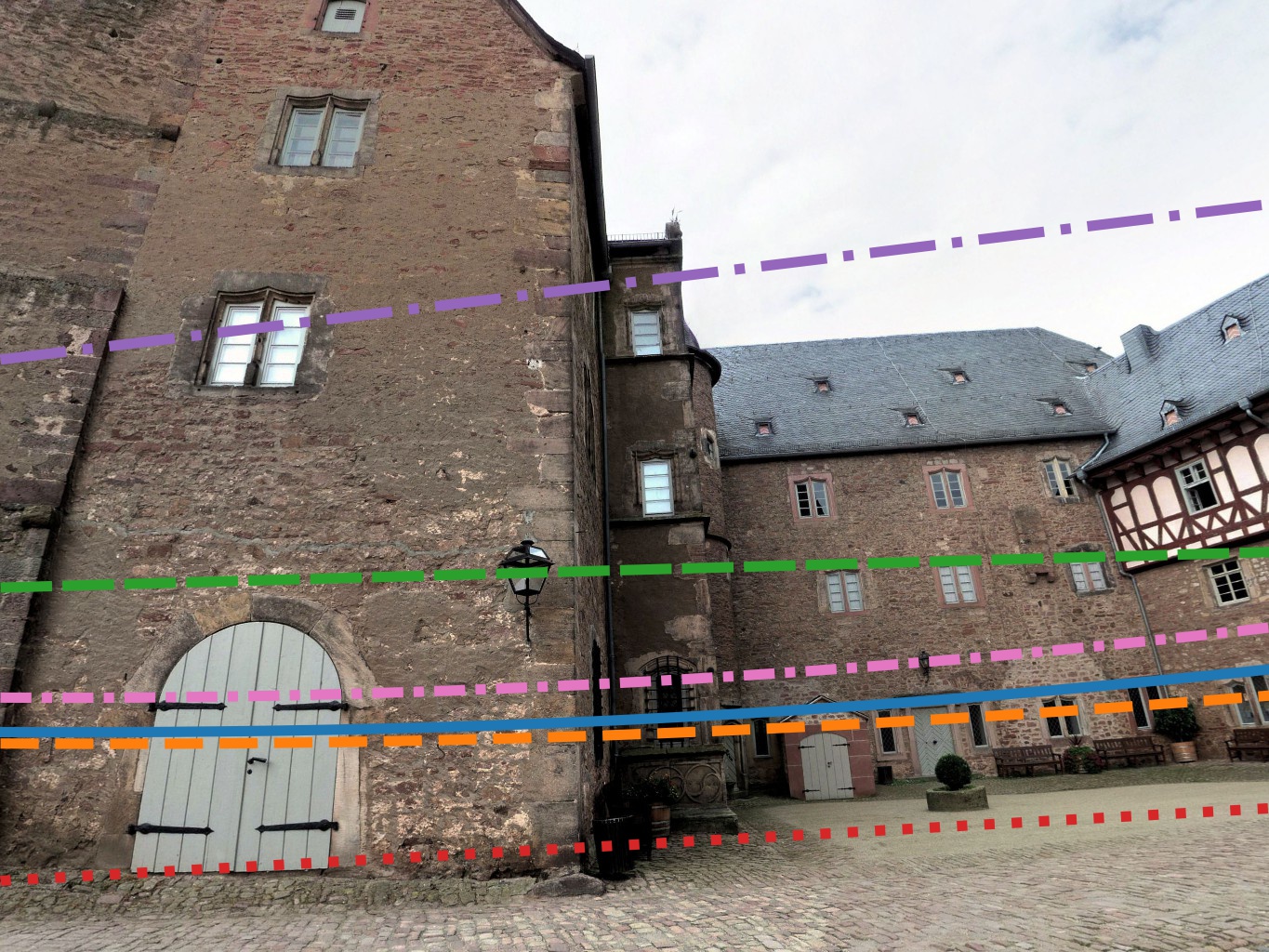} \\ 
            \includegraphics[width=0.23\linewidth]{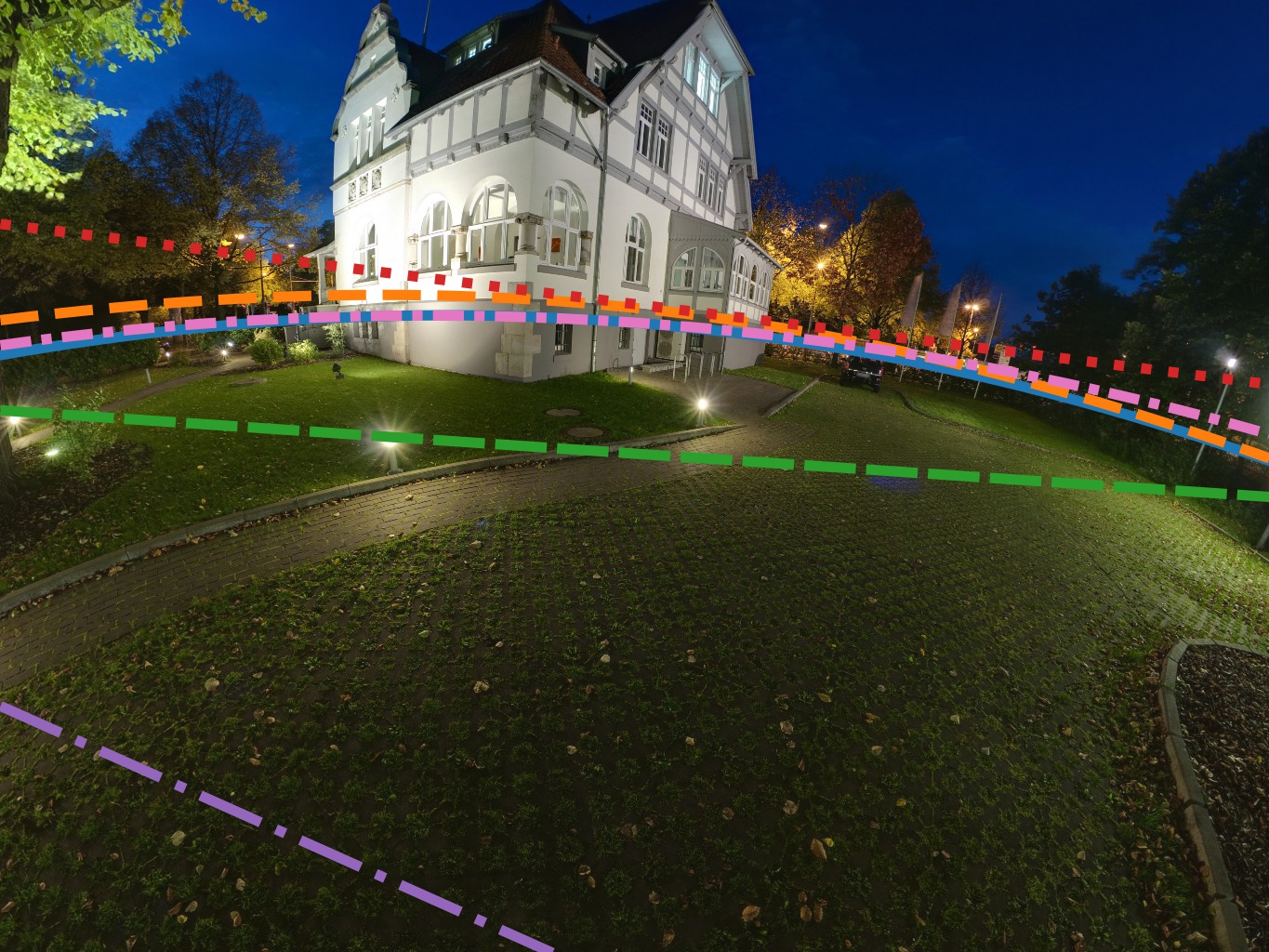} &
            \includegraphics[width=0.23\linewidth]{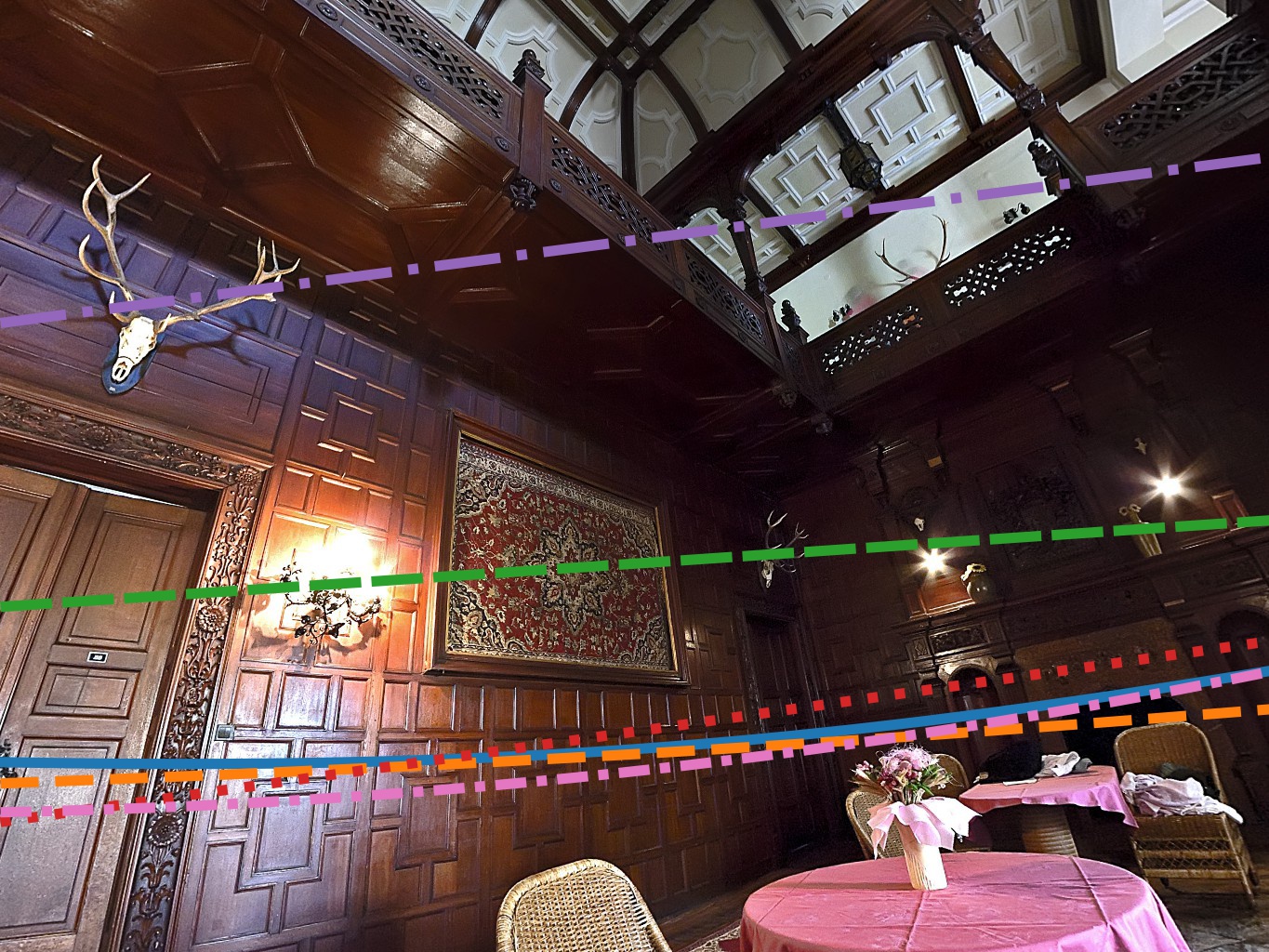} \\
        \end{tabular}
    }
    \subfigure[]{
        \begin{tabular}{cc}
            \includegraphics[width=0.23\linewidth]{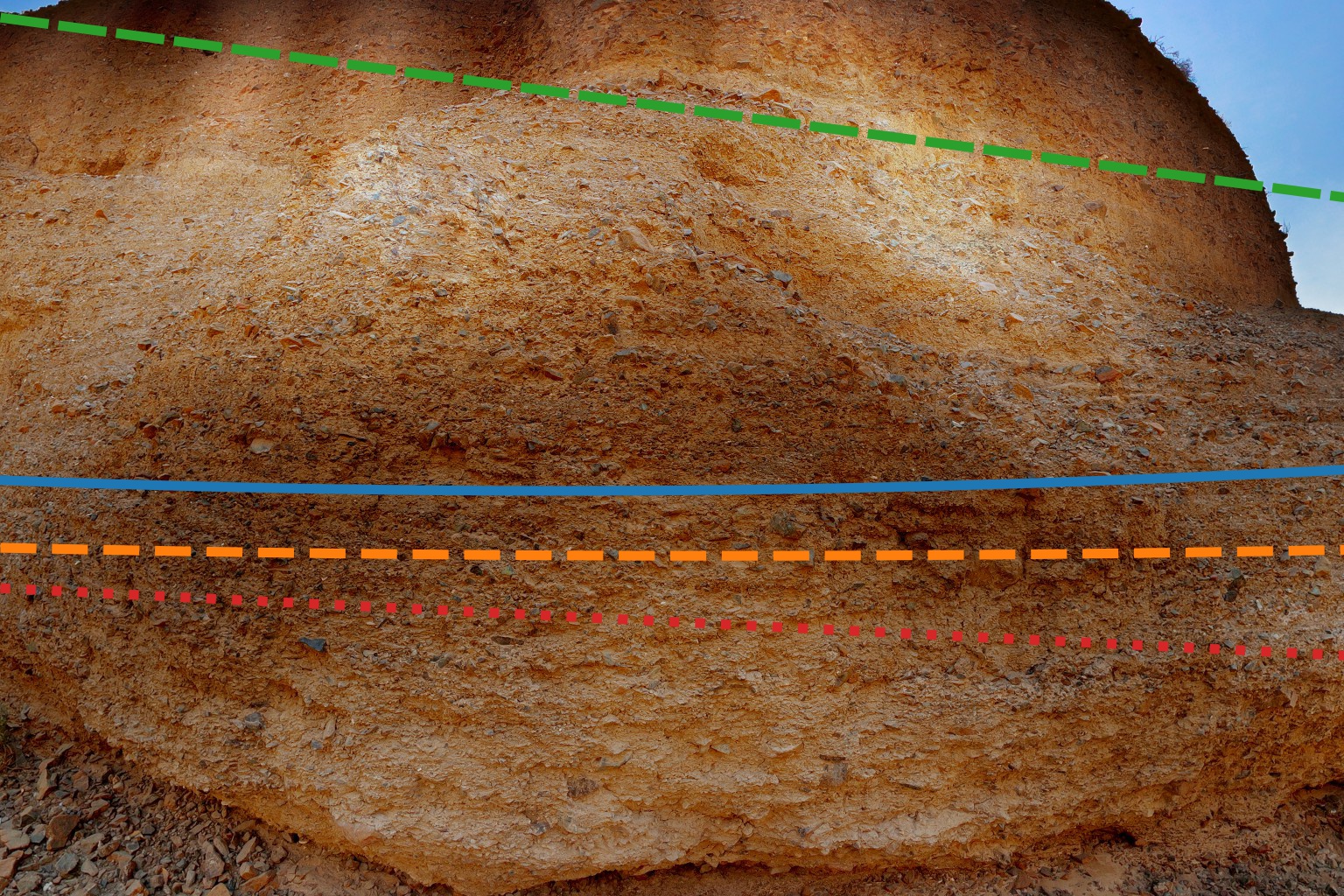} &  \includegraphics[width=0.23\linewidth]{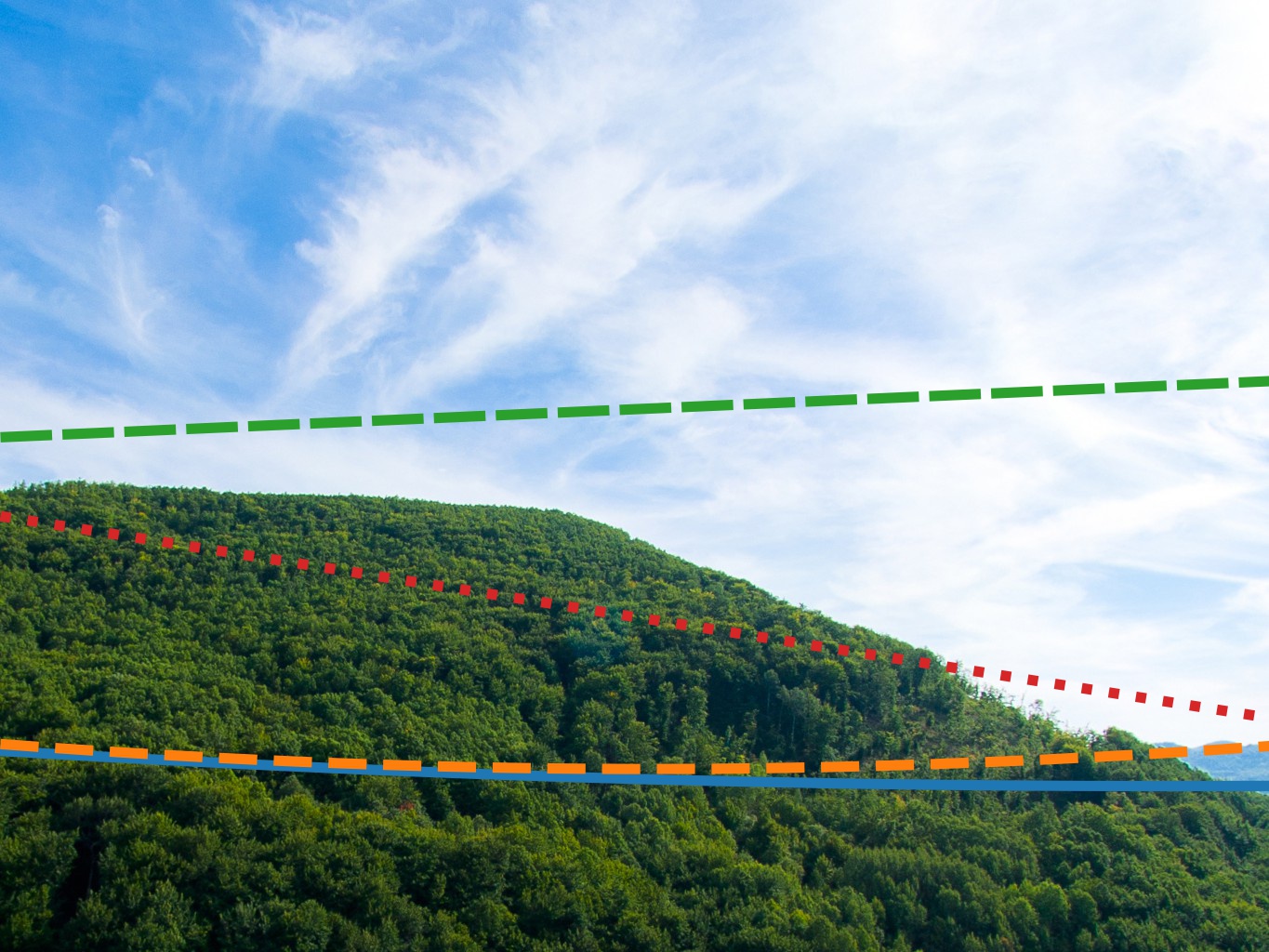} \\
            \includegraphics[width=0.23\linewidth]{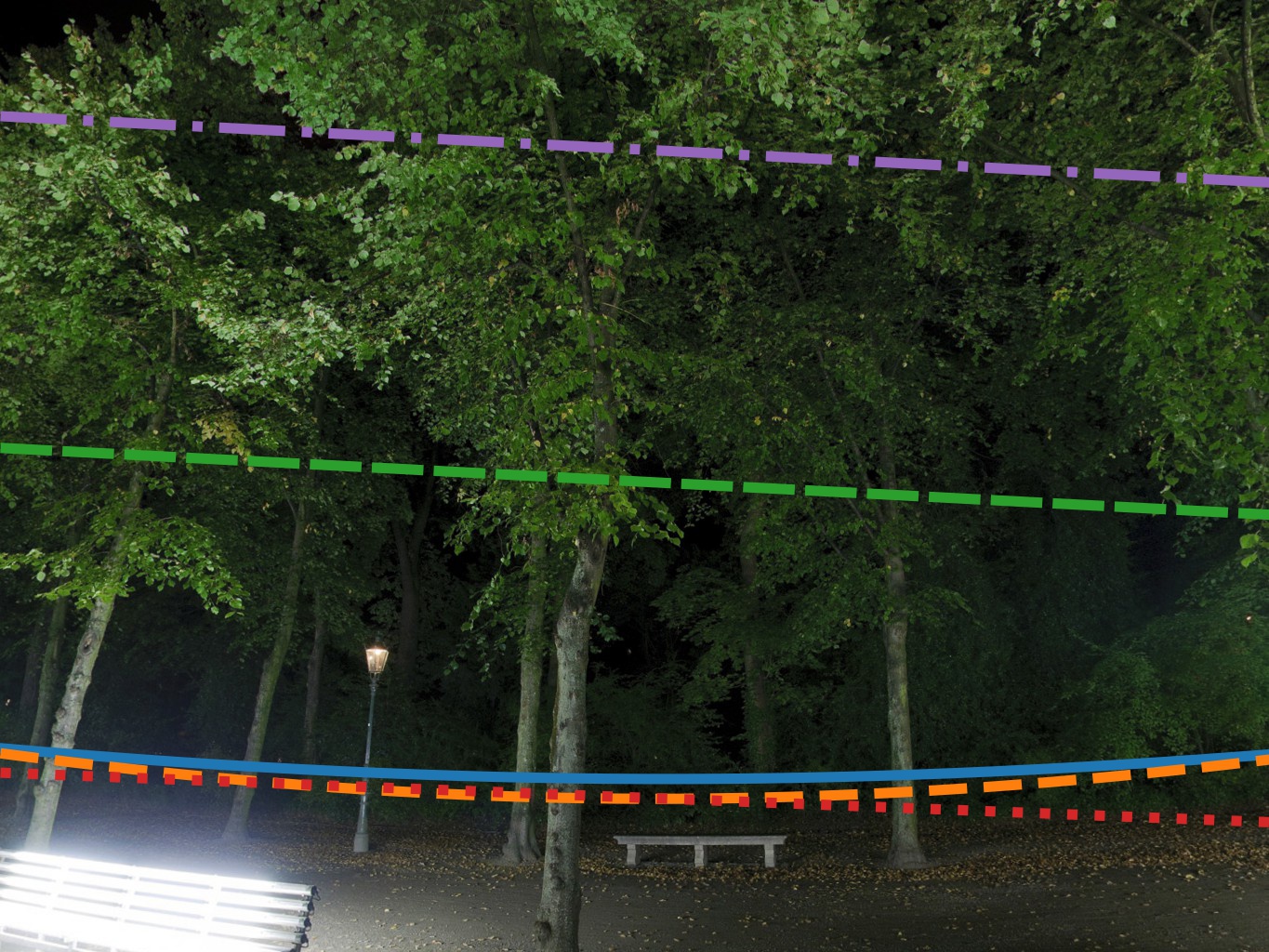} &
            \includegraphics[width=0.23\linewidth]{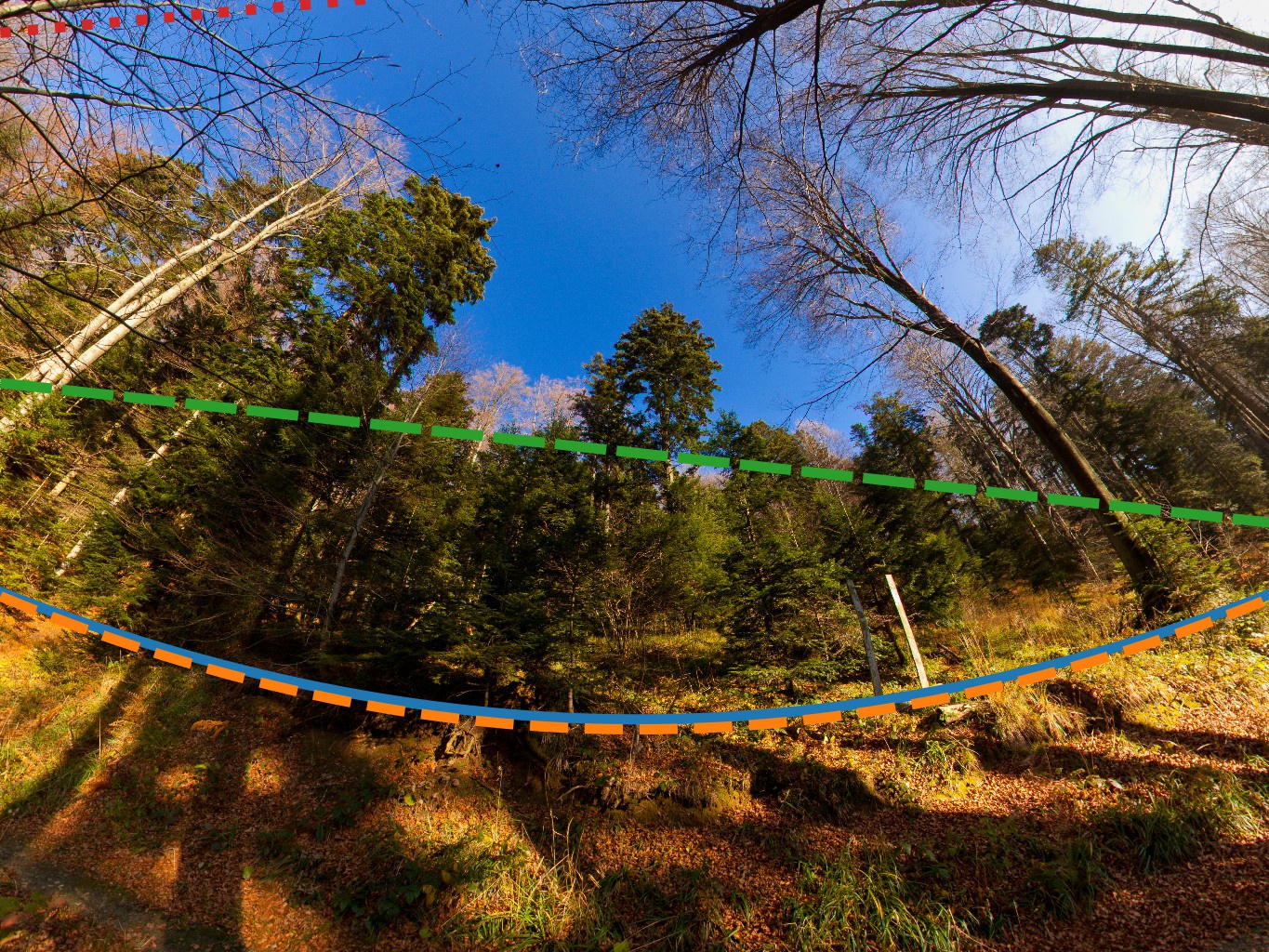} \\
            \includegraphics[width=0.23\linewidth]{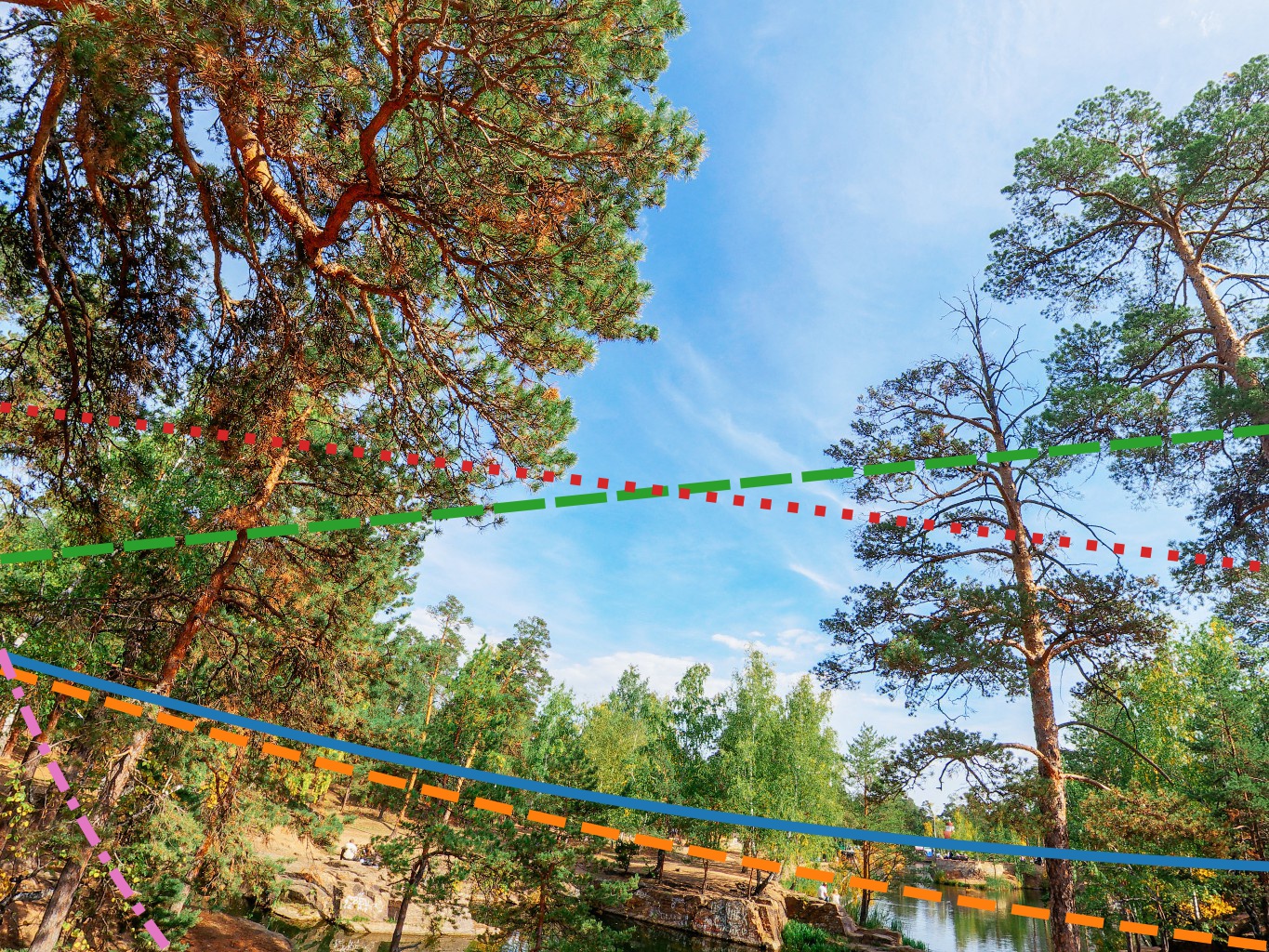} &  \includegraphics[width=0.23\linewidth]{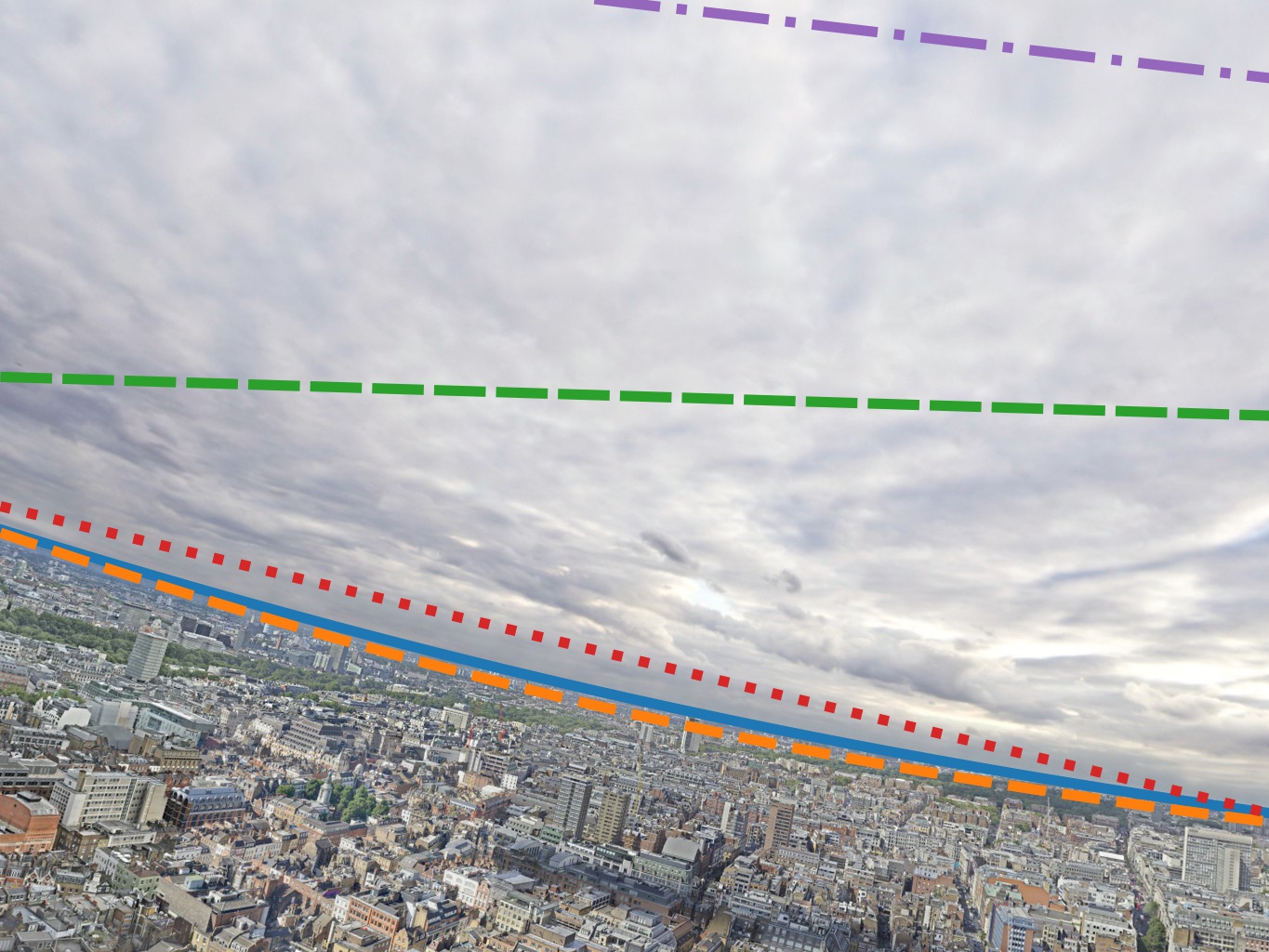} \\
        \end{tabular}
    }
    \subfigure{
    \centering
    \includegraphics[width=0.9\linewidth]{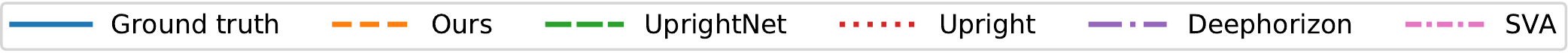}
    }
    \caption{Example results of horizon line estimation on our 360Cities test set, for images both (a) with and (b) without clear vanishing lines. We provide the ground truth field of view to UprightNet~\cite{xian2019uprightnet}. Only our method and SVA~\cite{Lochman_minimal_2021_WACV} allow for curved horizon lines (due to large distortions). Note how Upright~\cite{Jung:VR:upright:2019} and SVA perform well when sharp human-made objects are present in the scene, whereas deep learning methods offer a more robust performance across all scenes. Note also that SVA fails on 49\% of our test images. \textbf{More examples available in the supplementary material.}}
    \label{fig:results_qualitative}
\end{figure*}

\section{Image formation model}
\label{sec:ifm}


\subsection{Projection model}
\label{sec:projection}

We begin by reviewing the main parameters of the image formation model. 
The pixel coordinates $\mathbf{p}_\mathrm{im}$ of a 3D point $\mathbf{p}_\mathrm{w}$ in world coordinates are given by
\begin{equation}
\mathbf{p}_{\mathrm{im}} = [u, \; v]^\mathsf{T} = 
\mathcal{P}(\mathbf{R} \mathbf{p}_{\mathrm{w}} + \mathbf{t} ) \,,
\label{eq:formation}
\end{equation}
where $\mathcal{P}$ is a 3D-to-2D projection operator, including conversion from homogeneous coordinates to 2D. Here, $\mathbf{R}$ and $\mathbf{t}$ are the camera rotation and translation respectively in the world reference frame. 
Under the pinhole camera model and with the common assumptions that the principal point is at the image center, negligible skew and a unit pixel aspect ratio~\cite{Hartley2004}, the projection operator $\mathcal{P}$ would take the form of a matrix $\mathbf{K} = \mathrm{diag}([f \; f \; 1])$, where $f$ is the focal length in pixels and the $\mathrm{diag}$ operator builds a matrix with the specified elements on its diagonal and 0 elsewhere. 

\subsection{Intrinsic parameters}
\label{sec:intrinsics}

Instead of following these typical assumptions, we model $\mathcal{P}$ by including lens distortion and employ the unified spherical model~\cite{barreto2006unifying,mei2007single} since it features several advantages over the models discussed in sec.~\ref{sec:related_work}. First, the radial distortion can be represented by a \emph{single}, \emph{bounded} parameter $\xi \in [0, 1]$\footnote{$\xi$ can be slightly greater than 1 for certain types of catadioptric cameras~\cite{YingH:ECCV:04} but we ignore this scenario in this work.}. Second, this model is fully invertible which make it practical to both add/remove distortion to/from an image. Third, both the projection and back-projection operations admit closed-form solutions that can be computed very efficiently. Finally, unlike polynomial-based models, the spherical model is compatible with the entire family of single view points sensors, including perspective, fisheye and catadioptric cameras. 

The unified spherical model relies on a stereographic projection (fig.~\ref{fig:unified_spherical_model}), which makes the basis for our projection operator $\mathcal{P}$ from eq.~\ref{eq:formation}.
First, the 3D point $\mathbf{p}_\mathrm{c} = \mathbf{R}\mathbf{p}_\mathrm{w}$ (in camera coordinates, see sec.~\ref{sec:extrinsics}) is projected onto the unit sphere with $\mathbf{p}_\mathrm{s} = \mathbf{p}_\mathrm{c}/\|\mathbf{p}_\mathrm{c}\|$. Then, we project the spherical point $\mathbf{p}_\mathrm{s}$ onto the image plane, using $\mathbf{O}_\mathrm{c} = [0,0,\xi]$ as the center of projection. The distance~$\xi$ models the radial distortion of the camera. The projection operator $\mathcal{P}$, which projects a 3D point in camera coordinates $\mathbf{p}_\mathrm{c} = [x, y, z]^\mathsf{T}$ to pixel coordinates $\mathbf{p}_\mathrm{im} = [u, v]^\mathsf{T}$ can therefore be expressed as
\begin{equation}
\left[ u, v \right] = \mathcal{P}(\mathbf{p}_\mathrm{c}) \equiv \left[ \frac{xf}{\xi\alpha+z}+u_0,\frac{yf}{\xi\alpha+z}+v_0 \right]\,,
\label{SphProj1}
\end{equation}
where $\alpha = ||\mathbf{p}_\mathrm{c}||$ and $\left[ u_0 \: v_0 \right]$ is the principal point, assumed in this work to be the image center. 

As mentioned, one of the advantages of the spherical model is the closed-form solution of the inverse projection equation. This inverse projection is especially useful when applying the spherical lens model to panoramic images (see sec.~\ref{sec:dataset_generation}). Given a 2D image point $\mathbf{p}_\mathrm{im}=[u, v]^\mathsf{T}$, the back-projection from the image to the sphere is computed by $\mathbf{p}_\mathrm{s} = [\omega u, \; \omega v, \; \omega \! - \! \xi]^\mathsf{T}$, where
\begin{equation}
\omega = \frac{\xi + \sqrt{1 + (1-\xi^2)(u^2+v^2)}}{u^2+v^2+1} \,.
  \label{Eq::backPro}
\end{equation}
We can also derive the equation for the effective horizontal field of view $h_\theta$, noting $\hat{u} = -u_0/f$: 
\begin{equation}
h_{\theta} = 2\arccos{\left(\frac{\xi+\sqrt{1+(1-\xi^2)\hat{u}^2}}{\hat{u}^2+1} - \xi\right)} \,.
\end{equation}
Please refer to the supplementary material for more details. 

%
%

\subsection{Extrinsic parameters}
\label{sec:extrinsics}

The rotation matrix $\mathbf{R}$ can be parameterized by roll $\psi$, pitch $\theta$, and yaw $\varphi$ angles. Since there exists no natural reference frame to estimate $\varphi$ (left vs right) from an arbitrary image, we therefore constrain the rotation to only pitch and roll components, simplifying the extrinsic rotation matrix to $\mathbf{R} = \mathbf{R}_z(\psi) \mathbf{R}_x(\theta)$. For the same reason, we take the origin to be at the camera with $\mathbf{t} = \mathbf{0}$.

The camera extrinsic parameters are expressed by using the horizon line as an intuitive representation for these angles. As in \cite{Workman2016}, we parameterize the horizon line by its midpoint $v_\mathrm{m}$ and its angle $\psi$. The latter coincides with the camera roll, which we define as the angle between the horizon line in the image and an virtual horizontal line located at the midpoint. We define the midpoint $v_\mathrm{m}$ as the $v$-coordinate of its intersection with the vertical axis in the center of the image. The midpoint $v_\mathrm{m}$ can be derived from $\theta$ and $f$ as
\begin{align*}
    v_\mathrm{m} = \frac{2f\sin\theta}{h \left( \xi+\cos\theta \right)} \,,
    \label{eq:horizon_midpoint}
\end{align*}
where $h$ is the image height. In this normalized image units representation, the top and bottom of the image have coordinates 1 and $-1$, respectively.

%% file: sec_proposed.tex
\section{Proposed approach}
\label{sec:proposed}

\addtolength{\tabcolsep}{-5pt}
\begin{figure*}
    \centering
\begin{tabular}{cccc}
    \includegraphics[width=.247\textwidth]{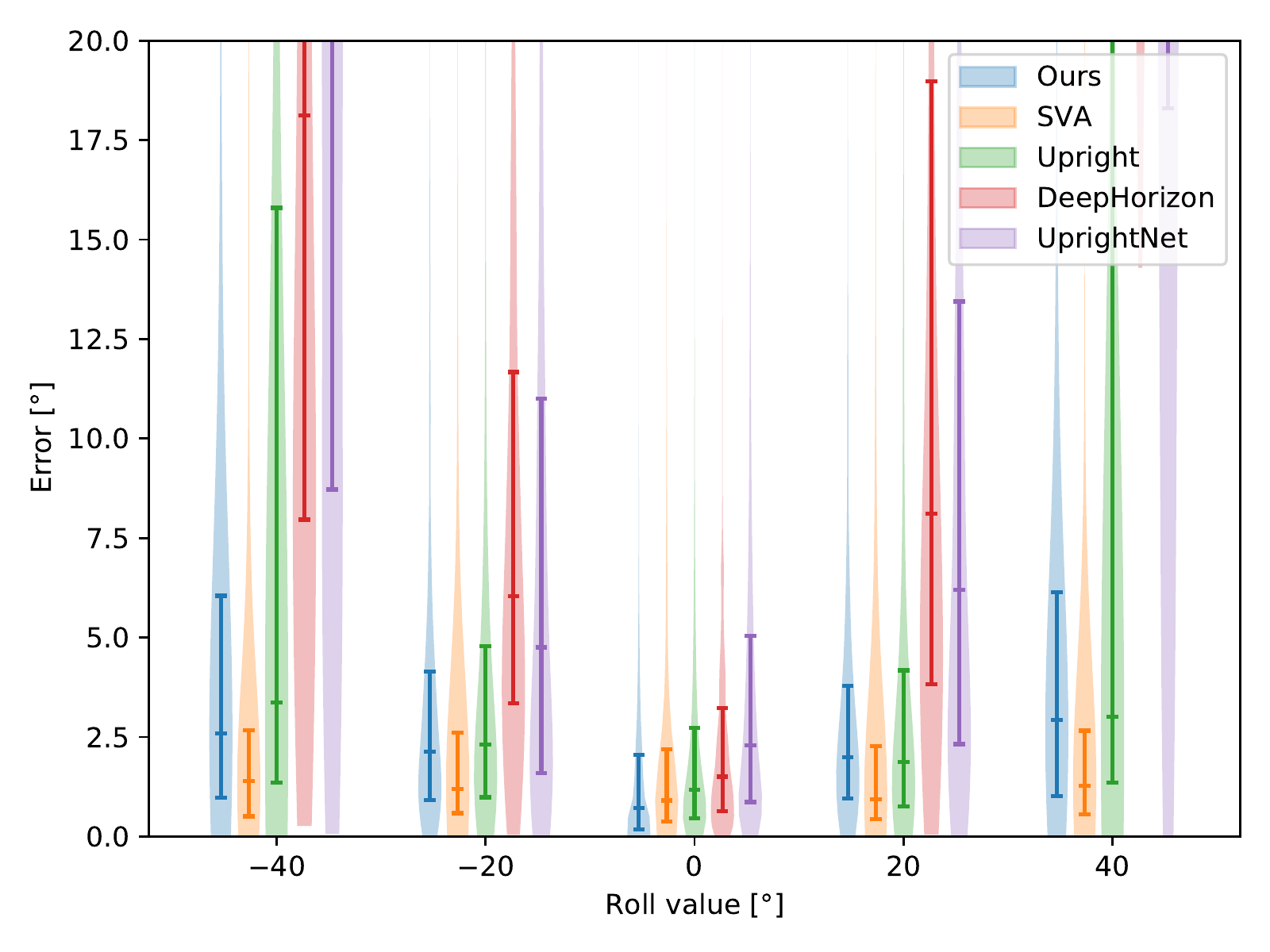} & 
    \includegraphics[width=.247\textwidth]{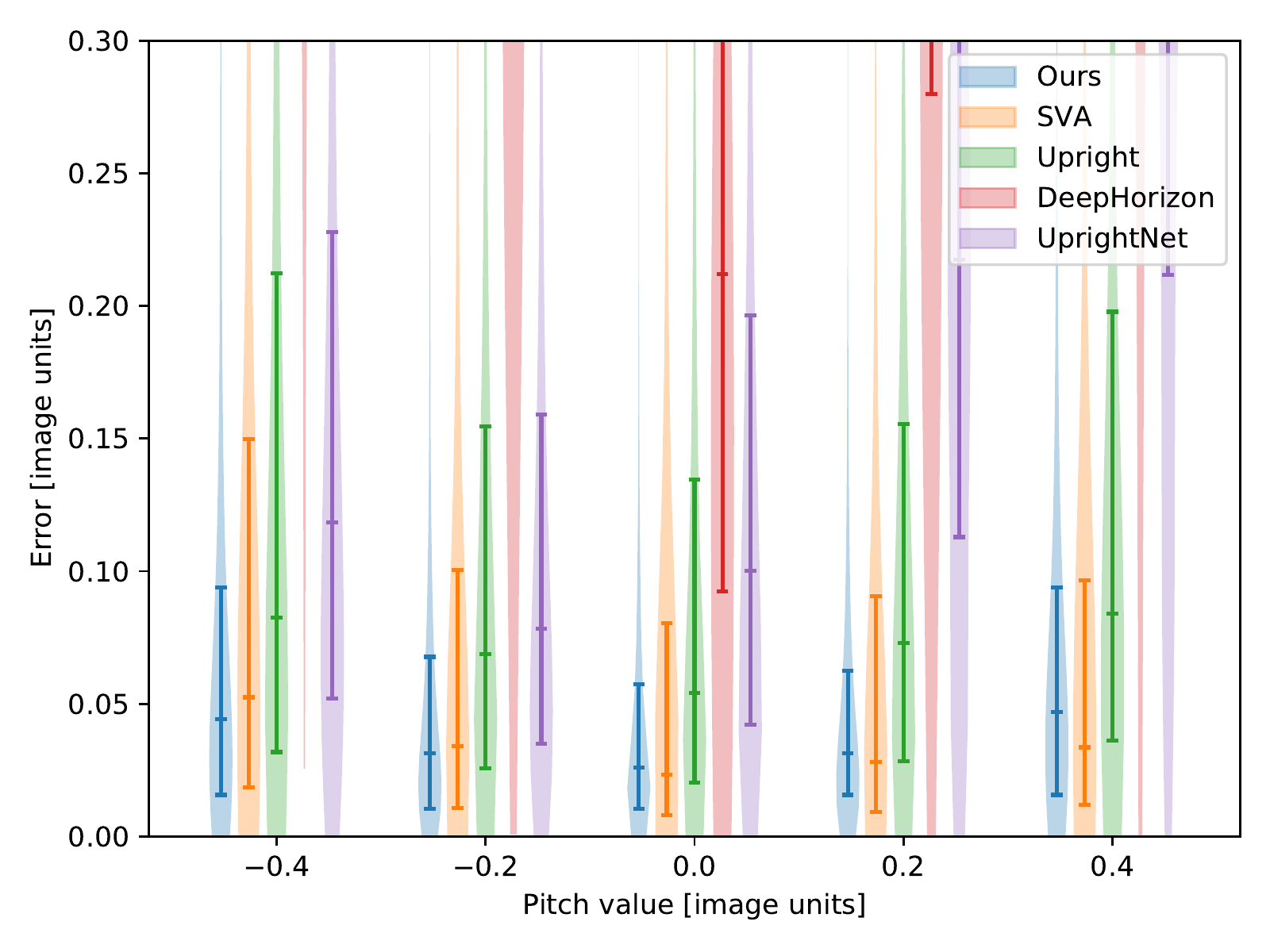} & 
    \includegraphics[width=.247\textwidth]{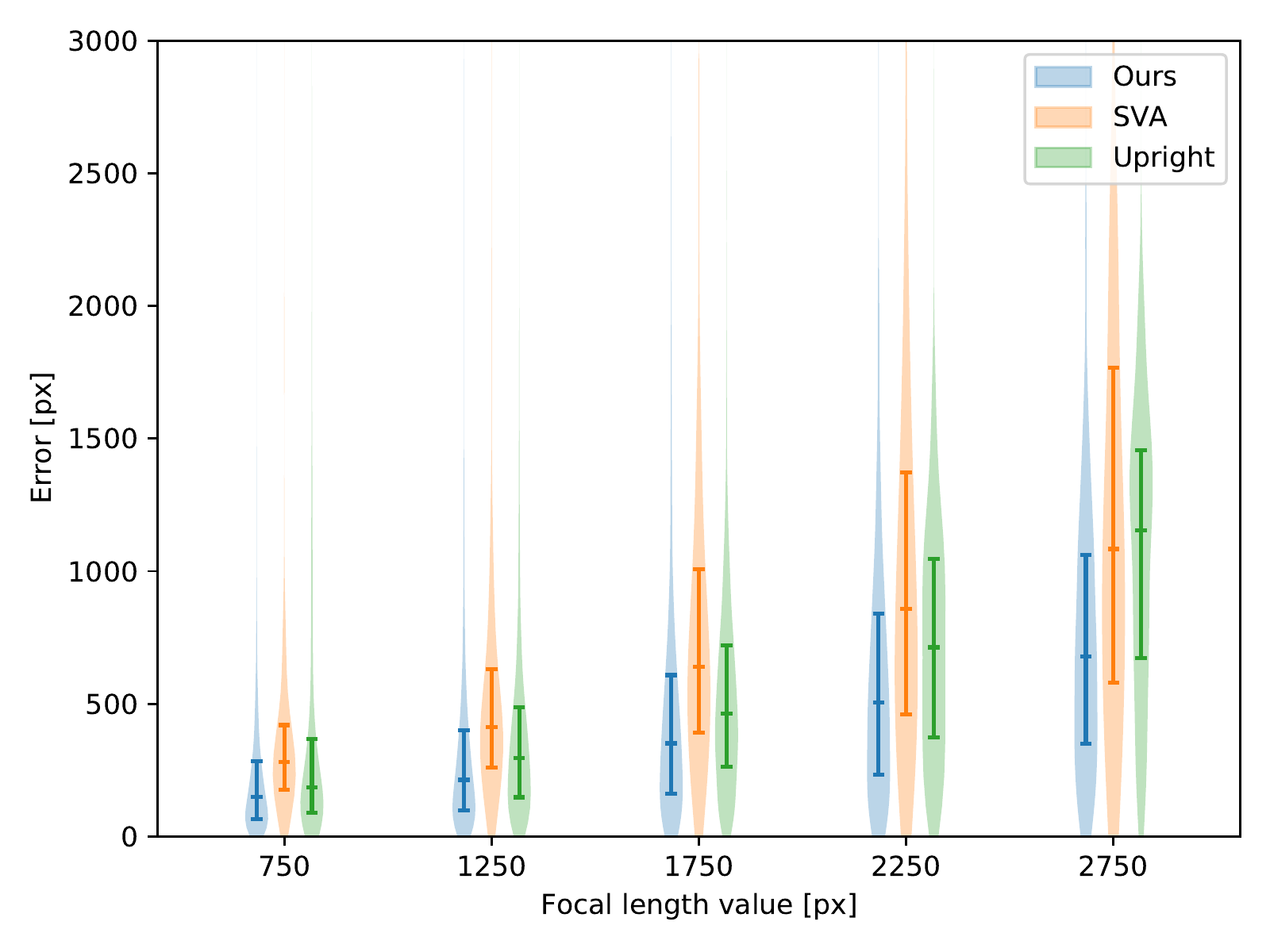} & 
    \includegraphics[width=.247\textwidth]{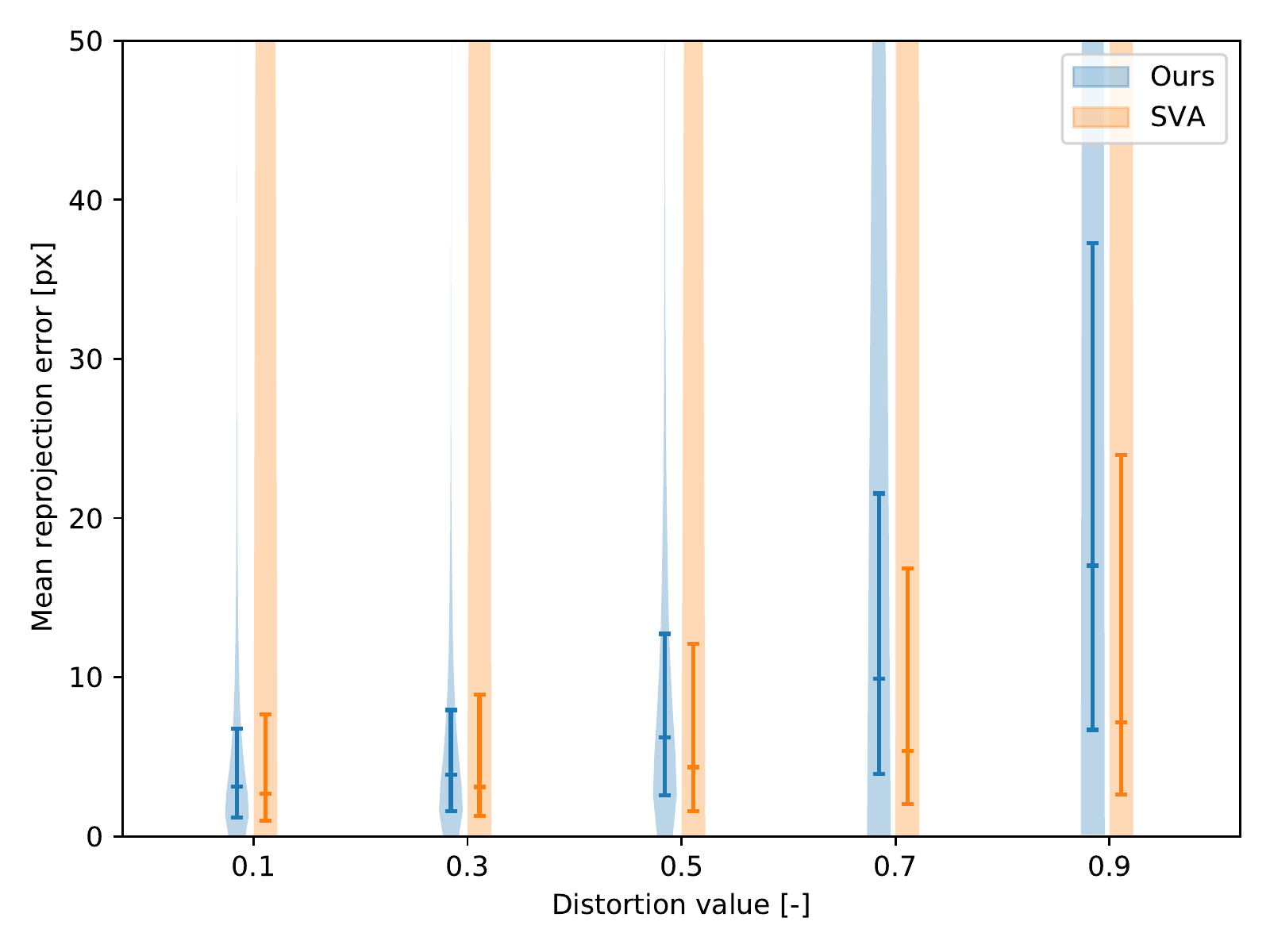} \vspace{-1em}\\
\hspace{1.5em} \tiny (a) & \hspace{1em} \tiny (b) & 
\hspace{1.5em} \tiny (c) & \hspace{1em} \tiny (d) \\
\end{tabular} \vspace{-0.8em}
    \caption{Quantitative comparison of (a) roll, (b) pitch, (c) focal length, (d) distortion. Note that SVA~\cite{Lochman_minimal_2021_WACV} fails on 49\% of the images, when there are not enough distinguishable edges to be detected with confidence. 
    }
    \label{fig:results_quantitative}
\end{figure*}
\addtolength{\tabcolsep}{5pt}

Our goal is to train a deep network to estimate the horizon line (parametrized by $\psi$ and $v_\mathrm{m}$), field of view $h_\theta$ and distortion $\xi$ from a single image. In this section, we present our CNN architecture for single image calibration and compare it to state-of-the-art estimation methods. To train this model, we need a large number of images and their corresponding camera parameters. \change{In the literature, \cite{Workman2016} provides images with ground truth horizon lines, a subset of which also have field of view annotations~\cite{Wilson2014}. However, these images are captured with cameras sharing similar parameters, with low fields of view and distortions.}

\subsection{Dataset}
\label{sec:dataset_generation}

\begin{table}[!t]
\centering
\footnotesize
\begin{tabular}{lll}
\toprule
Parameter & Distribution & Values \\
\midrule
Focal length (mm) & Lognormal & $\mu=14, \sigma=16$ \\
Horizon (image units) & Normal & $\mu=0.523, \sigma=0.3$ \\
Roll ($^\circ$) & Cauchy & $x_0 = 0$, $\gamma \in \{0.001, 0.1\}$ \\
Aspect ratio & Varying & $\{1{:}1, 5{:}4, 4{:}3, 3{:}2, 16{:}9\}$ \\
Distortion & Triangular & $c \in \{0.3, 1\}$ \\
\bottomrule
\end{tabular}
\caption{Sampling of camera parameters used to generate the dataset for the human sensitivity study.}
\label{tab:parameters-sampling}
\end{table}

To obtain a dataset of images and their ground truth camera parameters, we take inspiration from \cite{Workman2016,Hold-Geoffroy2017} and leverage a dataset of 30,000 $\; 360^\circ$ panoramas obtained from 360Cities\footnote{\url{https://www.360cities.net}, \change{obtained through a license allowing for publication but not redistribution.}}. We extract 7 rectified images from each panorama using our projection model from sec.~\ref{sec:intrinsics}. To obtain reasonable camera parameters, the following sampling strategies, summarized in table~\ref{tab:parameters-sampling}, were employed. Note that for the camera roll, two different Cauchy distributions $\{0.001, 0.1\}$ are sampled with 0.33 and 0.66 probability respectively. This was done to model the fact that many photos typically have a roll close to 0. Additionally, the distortion $\xi$ was sampled from a triangular distribution with low and high parameters of $0$ and $1$, and modes $\{0.03, 0.\}$ sampled with 0.8 and 0.2 probability, respectively. Those two modes are used to generate a long-tail distribution bounded in $\left[ 0, 1 \right]$, yielding a distribution of distortion values empirically close to images found in the wild. The aspect ratios correspond to popular image formats on Flickr and ImageNet images. Note that a larger probability (0.66) was given to the $4{:}3$ aspect ratio as it is the most common. The other aspect ratios are given a probability of $0.11$. We resize the extracted images to $224\times224$ to fit the neural network input size. This results in a dataset of 205,865 pairs of photos and their corresponding camera parameters which we split into a training set of 186,690 pairs, a validation set of 2,000 pairs and a test set of 17,175 pairs. \change{We made sure that no training panorama was used to compute a crop in the test set.}

\subsection{Network architecture} 
\label{sec:architecture}

We adopt a DenseNet~\cite{Huang2016} model pretrained on ImageNet~\cite{Russakovsky2015} and replace the last layer with four separate heads which estimate horizon angle $\psi$, the horizon distance to the center of the image $v_\mathrm{m}$, the field of view of the image $h_{\theta}$, and the distortion $\xi$ respectively. Instead of predicting $f$ and $\xi$, our network is trained to output $h_\theta$ and $\xi$. We experimented with all possible pairs of the three parameters and found that estimating $\xi$ and $h_\theta$ gave overall the most accurate results. We hypothesize that the dynamic range and distributions of those two parameters better fit the network initialization and training procedure. 
All output layers use the softmax activation function, which was also used in~\cite{Workman2016}. 
We adopt a range of $\left[ -\nicefrac{\pi}{2}, \nicefrac{\pi}{2} \right]$ for $\psi$ and $\left[ -1.6, 1.6\right]$ for $v_\mathrm{m}$. For roll $\psi$, we use smaller bins around 0 for finer estimations around those values with the relationship $a - b\exp\left( \text{-}2\psi^2 \right)$. We empirically choose $a\!=\!0.044$ and $b\!=\!0.04$. For $h_\theta$ and $\xi$, we use 256 uniform bins on the intervals $\left[ 0.33, \; 2.6 \right]$ rad and $\left[ 0, \; 1 \right]$, respectively. We employ the Kullback-Leibler loss on all outputs and train the model using the Adam optimizer \cite{Kingma2015}. 
Please see our supplementary material for training details.

%% file: sec_results.tex
\section{Results}

\subsection{Comparison \changeagain{with single image calibration methods}}

We now compare our method against multiple single-image camera parameters and horizon estimation methods, including Single View Autocalibration~\cite{Lochman_minimal_2021_WACV}, Upright~\cite{Lee2014}, UprightNet~\cite{xian2019uprightnet}, and DeepHorizon~\cite{Workman2016}. Qualitative results are shown in fig.~\ref{fig:results_qualitative}. 
\change{%
Note that \cite{Lochman_minimal_2021_WACV} and \cite{Lee2014} are not learning-based and \cite{xian2019uprightnet} requires ground truth per-pixel surface normals for training---as such, we do not retrain on our training set.}
We also observe that optimization-based methods \cite{Lee2014,Lochman_minimal_2021_WACV} provide good accuracy on urban scenes, where visual cues for vanishing points are abundant. 

Quantitative results are shown on fig.~\ref{fig:results_quantitative}. We note that our method provides state-of-the-art accuracy on our test set. \change{We hypothesize that our larger and more diverse training dataset and newer architecture provide enhanced generalization capabilities compared to training on fewer scenes~\cite{Workman2016} or indoor rooms only~\cite{xian2019uprightnet}.}

\subsection{Comparison \changeagain{with multi-image calibration methods}}
\label{sec:comparison_toolbox}


We now compare our method to state-of-the-art multi-image calibration methods. \changeagain{We emphasize that our work does not aim to compete with multi-image approaches which typically obtain subpixel reprojection accuracy. These methods require multiple images of a reference object or manual human annotations to work, which typically takes a long time to be performed. In contrast, our method can be executed very quickly on a generic scene (e.g., image from the internet). Despite this, our method still proposes visually pleasing results for most tasks related to image editing such as virtual object insertion, which we demonstrate in the rest of the paper. }

In table~\ref{Tab::CaMcal}, we evaluate the following calibration methods:
Mei's toolbox~\cite{mei2007single} (based on the spherical model), Zhang~\cite{Zhang:TPAMI:00} \changeagain{(using the Brown model~\cite{duane1971close})}, Fitzgibbon's division model~\cite{fitzgibbon2001simultaneous} and Scaramuzza's Toolbox~\cite{Scaramuzza:IROS:06}. Details on the camera setups are provided in the supp. material.
Our method provides competitive results on wide angle lenses. 
We also notice that the Brown model is not suitable to handle high distortion~\cite{kannala2006generic}, leading to many failure cases. 

\input{tab_compquant}

\subsection{Feature analysis} We use guided backpropagation~\cite{Springenberg2015} to understand the image features our CNN-based method focuses on to perform its estimation. We use the smoothgrad~\cite{Smilkov2017} version of guided backpropagation (SGB) to obtain a more stable analysis. Qualitative results are shown in fig.~\ref{fig:nn_analysis_smoothed-guided-back-propagation}. Note how edges representing vanishing lines are highlighted by SGB in accordance to the features used by geometry-based approaches and how human reason about such concepts. When no clear vanishing line is detected in the image, the CNN model tends to focus on boundaries between sky and land, as the horizon typically lies on or below this boundary. 
\changeagain{Such emergent behavior inspires us to quantify human sensitivity, which we investigate in sec.~\ref{sec:human_sensitivity_analysis}.}

%% file: tab_compquant.tex
\begin{table}[tb]
\centering
\footnotesize
\caption{Camera calibration results for different cameras (columns) and calibration methods (rows). 
All compared methods use multiple checkerboard pictures, while ours require a single picture of a general scene. Failures cases are noted N/A. Please see the supplementary materials for the estimated parameters.
} 
\begin{tabular}{lcccc}
\toprule
& Avenir & Avenir & \multirow{2}{*}{GoPro} & \multirow{2}{*}{Fisheye} \\
& 2.8mm & 4mm & & \\
\midrule
Ours                                        & 0.64  & 0.57  & 2.74    & 1.25 \\
Mei~\cite{mei2007single}                    & 0.12  & 0.13  & 0.10    & 0.10 \\
\changeagain{Fitzgibbon}~\cite{fitzgibbon2001simultaneous}  & N/A   & N/A   & 0.7     & 1.05 \\
\changeagain{Zhang}~\cite{Zhang:TPAMI:00}                 & 0.20  & 0.29  & N/A     & N/A  \\ 
Scaramuzza~\cite{Scaramuzza:IROS:06}        & 1.01  & 0.54  & 0.82    & 1.48 \\
\bottomrule
\end{tabular}
\label{Tab::CaMcal}
\vspace{-2mm}
\end{table}

%% file: sec_applications.tex
\section{Applications}
\label{sec:applications}

We now demonstrate different uses for camera parameter estimation from a single image. 
Please consult the supplementary material for additional results. 




\myparagraph{3D reconstruction}
\label{Sec::Reconstruction}
We acquired roughly 300 images of a building from a PointGrey Flea3 camera with an Avenir 4mm lens. 
We show the resulting 3D reconstruction obtained from COLMAP~\cite{schoenberger2016sfm,schoenberger2016mvs} in fig.~\ref{Fig:Recons3D}, using both undistorted images using our method (top) and the original images (bottom). Our method successfully corrects the large distortions present in the input images and allows for a robust 3D reconstruction of the scene, while the original sequence yields curved surfaces. 


\input{fig_recon3d}

\myparagraph{Undistortion of images in the wild}
\label{Sec::UnDistild}
To challenge the robustness of our algorithm, we propose to undistort a set of wide FoV images downloaded from the Internet for which no ground truth distortion parameters nor any information about the cameras or lenses are available. To generate undistorted images, we undo the lens distortion estimated by our method using eq.~\ref{Eq::backPro} and reproject the image on a flat plane. 
%
%
A set of representative results is available in fig.~\ref{fig:app_undistort}, showcasing our method robustness to diverse environments and lighting conditions.

\input{fig_undistort}

\myparagraph{Image retrieval} 
Our technique can be used to \change{match} images in large databases based on their camera parameters. To demonstrate this, we estimated the camera parameters using our technique on 10,000 images randomly selected from the Places2 dataset~\cite{Zhou2017} and computed the intersection of the horizon line with the left/right image boundaries (\change{which works better than matching the parameters themselves since this metric directly expresses camera parameters as pixel-based measurements}). 
Fig.~\ref{fig:applications_retrieval} presents the 4 closest matches in the dataset (according to the L2 distance) for 2 query images.

\input{fig_retrieval}

\myparagraph{Geometrically-consistent object transfer}
Transferring objects from one image to another requires matching the camera parameters. While previous techniques required the use of objects of known height in the image in order to infer camera parameters~\cite{lalonde-siggraph-07}, our approach estimates them from the image itself, and can be used to realistically transfer objects from one image to another, as shown in the supp. material. 

\begin{figure}[ht!]
\centering
\newcommand{\voiwidth}{0.36}
\begin{tabular}{@{}cc@{}}
\includegraphics[width=\voiwidth\linewidth]{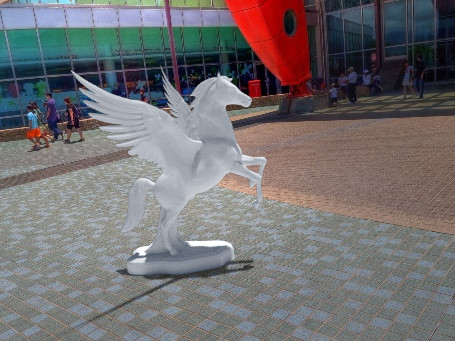} &
\includegraphics[width=\voiwidth\linewidth]{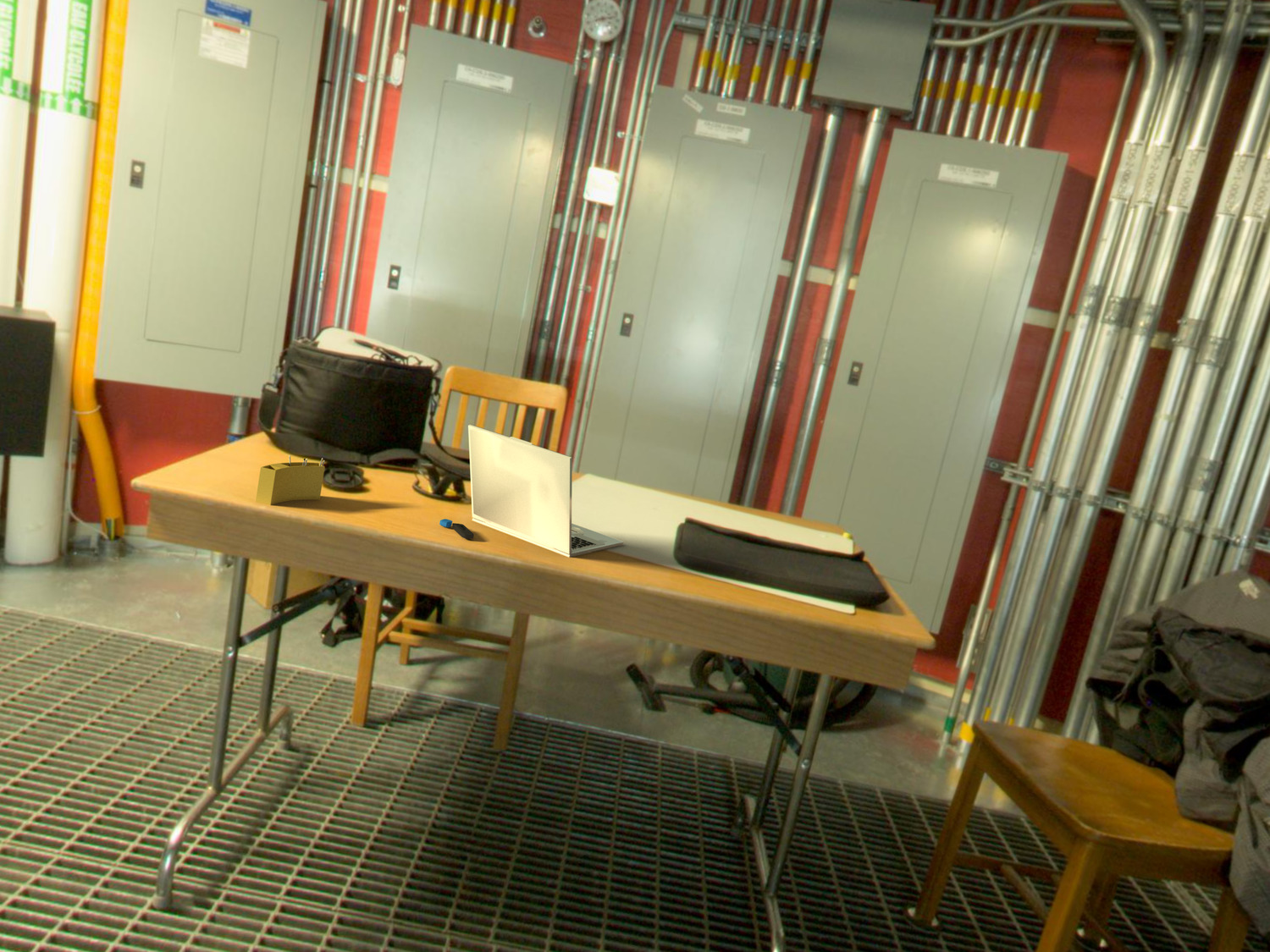} \\
\includegraphics[width=\voiwidth\linewidth]{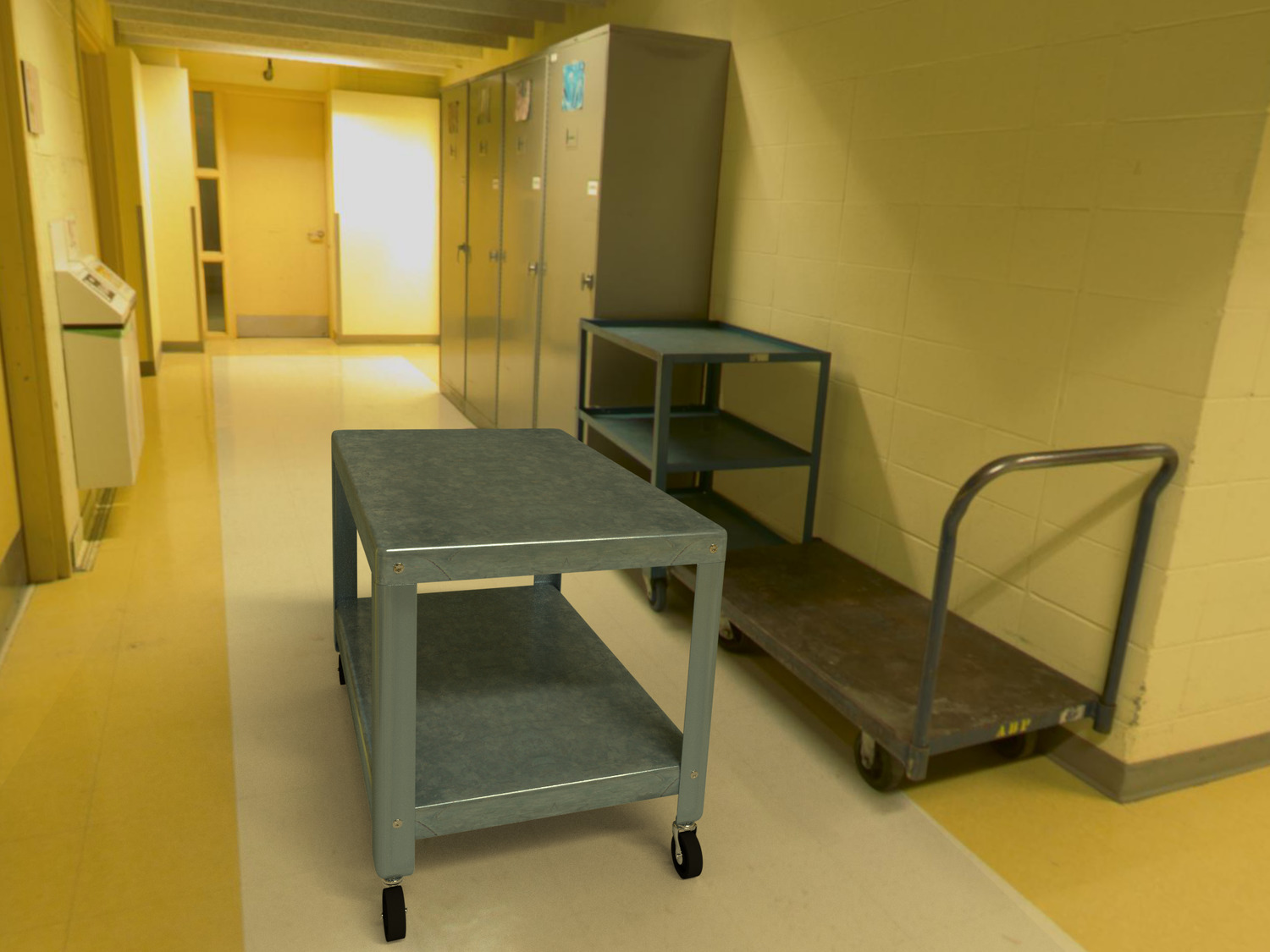} &
\includegraphics[width=\voiwidth\linewidth]{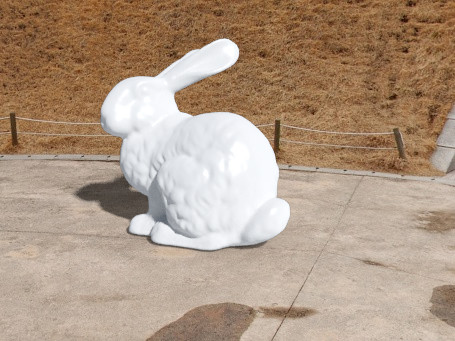}
\end{tabular}
\caption{Examples of virtual object insertions using our estimations\changeagain{, clockwise from top left: statue, laptop, bunny, cart.}}
\label{fig:applications_virtual_object_insertion}
\end{figure}

\myparagraph{Virtual object insertion}
Our approach also enables the realistic insertion of 3D objects in 2D images. As discussed in sec.~\ref{sec:human_sensitivity_analysis}, camera parameters are needed to plausibly align the virtual object with the background image. Given our automatic estimates, the user only needs to select an insertion point in the image and to specify the virtual camera height. Assuming the local scene around the object is a flat plane aligned with the horizon, we can automatically insert a virtual object and demonstrate several such examples in fig.~\ref{fig:applications_virtual_object_insertion}. For these results, the camera height was set to 1.6m and the lighting was automatically estimated with~\cite{Hold-Geoffroy2017,Gardner2017}.

%% file: fig_recon3d.tex
\begin{figure}[tb]
\centering
\footnotesize
\setlength{\tabcolsep}{1pt}
\begin{tabular}{cc}
\includegraphics[height=0.34\linewidth]{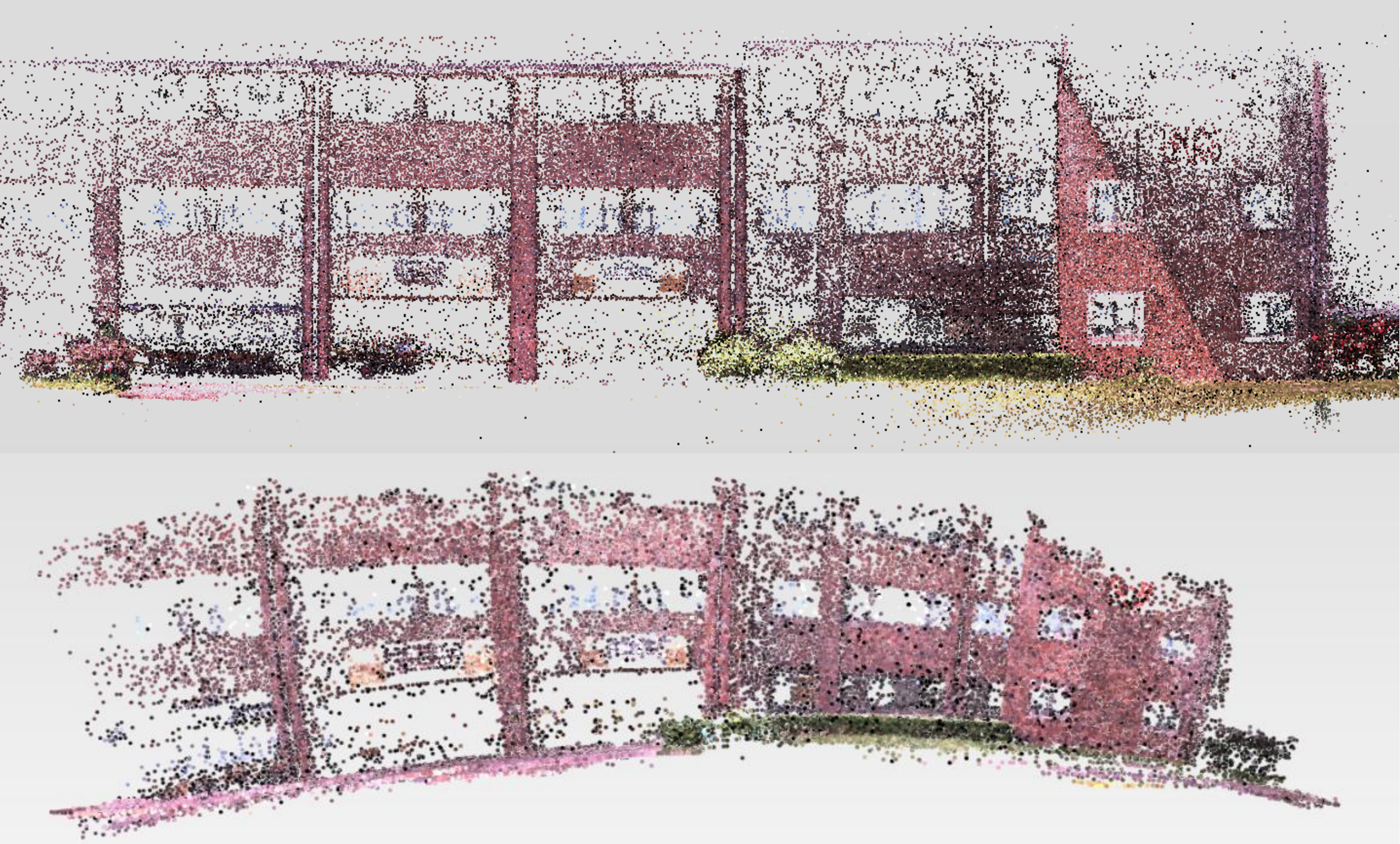} &
\includegraphics[height=0.34\linewidth]{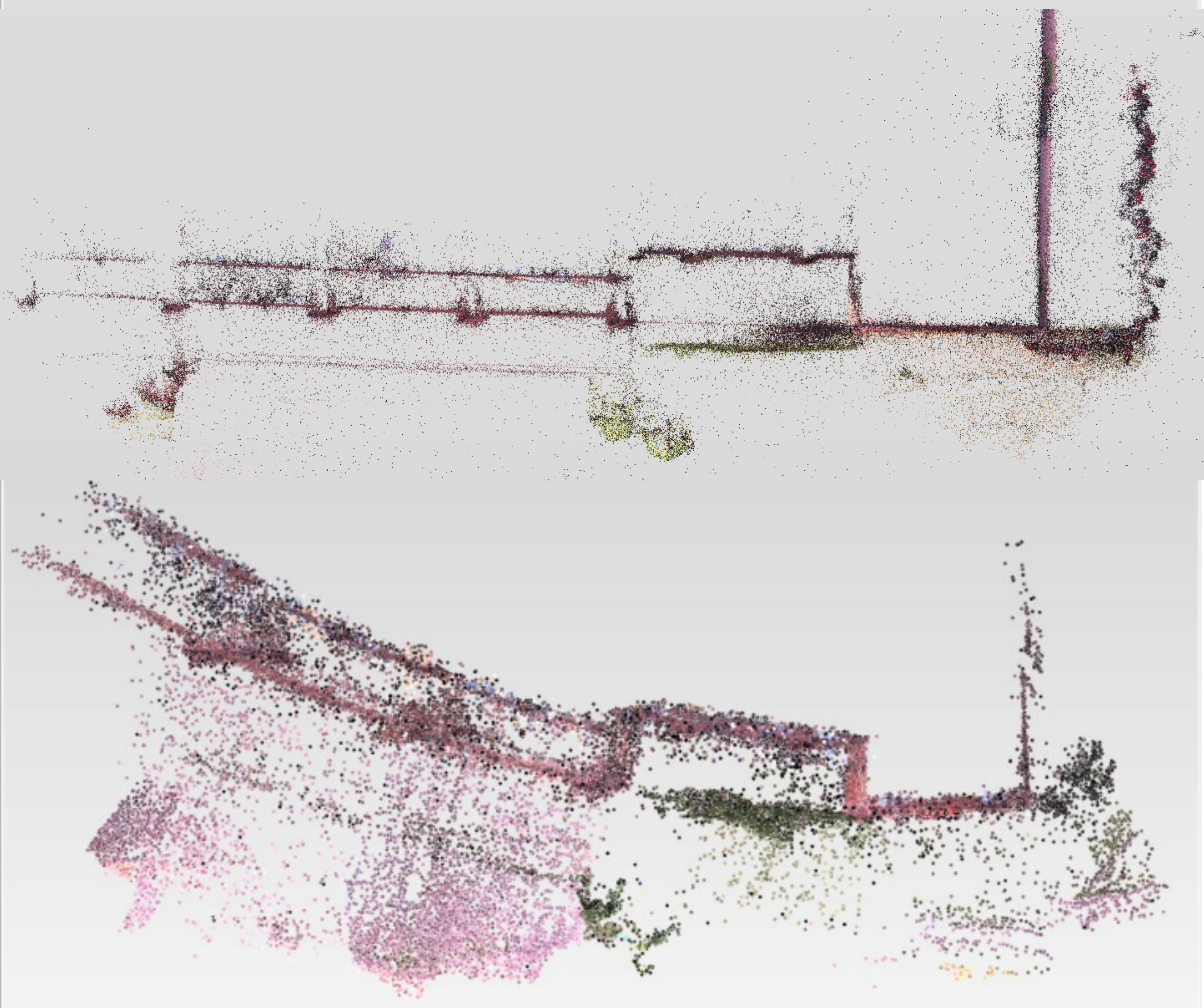} \\
(a) & (b)
\end{tabular}
\caption{3D reconstruction results: (a) front and (b) top views. Top row: results obtained with our parameters used for initialization step. Bottom: results obtained with the automatic initialization of intrinsic parameters.}
\label{Fig:Recons3D}
\end{figure}

%% file: fig_undistort.tex
\begin{figure*}[tb]
    \centering
    \newcommand{\undistortwidth}{0.324}
    \setlength{\tabcolsep}{2pt}
    \begin{tabular}{ccc}
         \footnotesize{$h_{\theta}=145\degree, \xi=0.93$}&\footnotesize{$h_{\theta}=143\degree, \xi=0.91$}&\footnotesize{$h_{\theta}=149\degree, \xi=0.81$}\\
         \includegraphics[width=\undistortwidth\linewidth,keepaspectratio]{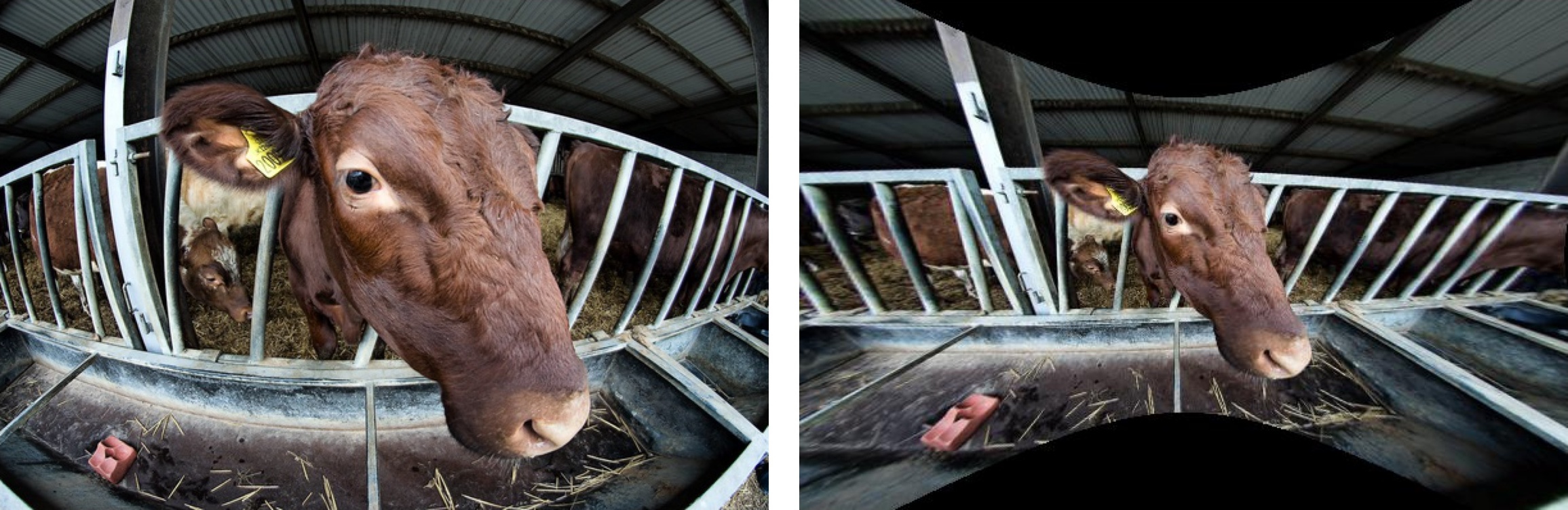}&
         \includegraphics[width=\undistortwidth\linewidth,keepaspectratio]{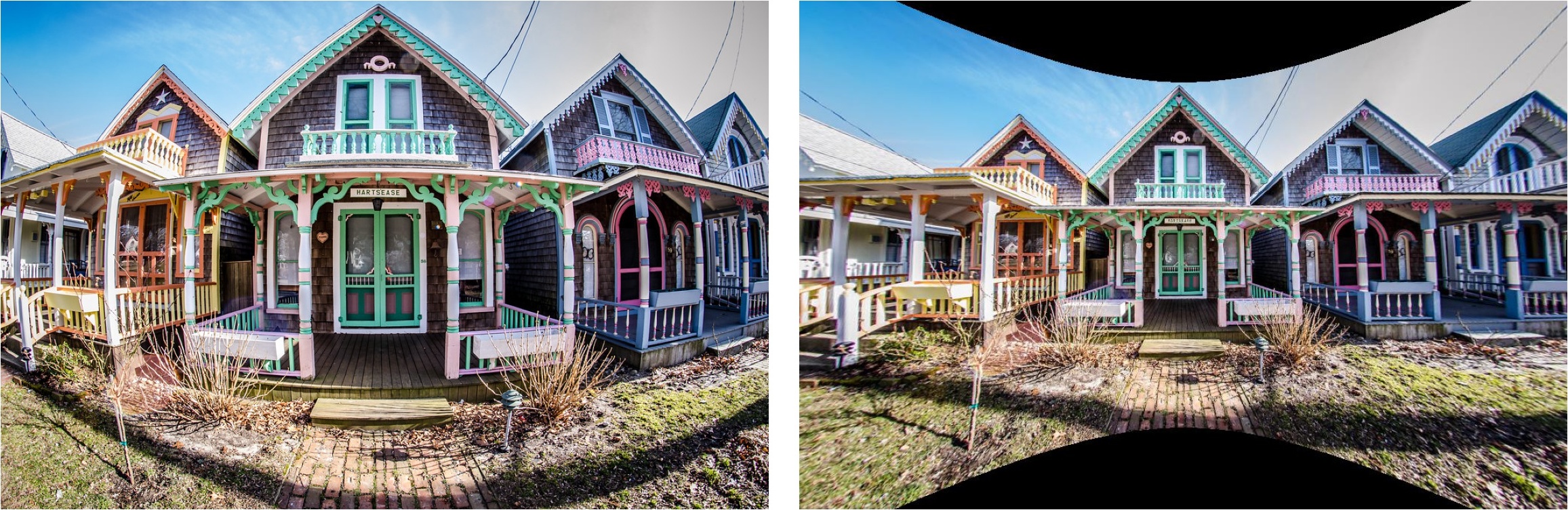}&
         
         \includegraphics[width=\undistortwidth\linewidth,keepaspectratio]{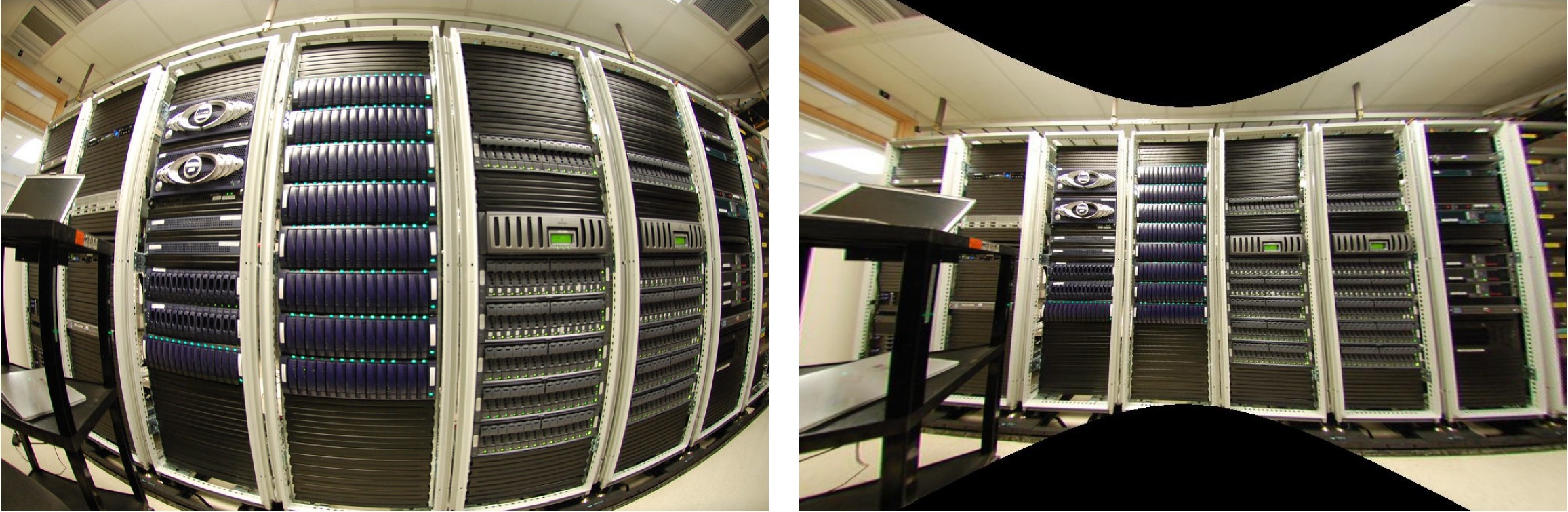}\\
        \footnotesize{$h_{\theta}=115\degree, \xi=$0.85}&\footnotesize{$h_{\theta}=149\degree, \xi=0.77$}&\footnotesize{$h_{\theta}=145\degree, \xi=0.98$}\\
         \includegraphics[width=\undistortwidth\linewidth,keepaspectratio]{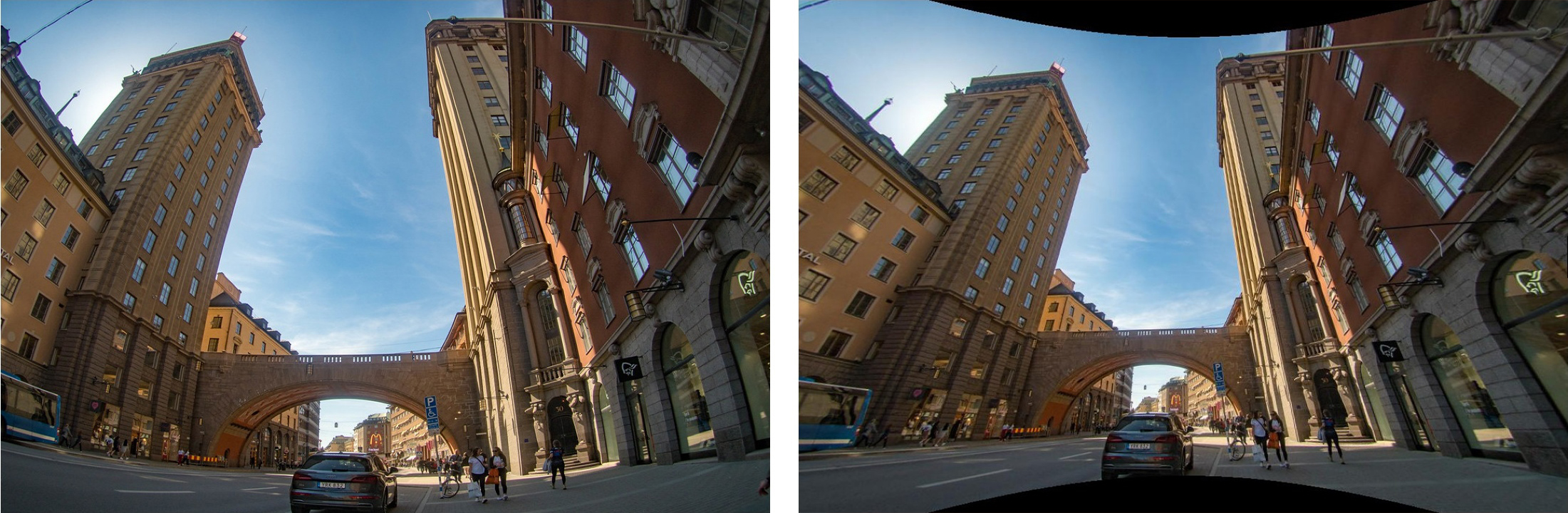}&
         \includegraphics[width=\undistortwidth\linewidth,keepaspectratio]{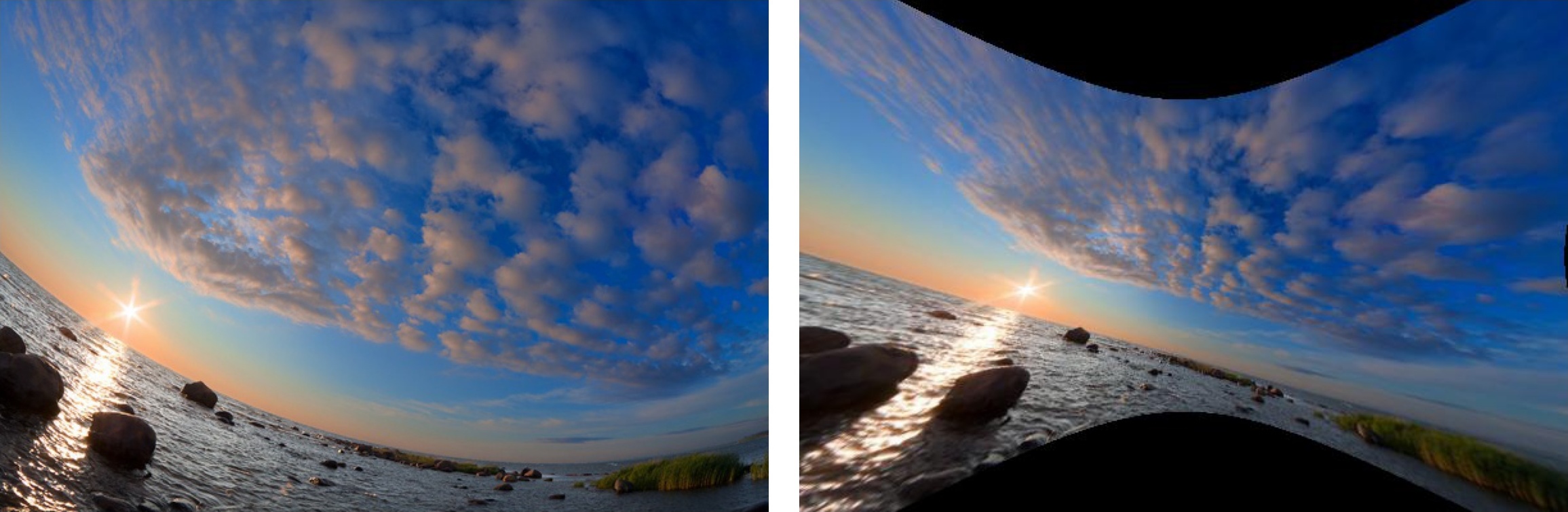}&
         \includegraphics[width=\undistortwidth\linewidth,keepaspectratio]{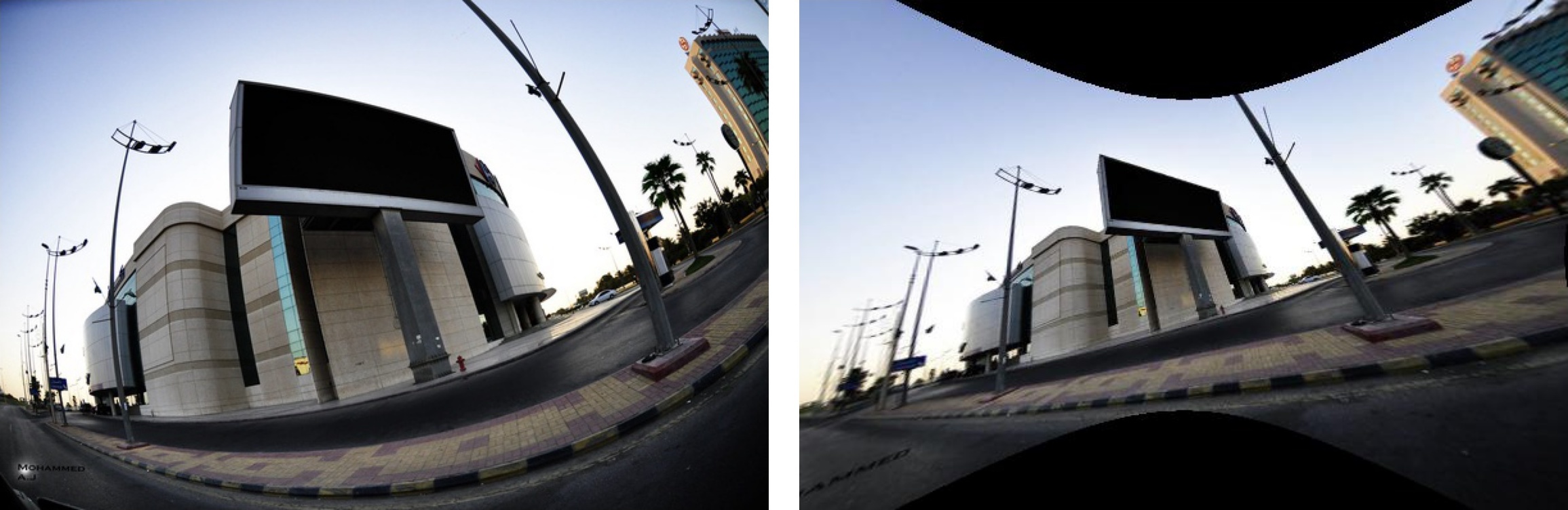}\\
         \footnotesize{$h_{\theta}=143\degree, \xi=0.87$}&\footnotesize{$h_{\theta}=149\degree, \xi=0.79$}&\footnotesize{$h_{\theta}=149\degree, \xi=0.90$}\\
         \includegraphics[width=\undistortwidth\linewidth,keepaspectratio]{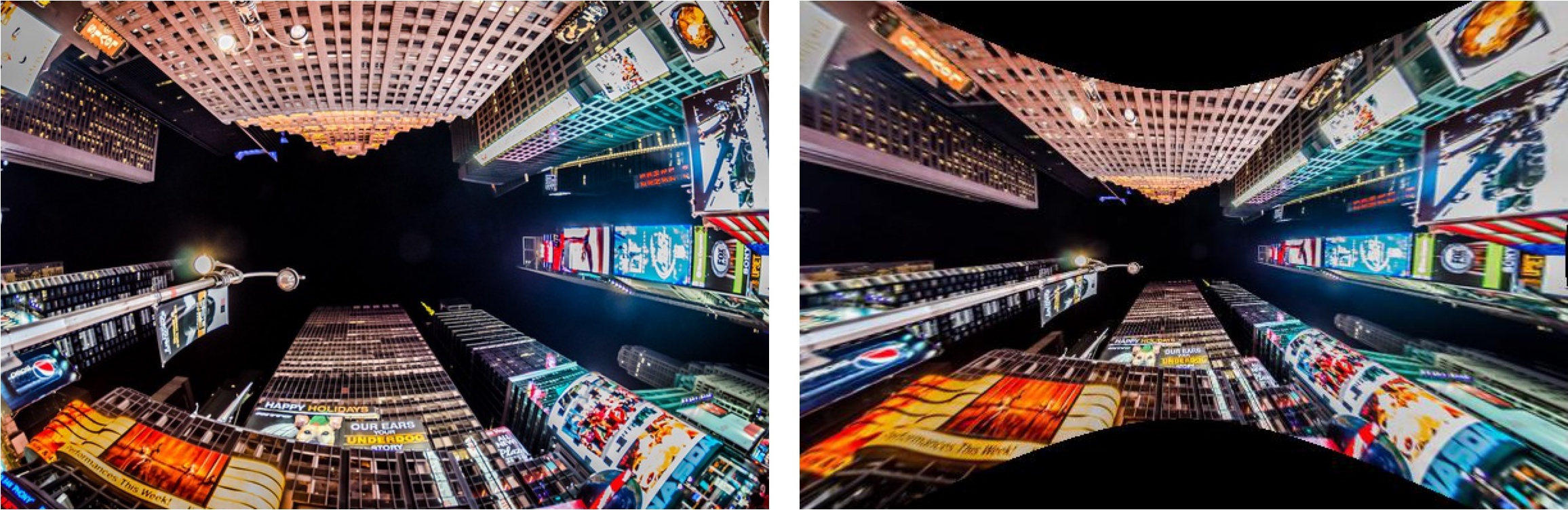}&
         \includegraphics[width=\undistortwidth\linewidth,keepaspectratio]{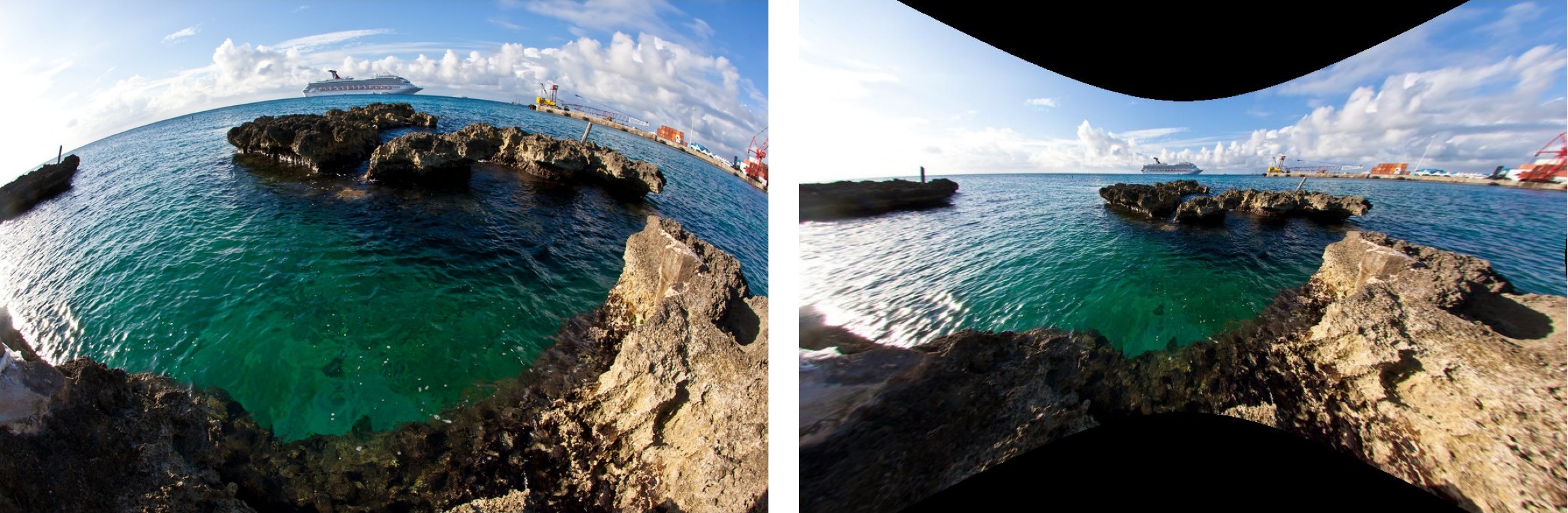}&
         \includegraphics[width=\undistortwidth\linewidth,keepaspectratio]{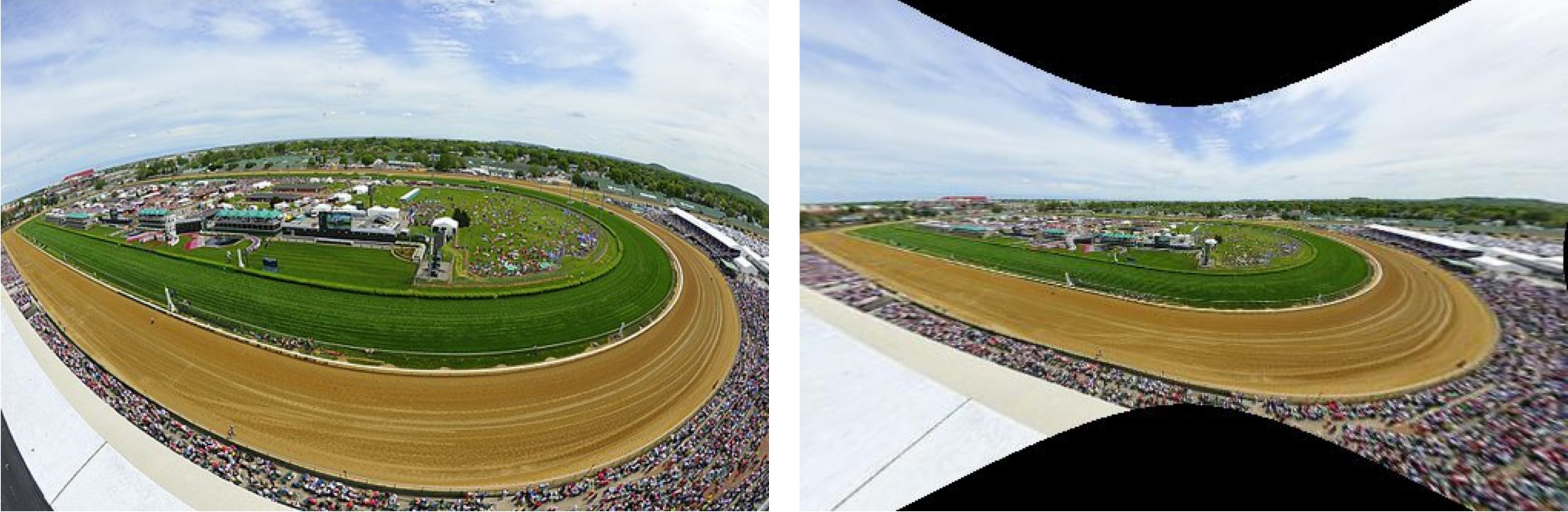}\\
    \end{tabular}
    \vspace{-1em}
    \caption{Examples of automatic undistortion results on images in the wild, \change{with the estimated field of view $h_\theta$ and distortion $\xi$}. Left: Original image. Right: output of our algorithm. Our approach works on a variety of images, including close/far objects, indoor/outdoor, true color/heavily edited, vertical/tilted viewpoint, and ground/aerial views. \textbf{See the supplementary material for more results.} }
    \label{fig:app_undistort}
\end{figure*}

%% file: fig_retrieval.tex
\begin{figure}[t]
\centering
\footnotesize
\setlength{\tabcolsep}{3pt}
\newcommand{\retrievalwidth}{0.125}
\begin{tabular}{c|cccc}
\includegraphics[height=\retrievalwidth\linewidth]{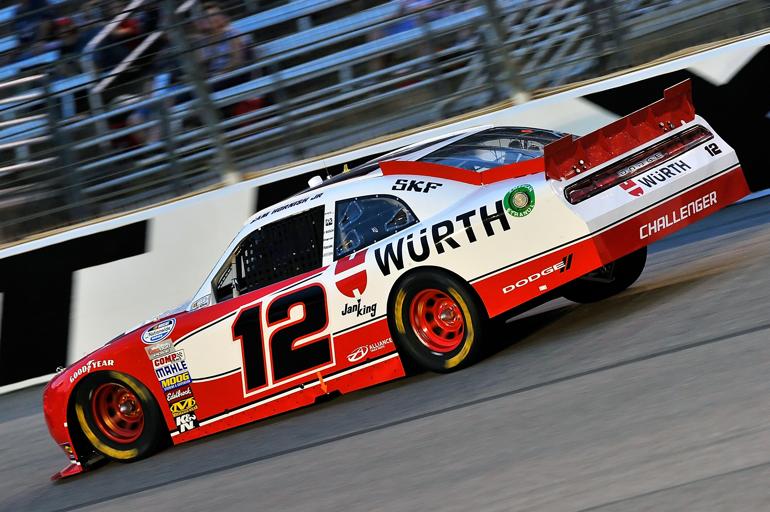} &
\includegraphics[height=\retrievalwidth\linewidth]{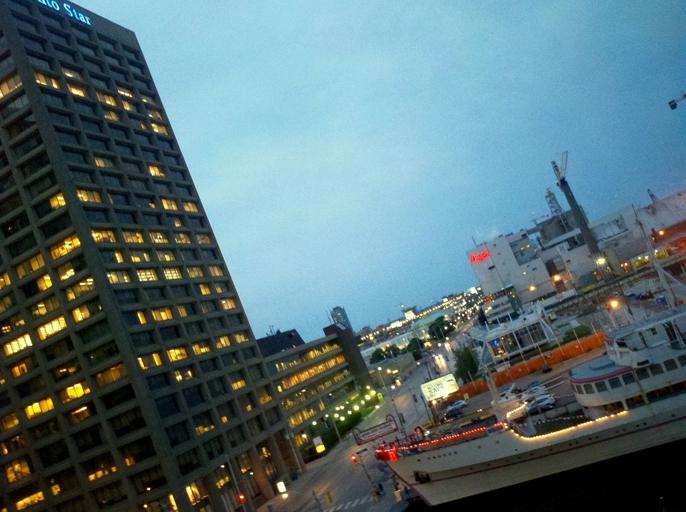} &
\includegraphics[height=\retrievalwidth\linewidth]{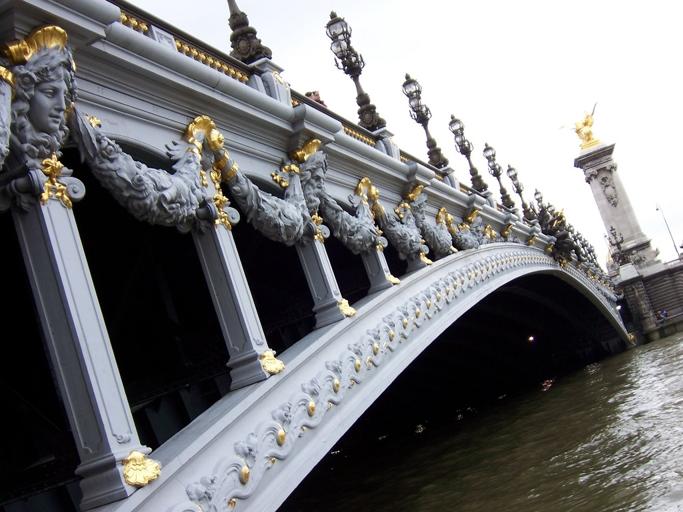} &
\includegraphics[height=\retrievalwidth\linewidth]{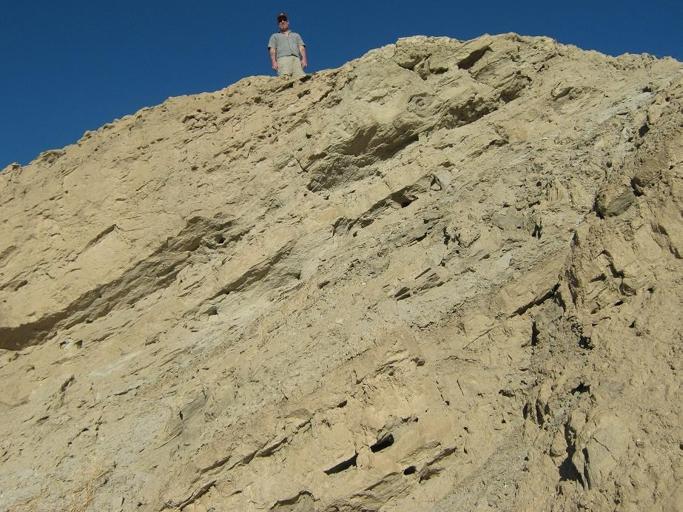} &
\includegraphics[height=\retrievalwidth\linewidth]{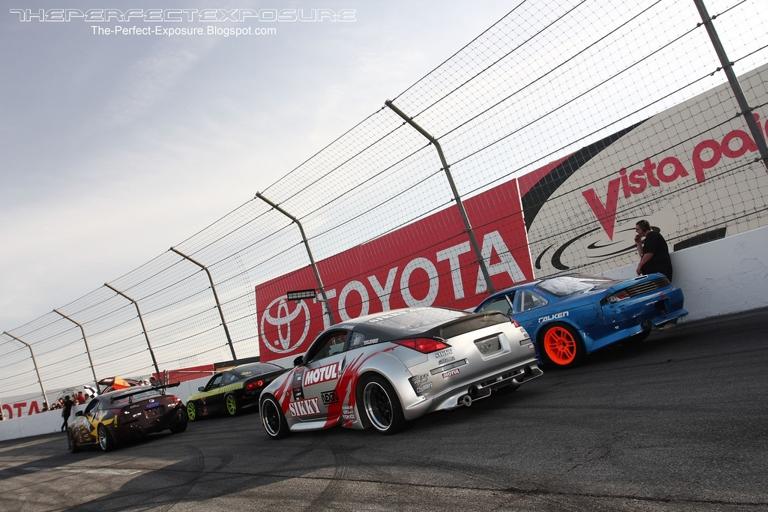} \\
\includegraphics[height=\retrievalwidth\linewidth]{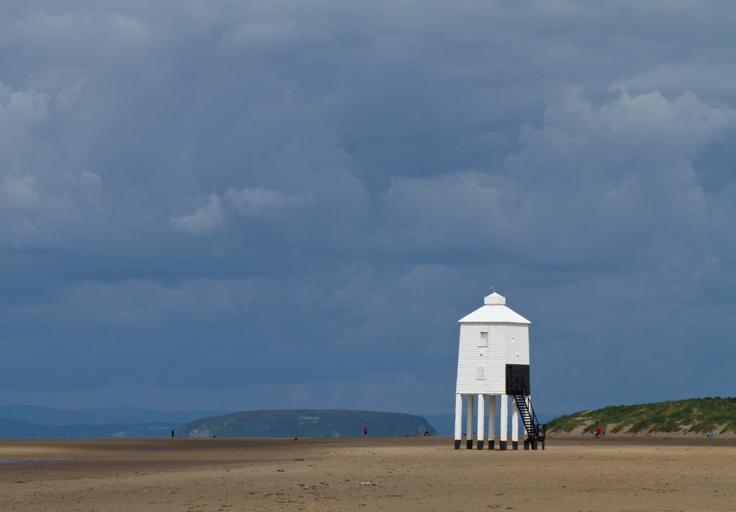} &
\includegraphics[height=\retrievalwidth\linewidth]{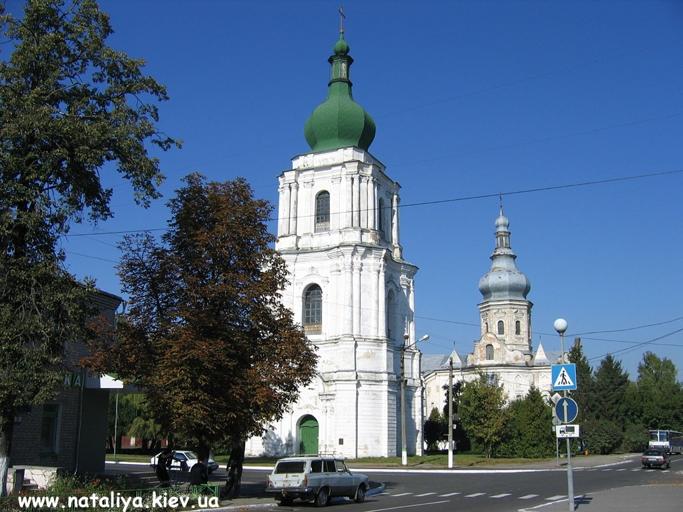} &
\includegraphics[height=\retrievalwidth\linewidth]{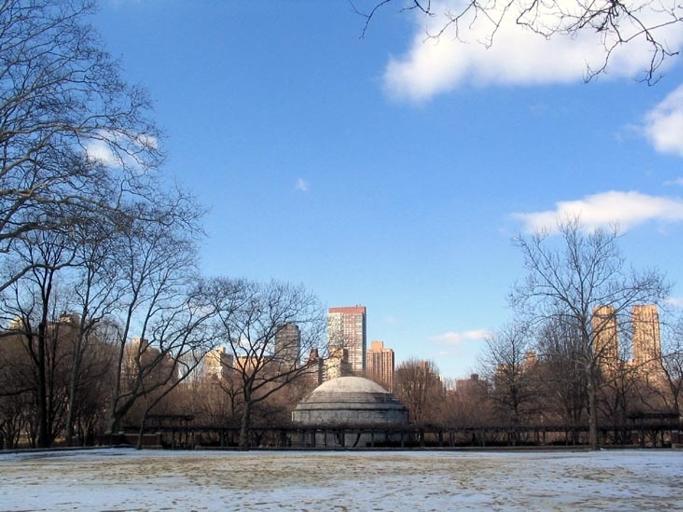} &
\includegraphics[height=\retrievalwidth\linewidth]{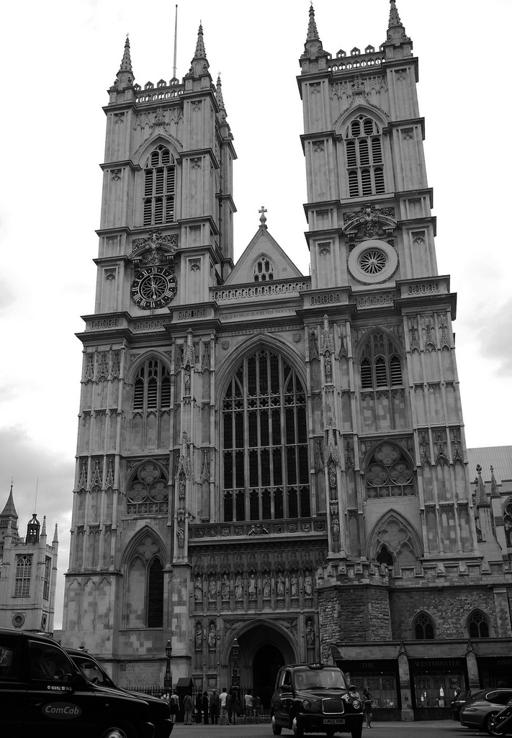} &
\includegraphics[height=\retrievalwidth\linewidth]{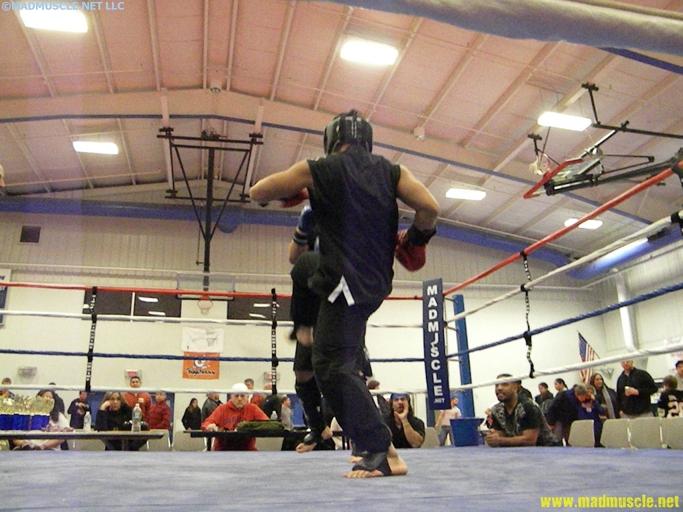} \\
Query & NN1 & NN2 & NN3 & NN4
\end{tabular}
\caption{Examples of image retrieval by horizon location on Places2. The horizon line is estimated using our method from the query image, and used to find closest matches in a 10k random subset from Places2. The top-4 matches are shown on the right.}
\label{fig:applications_retrieval}
\end{figure}

%% file: sec_human_perception.tex
\section{Human perception of calibration}
\label{sec:human_sensitivity_analysis}


\input{fig_backprop}

\changeagain{We now present two main studies to better understand the impact of camera calibration errors on human perception.}
We leverage virtual object insertion as a tool for evaluating whether an error in camera calibration can be detected by a human observer. Since the space of parameters is large, we \change{separate} it into two \change{different experiments on the sensitivity of human observers to the studied camera parameters: Experiment 1} evaluates sensitivity with respect to camera pitch, roll, and field of view (FoV) in the case where the FoV is narrow; \change{and Experiment 2} studies sensitivity to lens distortion wide FoV images, where it is most apparent. 
%
%
\change{We first proceed to describe aspects that are common to both experiments, and follow with per-experiment specifics.}

\subsection{Method}
\label{sec:stimuli_generation}

\myparagraph{Stimuli}
To understand human perception of geometric camera calibration errors, the stimuli generated for both experiments consists of two versions of the same real image, \changeagain{displayed side by side}, in which a virtual object has been inserted. The virtual object is inserted according to: the ground truth camera parameters for the first version; and to perturbed camera parameters in the second. To generate this dataset, we first use the same process as described in sec.~\ref{sec:dataset_generation} to \change{extract an image with ground truth camera calibration from a panorama.} 
The Cycles \change{rendering engine}~\cite{blender} is used to insert a single virtual \change{3D} object at a manually-selected point in the image \change{on a flat horizontal surface}. \change{To generate believable cast shadows around the object, a flat virtual ground plane\footnote{\change{Set in Cycles as a ``shadow catcher'' which only shows the shadows cast upon it but not the plane itself}.} was set at $y=0$.} \change{The techniques of \cite{Gardner2017} (indoor) and \cite{Hold-Geoffroy2017} (outdoor) were used to produce the virtual light sources}. The virtual camera is placed at a height of 1.6m.


\changeagain{The} first image is obtained by setting the parameters of the \change{Cycles} camera to the ground truth parameters of the background image. The second is obtained by distorting the \change{Cycles} camera parameters, yielding a virtual object that does not have the same camera parameters as the background. 
In the distorted renders, the virtual object was, \change{after rendering}, \change{translated} and scaled in order to appear at the same location and have the same size in the image as the ground truth render. This step is needed as apparent size is a function of field of view and/or lens distortion. The virtual objects were laid vertically on the ground plane and were randomly rotated about their vertical axis. 
To limit biases caused by the virtual object inserted, different virtual objects were used in both experiments. 

\myparagraph{Procedure}
\change{Recruited subjects observed both images displayed side by side horizontally (the order being randomly permuted) on their own (uncontrolled) monitor. After being shown the stimuli, the subjects were asked the question ``select the image where the orientation of the virtual object looks most natural to you'' and responded by clicking on the image of their choice.} \change{Subjects} were specifically instructed to ignore the color, texture, shadows or lighting on or around the object and to focus on its coherence with the background. 


\input{fig_distortion_center}

\subsection{\change{Experiment 1 design: pitch, roll, (narrow) FoV}}
\label{sec:experiment-1}

\textbf{\change{Stimuli}}
\change{For Experiment 1}, 530 panoramas (79 indoor, 451 outdoor) \change{were randomly selected and 20 crops were extracted for each} (see sec.~\ref{sec:dataset_generation}), resulting in 10,368 images. 
%
%
The \change{camera} parameters were altered by randomly adding or subtracting values sampled from a uniform distribution in $\left[ 1, 30 \right]^\circ$ for pitch, $ \left[ 0.5, 20 \right]^\circ$ for roll and $\left[ 5, 55\right]^\circ$ for field of view. \change{No lens distortion was used, i.e., $\xi=0$}.

\noindent\textbf{\change{Procedure}}
\change{Experiment 1} employed 8 different virtual objects for insertion, ranging from simple geometric primitives (sphere, cone), real objects with clear vertical directions (toy rocket, metal barrel, the Eiffel tower) and objects with a somewhat organic shape (the Stanford bunny and a horse statue). Several examples of images generated using this process are shown in fig.~\ref{fig:pstudy_sensitivity_per_parameter} and in the supp. material. 

\changeagain{Due to the large number of parameters to evaluate}, participants were recruited on the Mechanical Turk platform. In total, 376 \change{subjects} provided 145,720 submissions, from which 124,740 were accepted, leading to 11 different \change{subjects} annotating each image on average. 4,319 of those submissions had a single distorted parameter, while the remaining 5,947 had two or more distorted parameters. \change{To handle inattentive subjects, we use the sentinels system of Gingold et al.~\cite{Gingold2012}. In this system, subjects are shown some ``sentinels'' to perform, which consist of images with easily identifiable ground-truth. Subjects} were presented sets of 20 image pairs at a time which contained 2 sentinels. 9 subjects were \change{removed since they failed the sentinel tasks}. An additional 520 annotations were rejected for inattentive \change{subjects} who failed some sentinels over a small time lapse. The median time spent on a single pair of images was 4 seconds. 

\myparagraph{Observations}
We found no statistical difference in the results obtained across the virtual objects, except for two: the sphere and the Stanford bunny where subjects were unable to identify the ground truth image except for significant field of view variations (at least $30^\circ$). The sphere being rotationally symmetric, it is unsurprising that it does not serve as a good instrument to determine object insertion errors. We theorize that the bunny, with its round shape, shares similarities to the sphere in that respect. 

We also found no statistical difference in the results when analyzing the impact of the object size, computed as the relative height of the object with respect to the image. Participants showed similar accuracy regardless of object size (ranging from 10\% to 85\% image height in our dataset). We manually labeled the images into pairs of outdoor-indoors and built-natural, and found no statistical difference between either subset. Finally, we analyzed the impact of the mean image brightness, hypothesizing that a mismatch in camera parameters may be more easily observable in brightly-lit images. We found no statistical differences between images of different mean intensities.

\input{fig_pstudy_overall}

\input{fig_pstudy_perparameter}

\subsection{\change{Experiment 2 design: distortion, (wide) FoV}}
\label{sec:experiment-2}

As opposed to errors in pitch, roll or field of view, human perception of errors in lens distortion seems to be highly dependent on background image content and on the position of the virtual object in the image. We hypothesize that: \change{1) humans perceive distortion better on urban scenes, where there are a multitude of straight lines, than on scenes in nature; and 2) the sensitivity to distortion is reduced when virtual objects are inserted near the image center.} 

\myparagraph{\change{Preliminary Experiment 2a}} To evaluate the significance of these two effects on our perception of errors in distortion, we begin by conducting \change{preliminary Experiment 2a} on 27 images of three different scene types: artificial (\change{cube with graduated lines, see fig.~\ref{fig:study-distortion-center}}), urban and nature. \change{This preliminary Experiment 2a aims to answer the question ``How does the image content affects human sensitivity to lens distortion?''}. The position of the object varies horizontally while the error on distortion is kept at a constant high value $\left(|{\Delta\xi}|=1\right)$. 18 participants \change{were recruited} from research labs, \change{and each observed 54 image pairs} corresponding to a total of 972 pairs analyzed. 

The \change{preliminary Experiment 2a results, displayed in fig.~\ref{fig:study-distortion-center}, show that observers} are much less sensitive to errors in distortion for nature scenes (average accuracy of 43\%) than for artificial (75\%) or urban (74\%) scenes (see fig.~\ref{fig:study-distortion-center}). Based on this observation, only urban images were used in \change{Experiment 2}. These results also show that the position of the object in the image does not have an impact on human sensitivity to errors in distortion, except when the object is inserted exactly in the center of the image. In this case, accuracy falls to nearly 50\%, meaning humans can not differentiate between the ground truth and distorted objects. Following these results, the objects were \change{not inserted in the middle 20\% of the image in Experiment 2}.

\myparagraph{\change{Stimuli}} We then conduct the full \change{Experiment 2}, where the error on distortion is varying. \change{This experiment explores ``How sensitive are humans to calibration errors in camera rotation, field of view and lens distortion?''. To answer this, } we generate 200 images (56 indoors and 144 outdoor), with distortion errors $\Delta\xi$ uniformly sampled from -1 to 1, while keeping the value $\xi \in [0, 1]$ and keep all the other parameters fixed. \change{To remove the variance related to shape in our distortion analysis, we only insert the square prism in all of the images since round shapes provide cues that are too ambiguous to assess distortion (see Experiment 1 in sec.~\ref{sec:experiment-1}).} \changeagain{Sentinels~\cite{Gingold2012} were also used in this experiment.}

\myparagraph{\change{Procedure}}
\change{122 participants, different than for Experiment 2a, were recruited from research labs. Each participant was shown a randomly-selected subset of up to 50 images, corresponding to 4843 pairs of images. } 

\subsection{\change{Experimental results}}

We now report on the results of \change{Experiments 1 and 2 jointly}. We \change{report on the observers sensitivity to errors in camera parameters first as a function of the ground truth value, and second as a function of both the ground truth and the error.}

\subsubsection{Error in camera parameters} 

We evaluate the sensitivity of \change{human observers} to errors in camera parameters for each investigated parameter independently and illustrate the results in fig.~\ref{fig:pstudy_overall_sensitivity}. This curve \change{was generated by computing} the percentage of times observers preferred the ground truth over the distorted image, and compute the median and percentiles across all images (and different virtual objects) that share the same amount of distortion. The higher the percentage in the $y$-axis, the more humans are prone to detect errors. Conversely, 50\% indicates perfect confusion: \change{subjects} were unable to distinguish between the ground truth and the distorted version. 

First, we note that when the error in camera parameters is close to 0, confusion nears 50\%, which is expected. What is interesting is how quickly sensitivity rises when increasing the error. 
For example, \change{subjects} could tolerate an error in pitch up to 0.2 in rescaled image units (see sec.~\ref{sec:ifm}), but beyond this threshold, \change{subjects} started to distinguish the distortion (fig.~\ref{fig:pstudy_overall_sensitivity}-a). The high sensitivity to roll errors is most prominent (fig.~\ref{fig:pstudy_overall_sensitivity}-b), where errors of $12^\circ$ and more are almost systematically detected and only a small range of approximately $\pm2.5^\circ$ roll error go unnoticed.

We note a large tolerance to negative errors in field of view (fig.~\ref{fig:pstudy_overall_sensitivity}-c). Large positive errors (right side of the plot) translate to rendering an object with a field of view that is larger than that of the background. This results is increased perspective effects on the object, which tend to be visible. On the other hand, negative errors indicate that the perspective effect is not as pronounced on the object as it should be with respect to the background image. In this scenario, \change{subjects} seemed to have been unable to differentiate between the ground truth and the distorted object. 
A similar trend for errors in lens distortion is observed in fig.~\ref{fig:pstudy_overall_sensitivity}-d). On images with negative errors, meaning the object has less distortion (i.e. is straighter) than the background, \change{subjects} could not reliably identify the object with correct calibration. However, \change{subjects} were increasingly sensitive to positive errors in distortion, where the object has a greater distortion than the background.

\subsubsection{Joint space of error in camera parameters and absolute parameter value}

To evaluate \change{if there are regions in parameter space where errors are more noticeable}, we plot \change{observer sensitivity to errors in camera parameters according to the joint} 2D space of errors and absolute parameter value in fig.~\ref{fig:pstudy_sensitivity_per_parameter}.
Note that white squares in these plots indicate impossible parameter/error configurations (e.g., ground plane is not visible in the image and thus no virtual object can be inserted, lens distortion values outside its admissible range, etc.).


The results suggest that observer sensitivity to pitch error \change{in the inserted object} does not correlate strongly with \change{pitch of the background image}: \change{larger error is more noticeable across the board}. Similarly, sensitivity to errors in field of view seems to be constant across all fields of view. However, camera roll appears to have an influence on the perception of roll error: images with high roll (\change{positive or negative}) allow more room for roll estimation errors. The lens distortion also has an influence on the perception of its error: only when the background has relatively little distortion do observers notice the discrepancy when the inserted object is significantly curved, \change{but not the other way around}. 

\subsubsection{Summary of findings}

\change{In summary, our Experiments reveal that: 1) larger absolute errors in pitch and roll are more noticeable; 2) larger positive errors in field of view and roll are more noticeable; 3) large roll, or low distortion, allow for errors to be less perceivable.}

\subsection{Evaluation on a perceptual measure}

\begin{figure}
    \centering
\begin{tabular}{cc}
    \includegraphics[trim=10 0 40 36,clip,width=.49\textwidth]{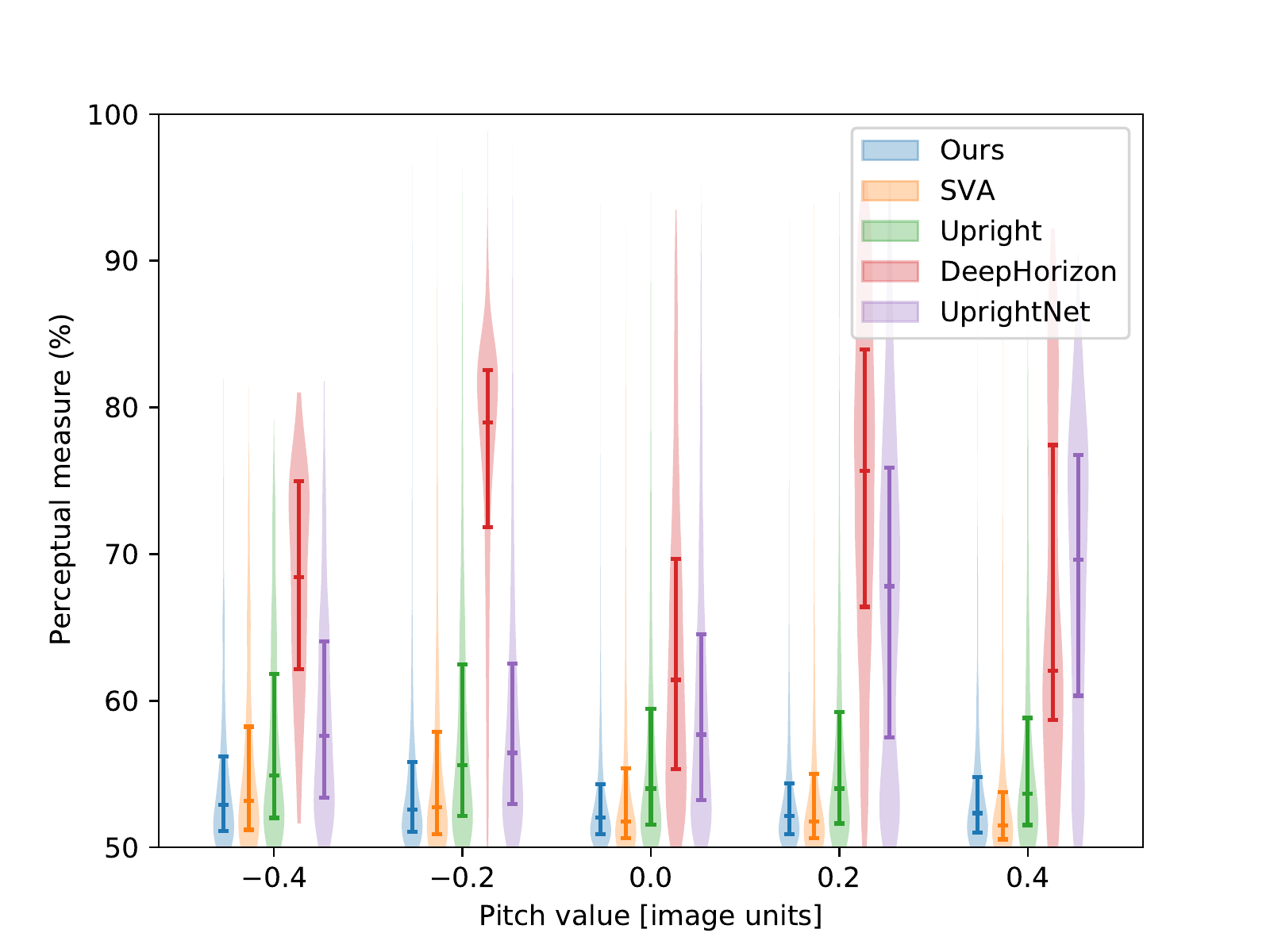} & 
    \includegraphics[trim=10 0 40 36,clip,width=.49\textwidth]{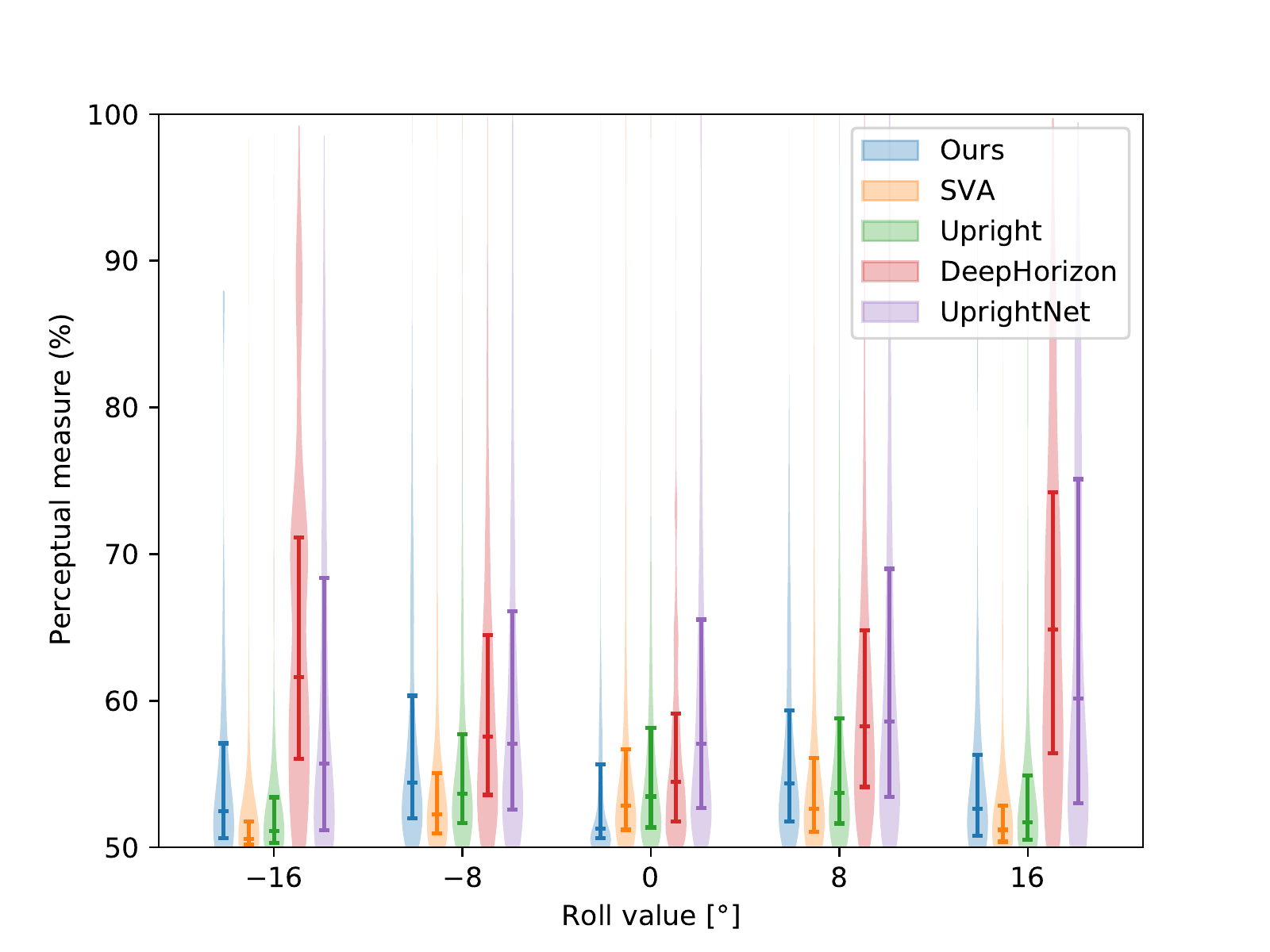} \vspace{-1em} \\
    \hspace{1.5em} \tiny (a) & \hspace{1em} \tiny (b) \\
    \includegraphics[trim=10 0 40 36,clip,width=.49\textwidth]{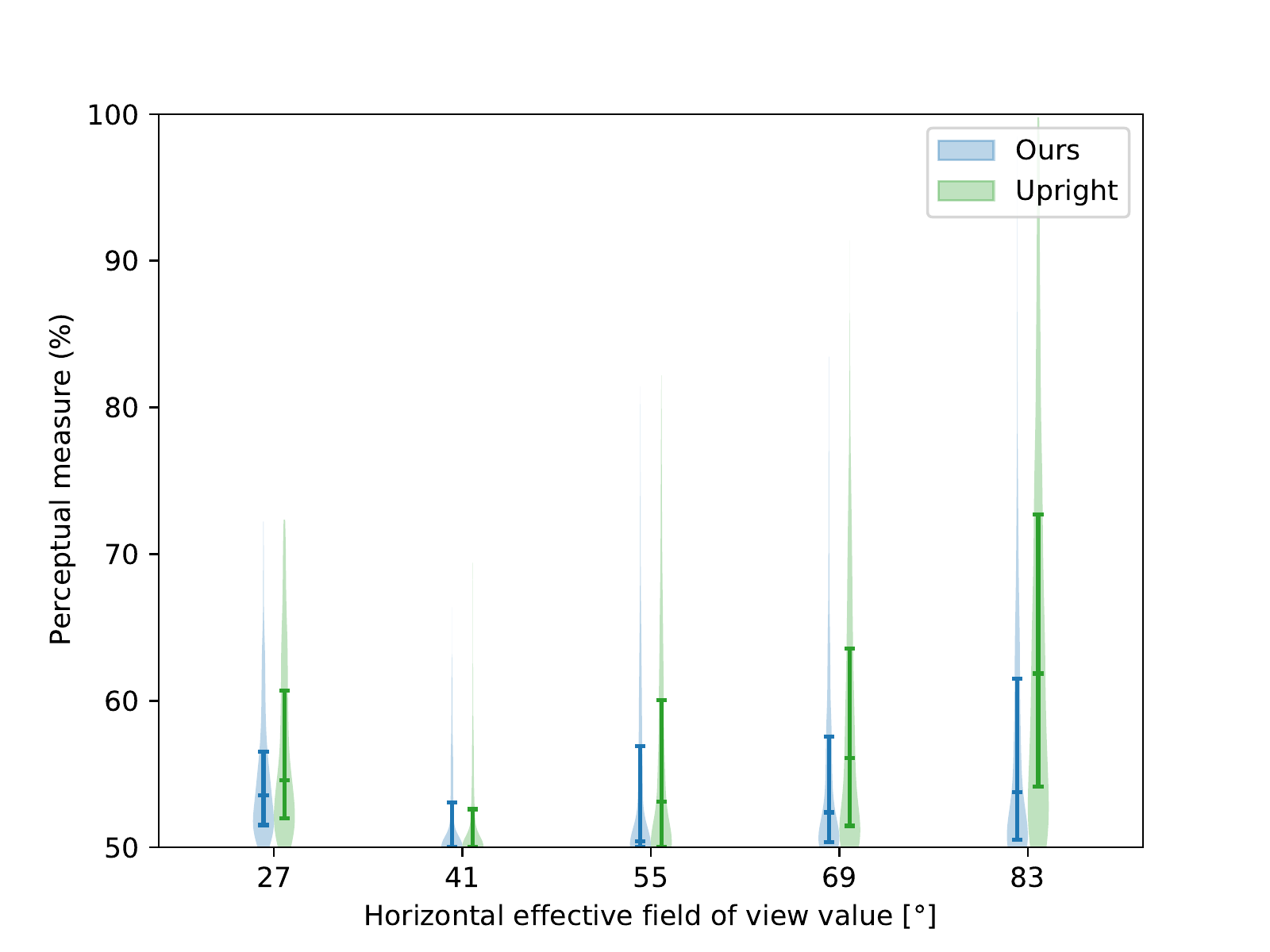} & 
    \includegraphics[trim=10 0 40 36,clip,width=.49\textwidth]{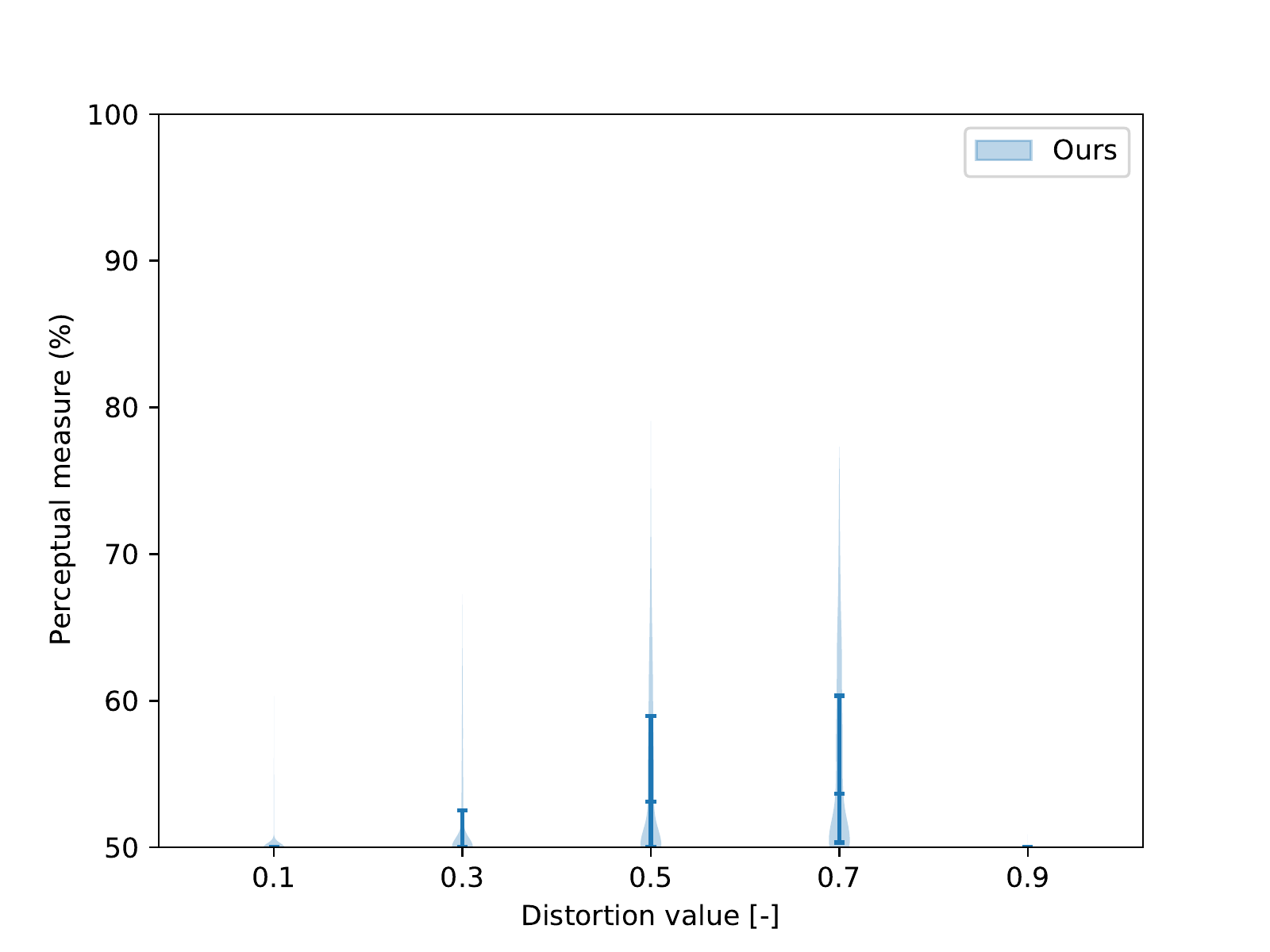} \vspace{-1em}\\
    \hspace{1.5em} \tiny (c) & \hspace{1em} \tiny (d) \\
\end{tabular} \vspace{-0.8em}
    \caption{Quantitative comparison of (a) roll, (b) pitch, (c) horizontal effective field of view, (d) distortion on our perceptual measure (lower is better, \change{50\% means the network can fool a human}). SVA~\cite{Lochman_minimal_2021_WACV} could not be evaluated on our perceptual measure for the last two parameters, because they use a different distortion model (division model) than ours (spherical model). Note that SVA fails on 49\% of the test images. 
    }
    \label{fig:perceptual_measure}
\end{figure}

We use the perception data gathered during these experiments to build a quantitative perceptual measure for evaluating and comparing approaches. For this, the data is binned in a $7\times7$ histogram on the 2D space of errors and absolute parameter (as in fig.~\ref{fig:pstudy_sensitivity_per_parameter}), and piece-wise linear functions are fitted on the bins. The resulting function indicates, for a particular set of camera parameters and errors on the estimation of these parameters, whether humans would perceive that error (value of \change{100\%}) or not (value of \change{50\%}). Fig.~\ref{fig:perceptual_measure} compares approaches in how humans would perceive their errors, and shows that our approach yields errors that are least noticeable amongst all approaches \change{according to a human perceptual measure}. 


\addtolength{\tabcolsep}{-5pt}
\addtolength{\tabcolsep}{5pt}

%% file: fig_backprop.tex
\begin{figure*}
    \setlength{\tabcolsep}{2pt}
    \centering
    \subfigure[]{
        \begin{tabular}{cc}
            \includegraphics[width=0.23\linewidth]{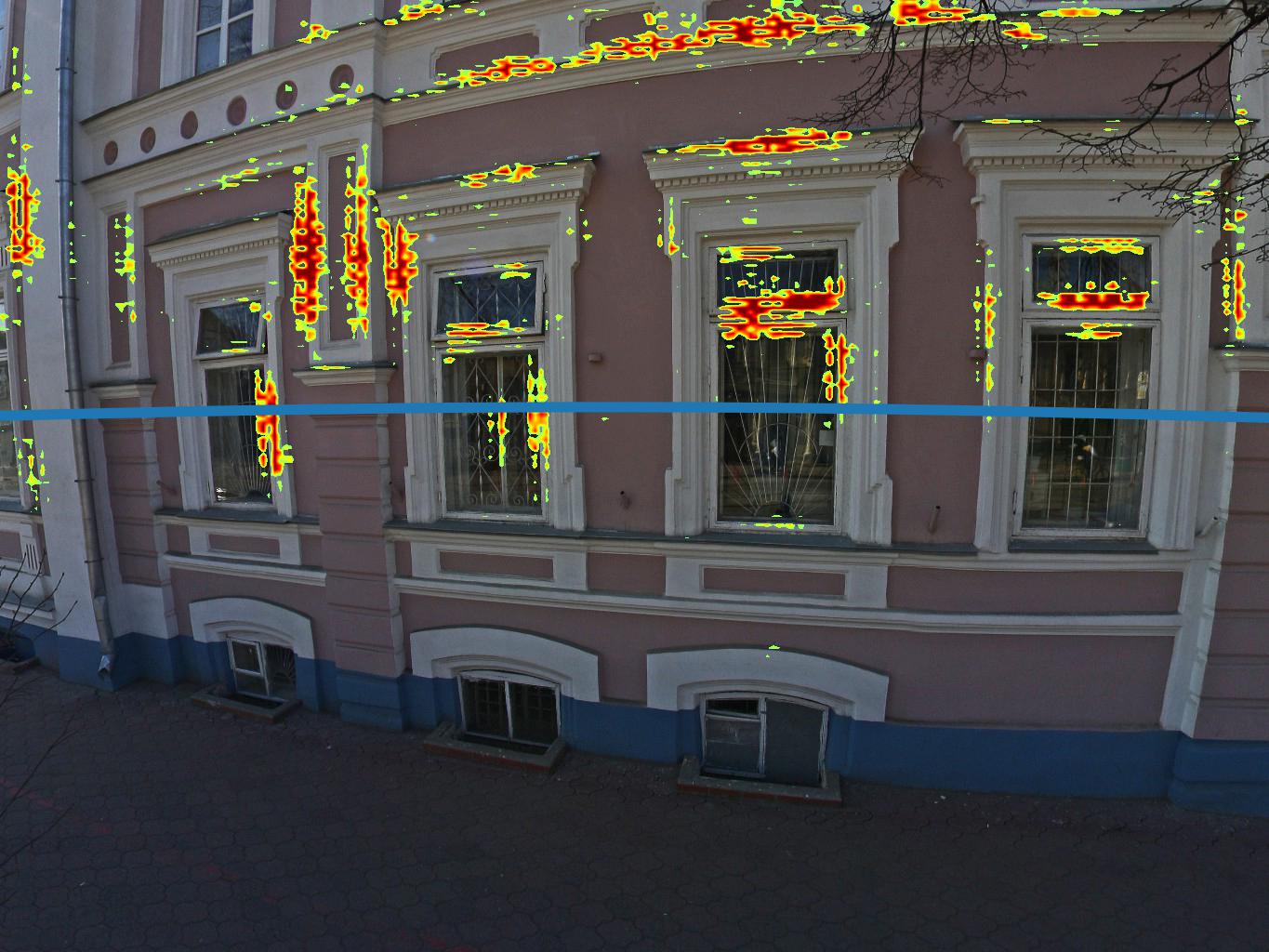} &  \includegraphics[width=0.23\linewidth]{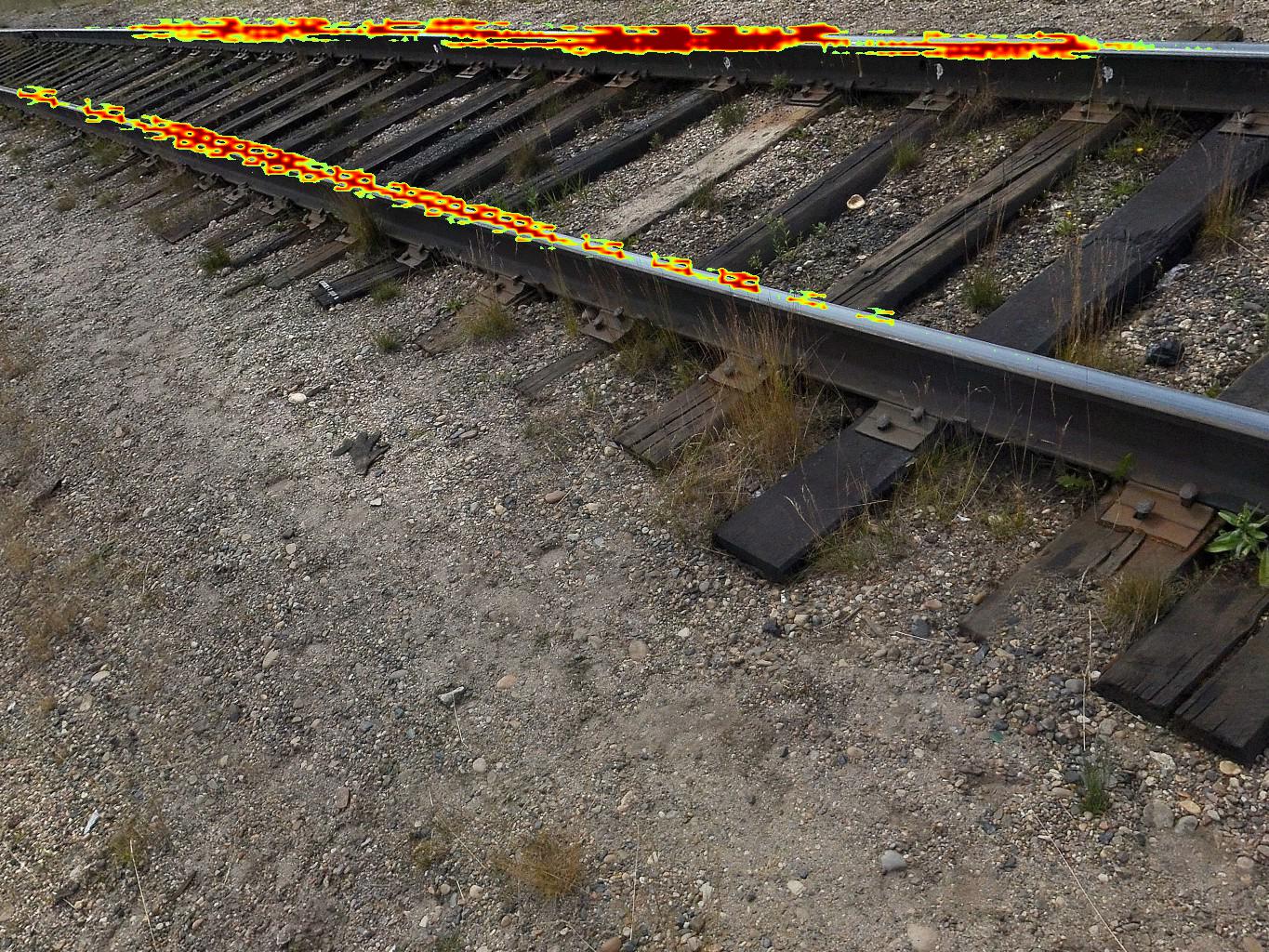}\\
            \includegraphics[width=0.23\linewidth]{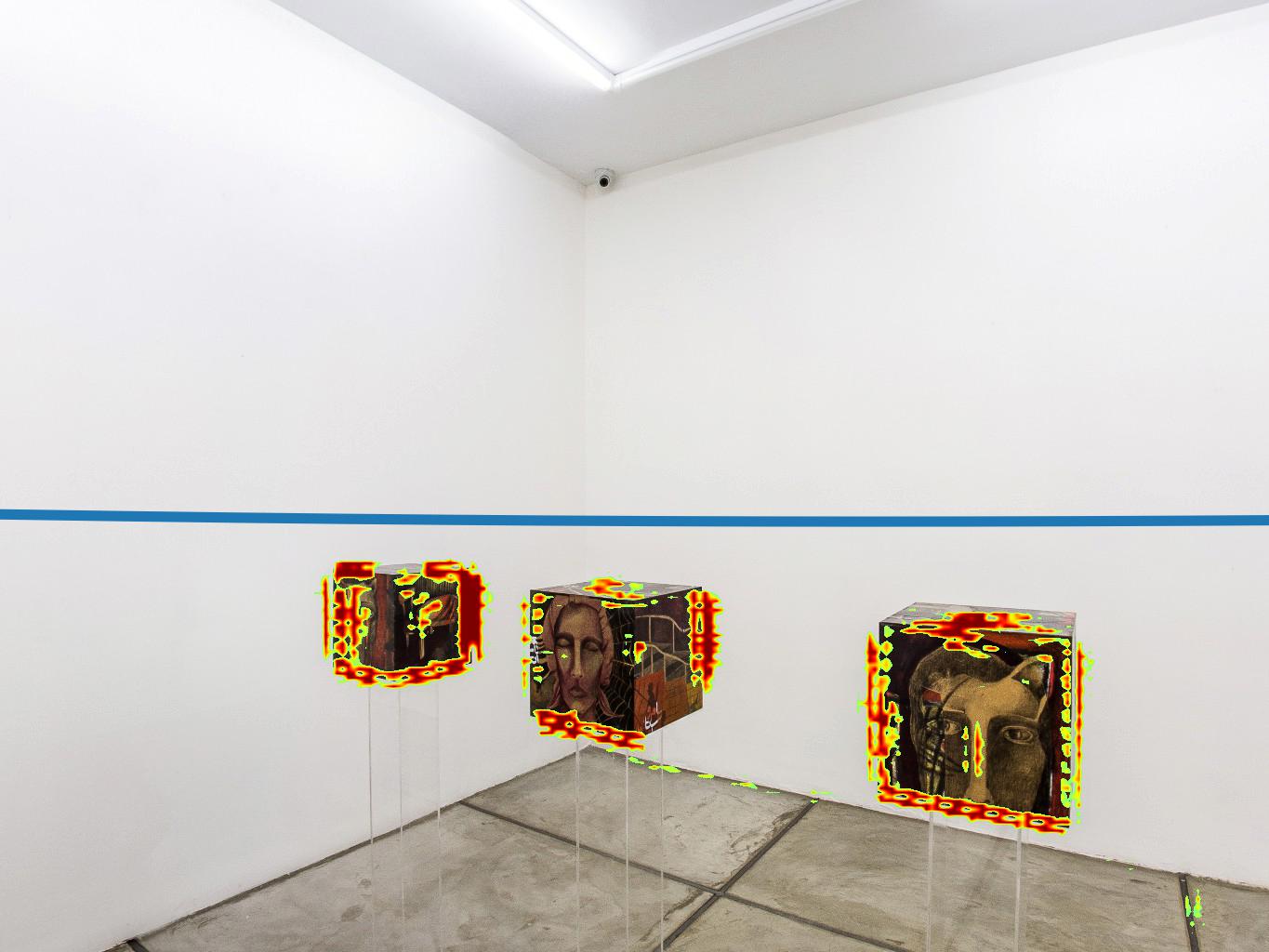} &
            \includegraphics[width=0.23\linewidth]{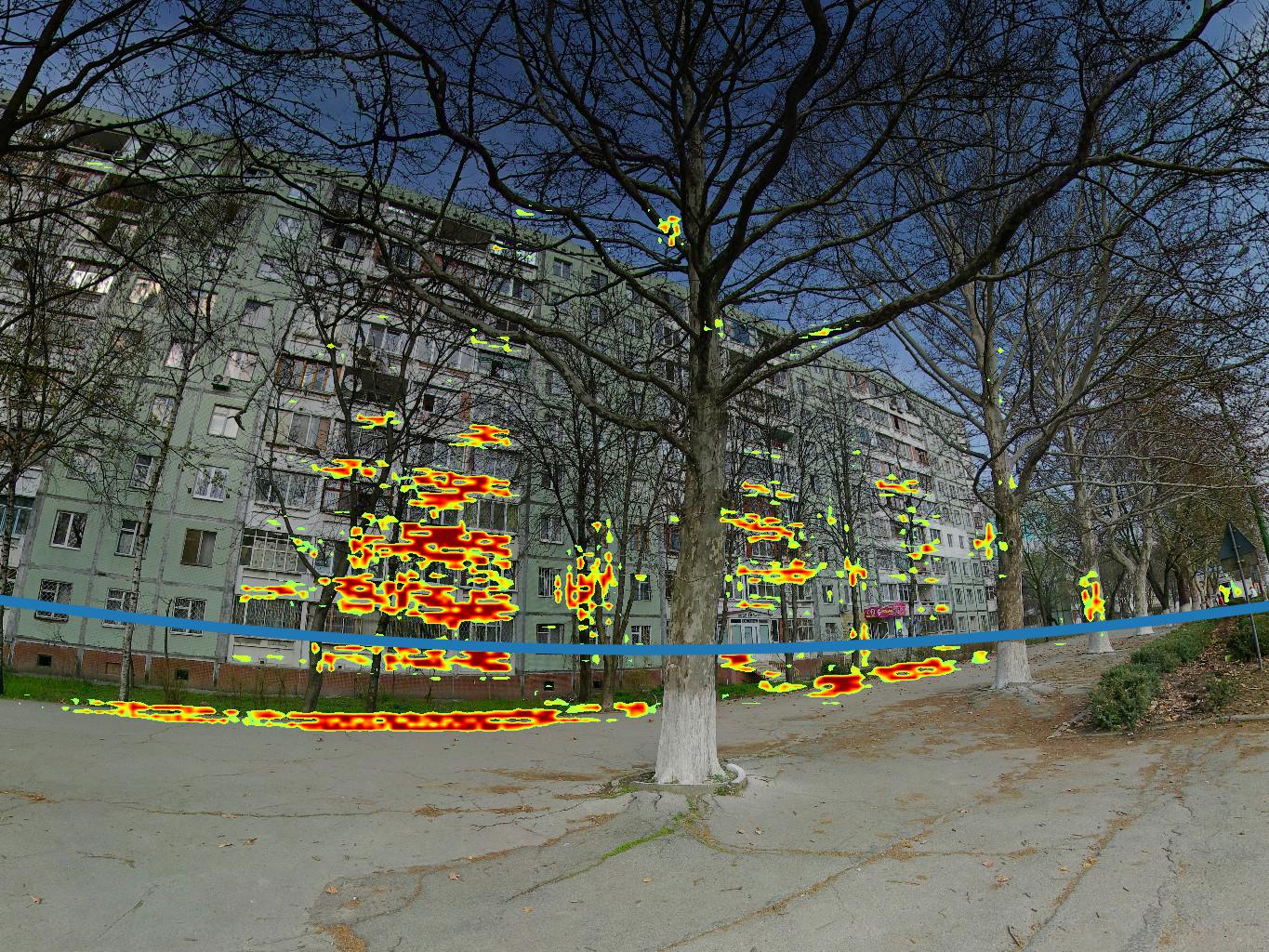}
        \end{tabular}
    }
    \subfigure[]{
        \begin{tabular}{cc}
            \includegraphics[width=0.23\linewidth]{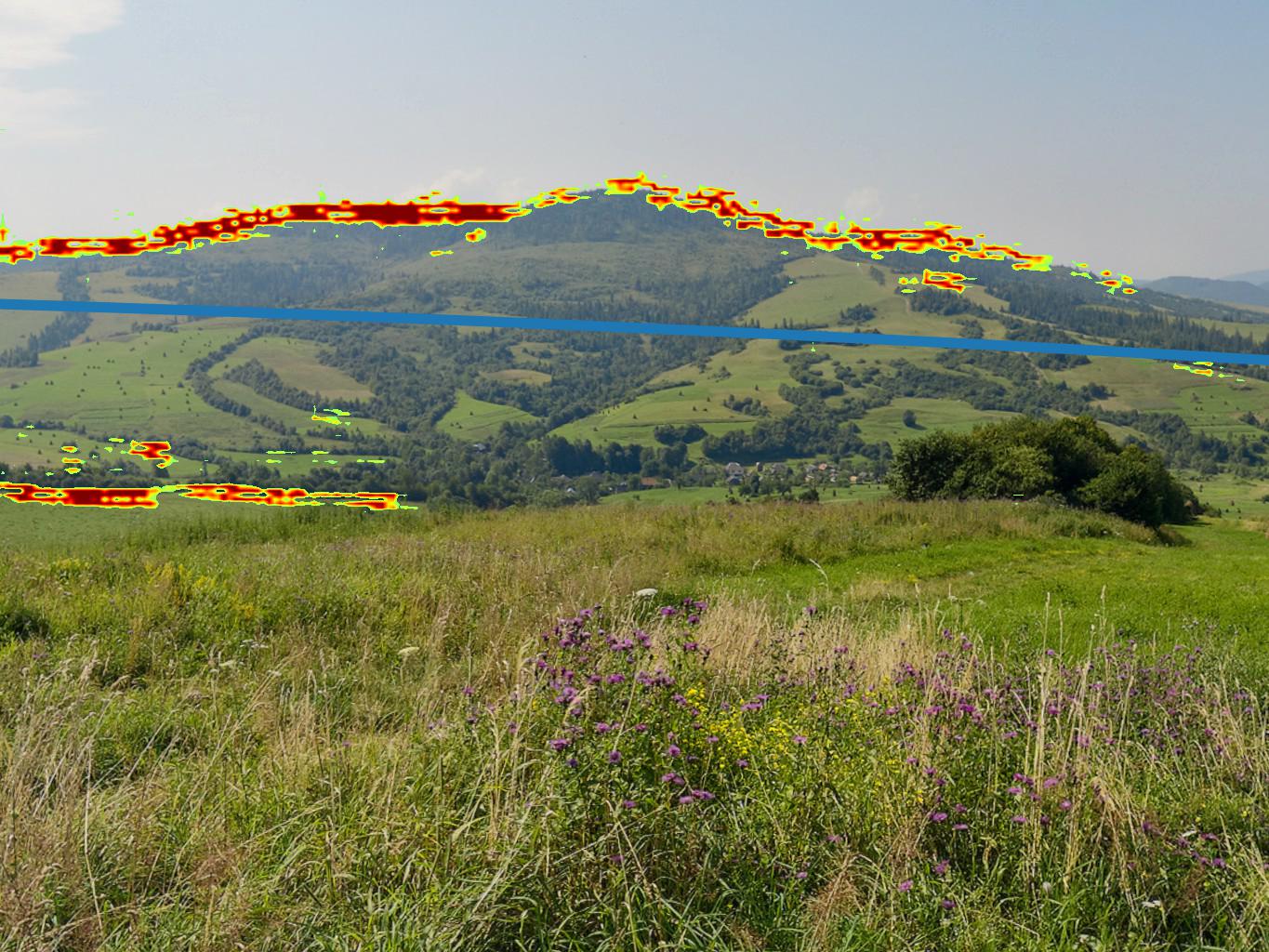} &  \includegraphics[width=0.23\linewidth]{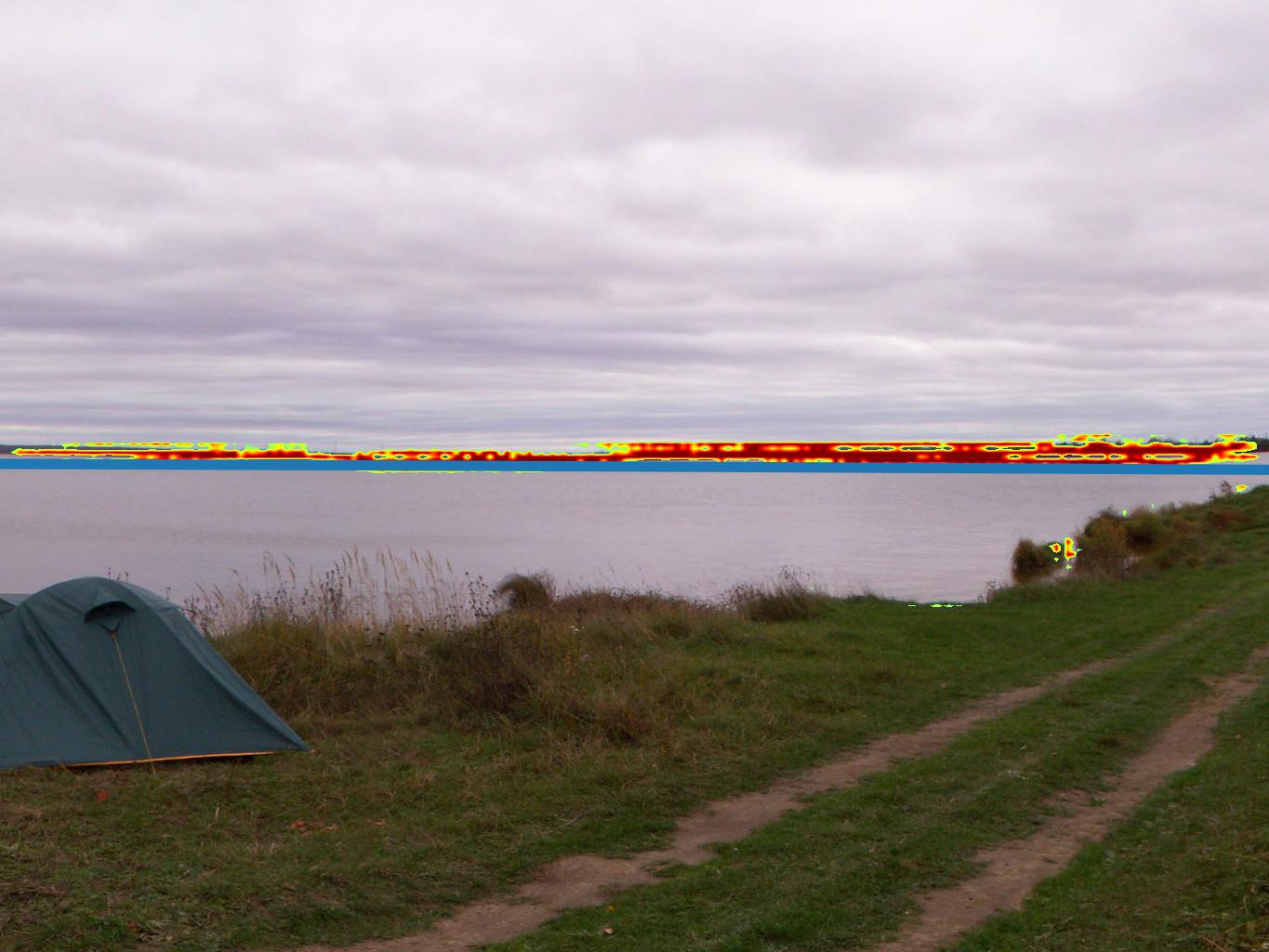}\\
            \includegraphics[width=0.23\linewidth]{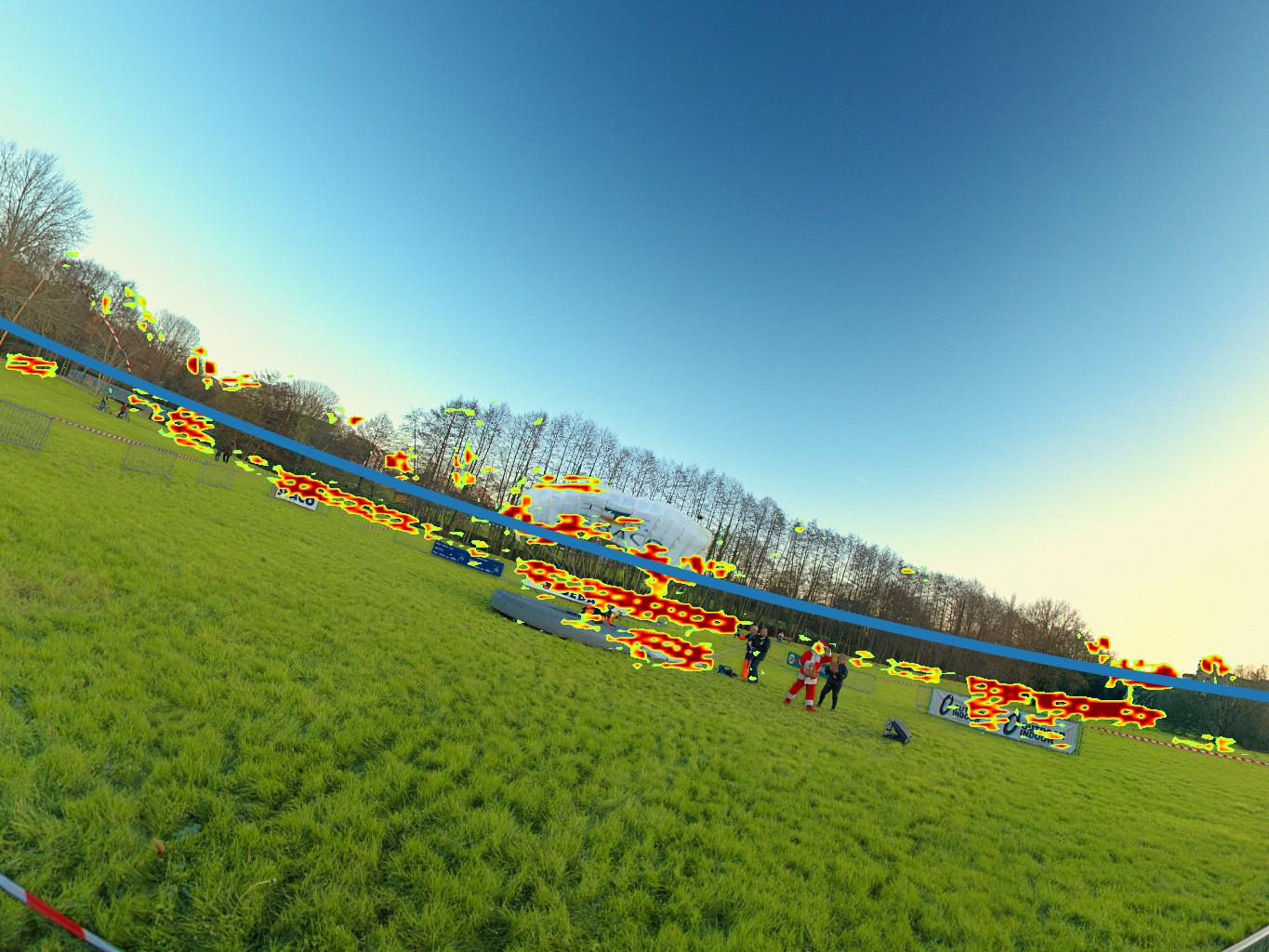} &
            \includegraphics[width=0.23\linewidth]{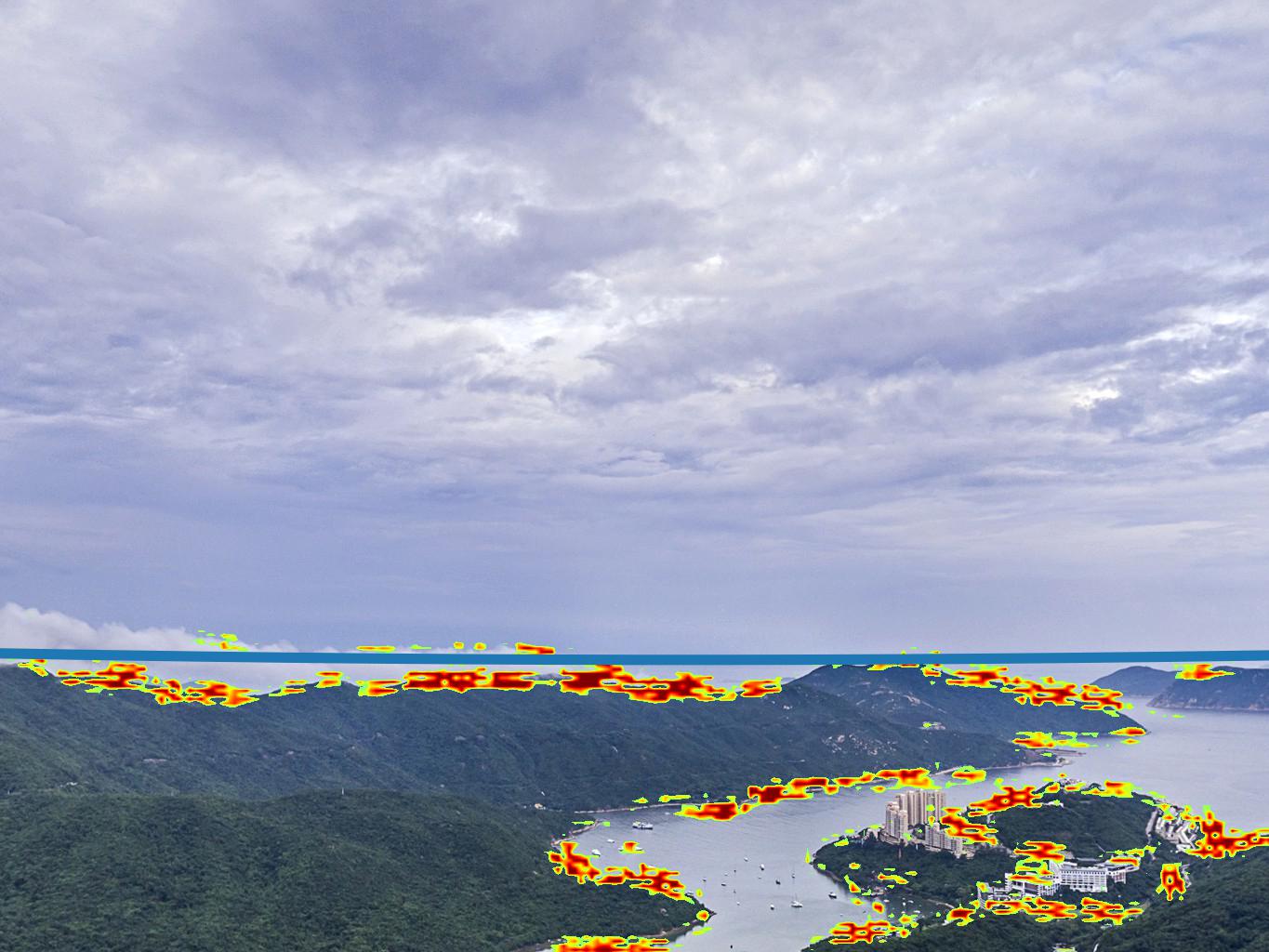}
        \end{tabular}
    }
    \caption{Analysis of the neural network focus. The result of smoothed guided backpropagation is displayed as a jet overlay, \change{and the estimated horizon line in blue}. (a) When present, edges corresponding to important vanishing lines are highlighted while other edges are discarded. (b) When no clear horizontal vanishing lines are detected, the neural network seems to look for the boundaries of either sky or land textures while dismissing the clouds or objects like trees, probably hinting bounds on horizon location in the image. \textbf{Refer to the supplementary material for more examples \change{and a comparison of the network focus before and after training}.}}
    \label{fig:nn_analysis_smoothed-guided-back-propagation}
\end{figure*}

%% file: fig_distortion_center.tex
\begin{figure}[t!]
\subfigure{
  \centering
  \includegraphics[width=0.3\linewidth]{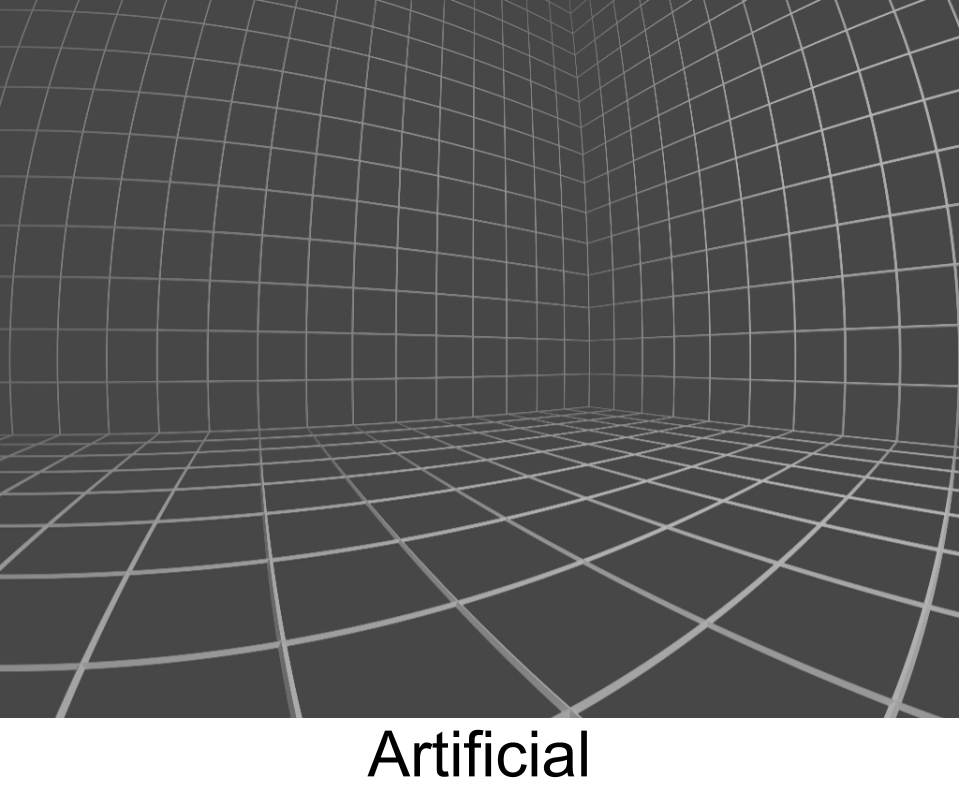}
}
\subfigure{
  \centering
  \includegraphics[width=0.3\linewidth]{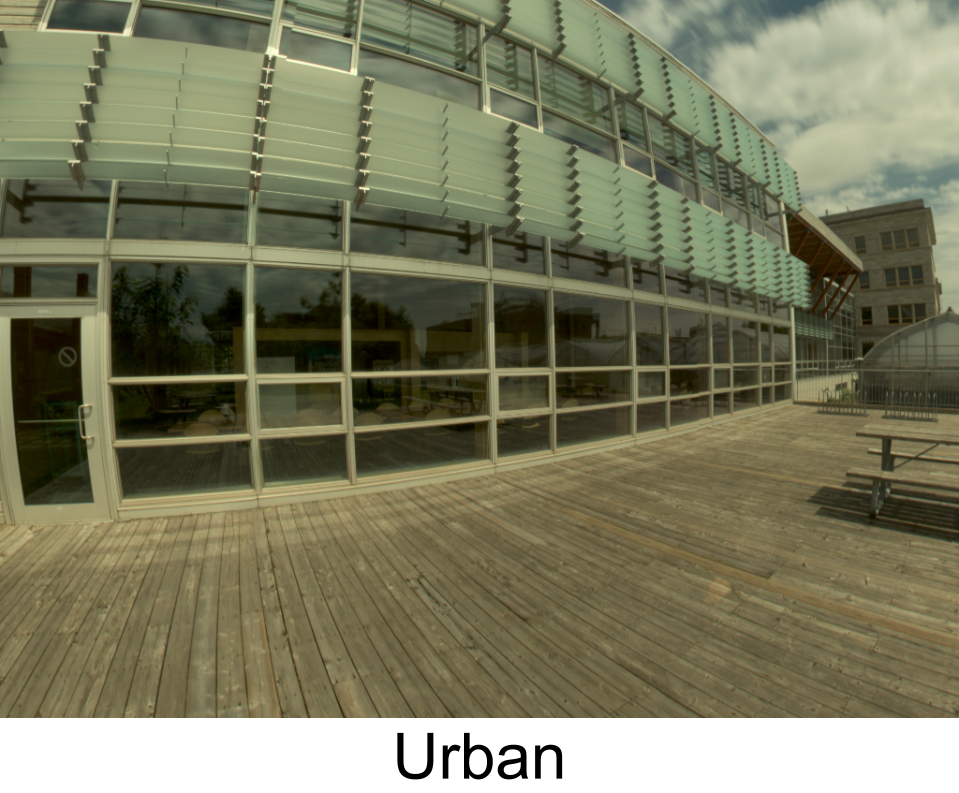}
}
\subfigure{
  \centering
  \includegraphics[width=0.3\linewidth]{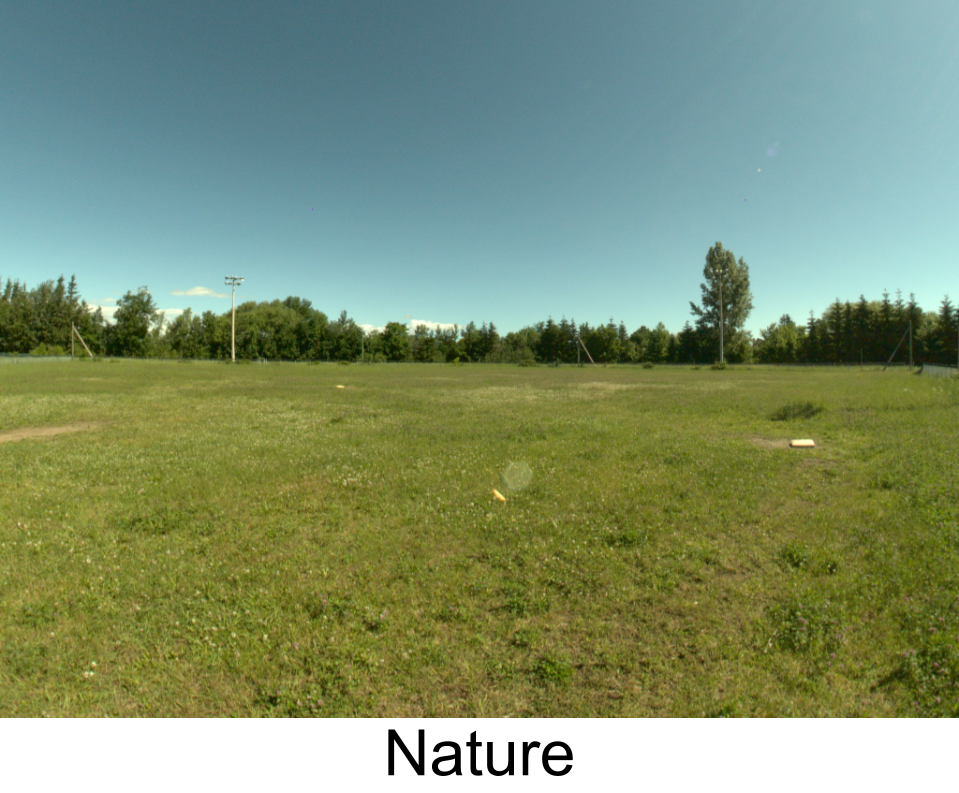}
}
\\
\vspace{-1.1em}
\subfigure{
   \centering
    \includegraphics[trim=12 5 15 21,clip,width=0.90\linewidth]{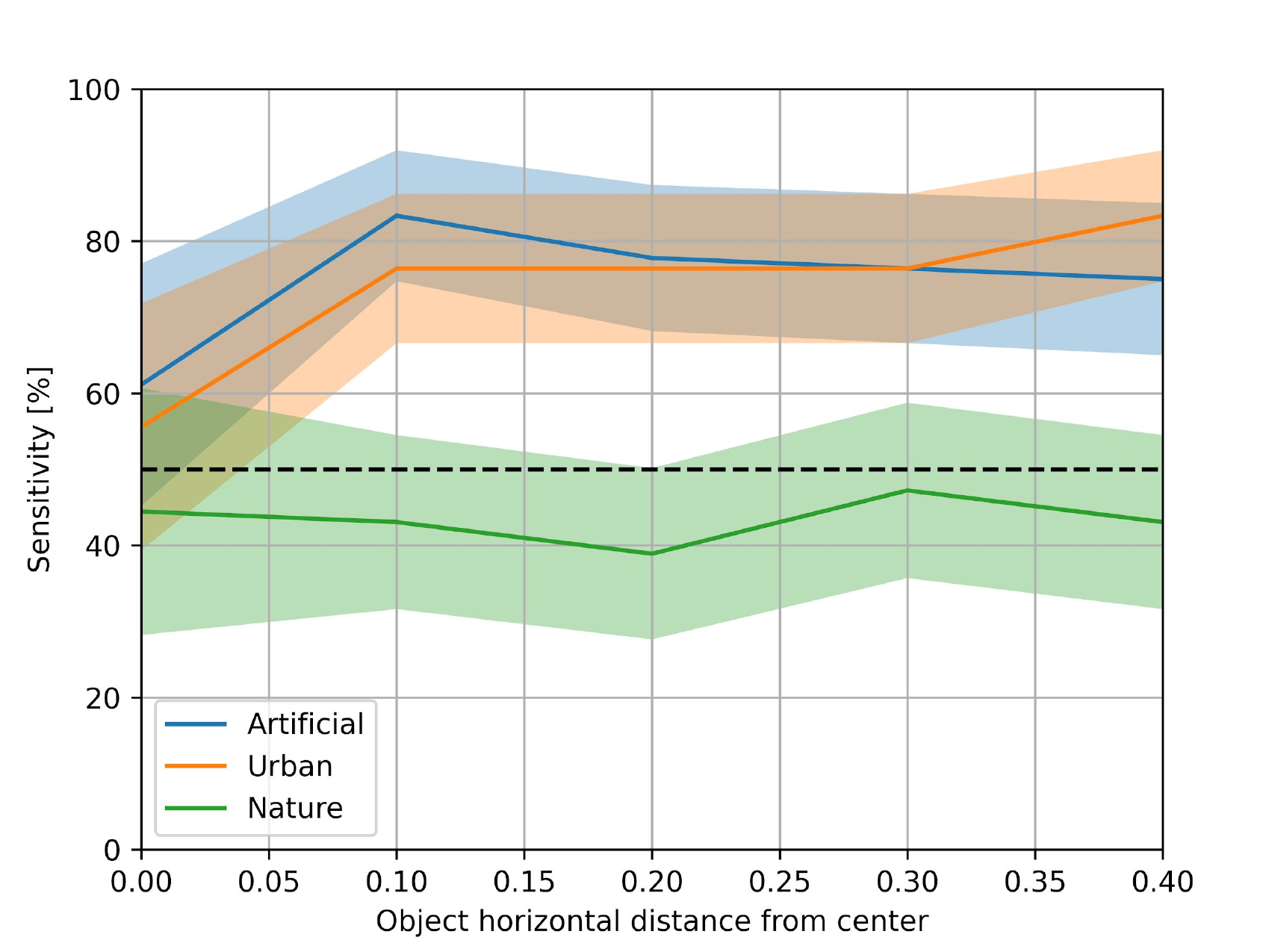}
}
\caption{\change{Results of Preliminary Experiment 2a: sensitivity of observers} to lens distortion as a function of \change{horizontal position of the virtual object}, for three scene types: artificial, urban and nature. Sensitivity is quantified as the percentage of users choosing the ground truth image, i.e., where the object has the same amount of distortion as the background. A sensitivity of 50\% means that humans are not sensitive to errors in distortion (confusion), whereas a sensitivity of 100\% means that humans are very sensitive to errors. The shaded areas correspond to the 95\% confidence intervals.}
\label{fig:study-distortion-center}
\end{figure}

%% file: fig_pstudy_overall.tex
\begin{figure}[t]
\centering
\includegraphics[trim=12 12 24 12,clip,width=\linewidth]{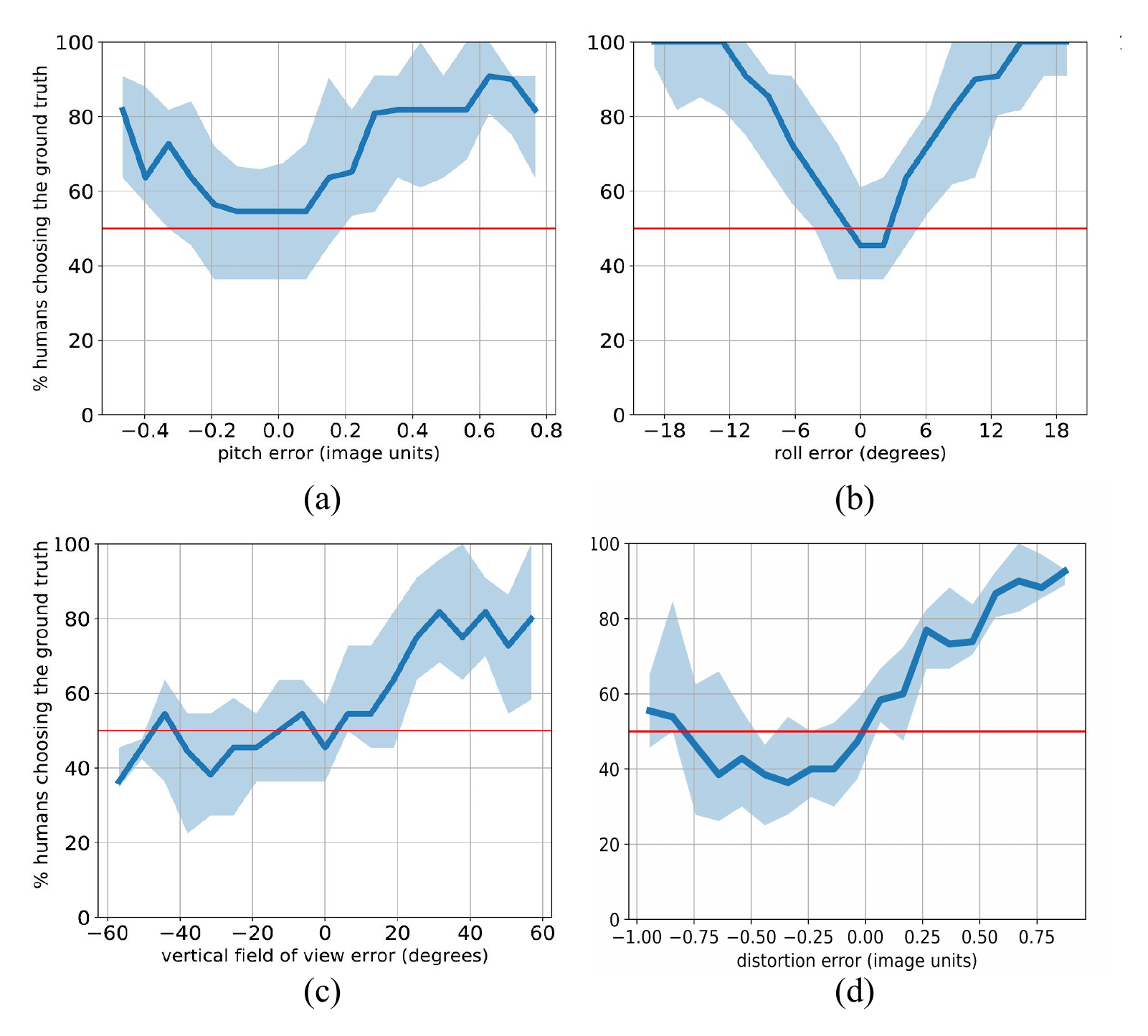}
\caption{Human sensitivity to errors in (a) pitch, (b) roll, (c) field of view and (d) distortion. We use the percentage of users choosing the image with an object inserted with the ground truth calibration as a measure of human sensitivity. 50\% represents perfect confusion: users are as likely to choose the image with a distorted calibration than the ground truth, and 100\% means that all humans could detect the ground truth image. The solid line is the median of the percentage of people who pick the ground truth for each image, and the light shaded regions \changeagain{delimits the region between} the first and third quartile. \changeagain{The mean interquartile ranges, corresponding to the average height of the light shaded regions, are (a) 30.8\% (b) 23.2\% (c) 26.0\% and (d) 20.8\%.}}
\label{fig:pstudy_overall_sensitivity}
\end{figure}

%% file: fig_pstudy_perparameter.tex
\begin{figure*}[!t]
\centering
\includegraphics[width=\linewidth]{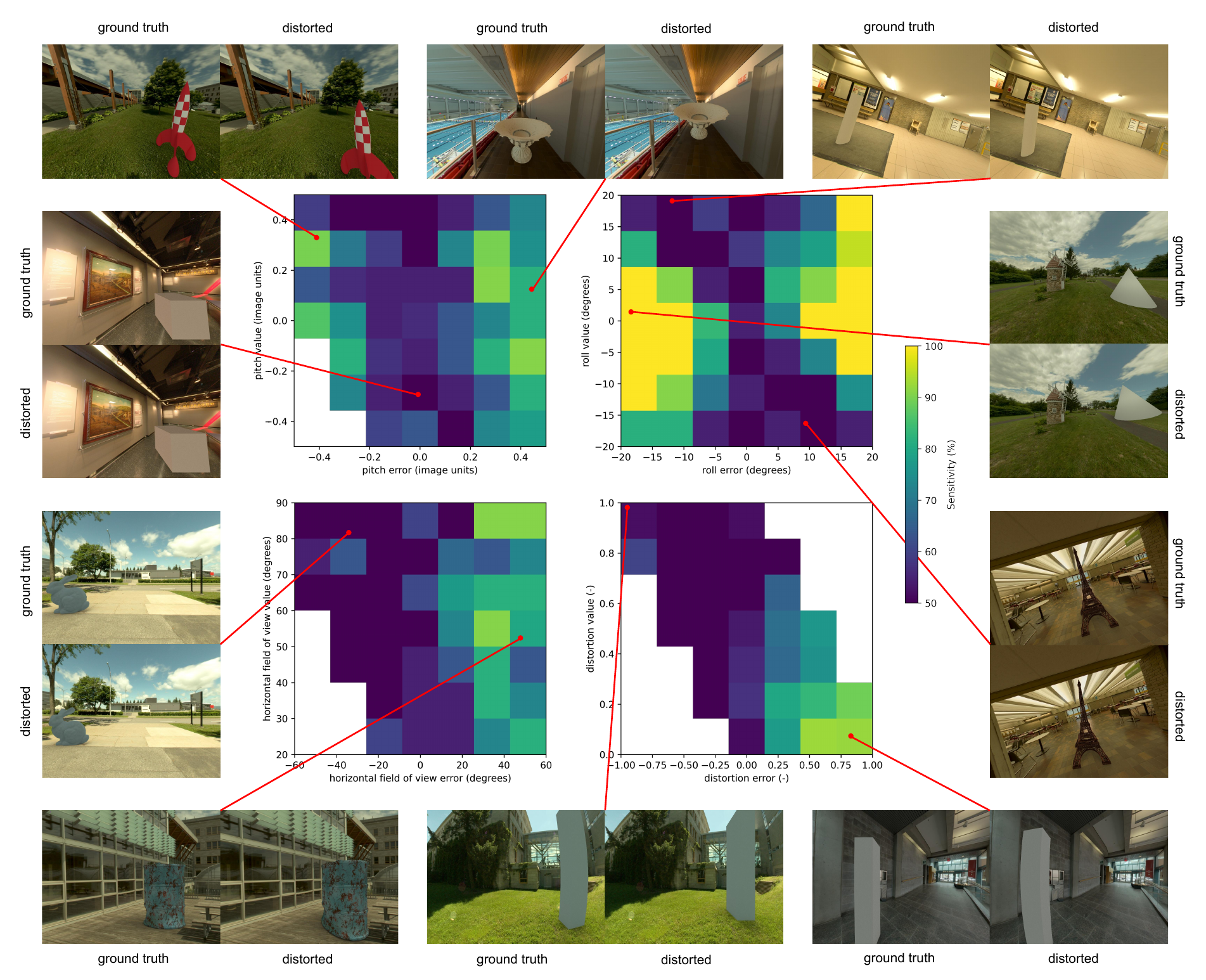}
\caption{Human sensitivity to errors in field of view (top left), roll (top right), pitch (bottom left) and distortion (bottom right) as a function of individual parameter values, along with examples of image pairs shown to the users. We bin the percentage of people choosing the ground truth per image of the user study. The colors in the plot represents the median over all values in each bin. Note the strong relation between the roll value and its human sensitivity to error. Some combinations of parameters and errors makes it impossible to perform insertion leading to missing values in the figure (e.g., the ground is not visible anymore in the image (bottom left in pitch), the resulting field of view would be negative (bottom left in field of view) or the resulting distortion would be negative (bottom left and top right in distortion). \textbf{See the supplementary material for more analysis}, including joint modeling of distortions on multiple parameters.}
\label{fig:pstudy_sensitivity_per_parameter}
\end{figure*}

%% file: sec_conclusion.tex
\section{\change{Discussion}}

\myparagraph{Limitations}
Despite providing competitive accuracy, our method bears some limitations. 
Notably, the unified spherical model has an ambiguity between the focal length and distortion parameter~\cite{hartley2007parameter, li2005non, cornelis2002lens}, leading to multiple parameters yielding the same (or very similar) projection of a world point in the image. A promising extension of our approach is to remove this ambiguity in a more compact representation. 
\change{In addition, images with extreme roll (greater than $60^\circ$) or pitch (looking straight up/down) angles are not properly estimated due to the way we perform our data sampling. If the image is cropped, the principal point is no longer at the center of the image, which breaks the radial symmetry and leads to errors in distortion estimation. Finally, images devoid of textures or recognizable scene elements (e.g. looking at a flat wall) do not possess sufficient cues for accurate camera calibration. Please see the supp. mat. for visual examples.
Lastly, we experimented with our perceptual measure to train our network. Unfortunately, doing so did not improve performance improve over regular training. We hypothesize that since the optimum is at the same location for both perceptual and regular losses---i.e., when the errors are 0---the neural network is optimized to a similar minima. Furthermore, the DenseNet architecture used as backend is not specifically tailored for the camera calibration task. We are hopeful that injecting domain knowledge into the neural network architecture will prove to be a fruitful direction for research in the near future.}

\myparagraph{Conclusion}
In this paper, we present a method to automatically estimate camera parameters from a single image. Our method is the first to jointly estimate camera pitch, roll, field of view, and lens distortion---and can be applied to both narrow and wide-angle fields of view. We then establish a link between the performance of our approach and human sensitivity to camera calibration errors by using virtual object insertion as a test bed and by running \change{two} large scale \change{perceptual} experiments. \change{In this sense, our work aligns with the recent relationships drawn between human perception and deep learning on 2D images  \cite{zhang2018unreasonable,johnson2016perceptual}, and extend those ideas to 3D and the camera itself.}

\change{In our experiments,} we find that humans are sensitive to calibration errors in a non-uniform way: there are configurations for which small errors are perceptible and others where large errors go unnoticed. Notably, we found that humans are particularly sensitive to errors in roll only when the image is level with the ground. When in doubt, humans tend to prefer images with lower perspective distortion. The human perception data can be aggregated into a quantitative perceptual measure which shows that our approach also yields the best results according to human perception. 

%% file: sec_bios.tex


\begin{IEEEbiography}[{\vspace{-0.6cm}\includegraphics[width=1in,height=1.15in,clip,keepaspectratio]{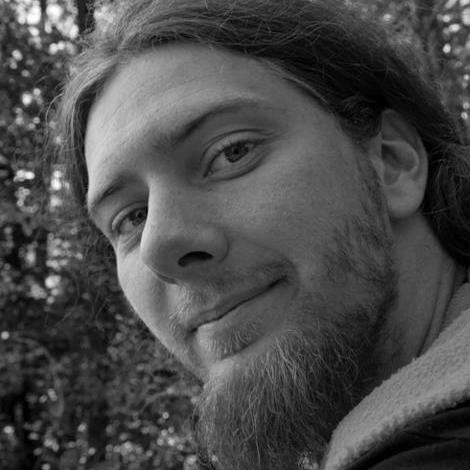}}]{Yannick Hold-Geoffroy}
is a Senior Research Scientist at Adobe in San Jos\'{e}. He received a Ph.D. degree in Electrical Engineering with a mention on the dean's honor roll from Universit\'{e} Laval, Canada, in 2018, for which he was awarded the CIPPRS Doctoral Dissertation Award in 2019. 
His research interests lie in the understanding of natural images through learned priors, aiming to help artists and engineers alike. 
\vspace{-1.5cm}
\end{IEEEbiography}

\begin{IEEEbiography}[{\vspace{-0.6cm}\includegraphics[width=1in,height=1.15in,clip,keepaspectratio]{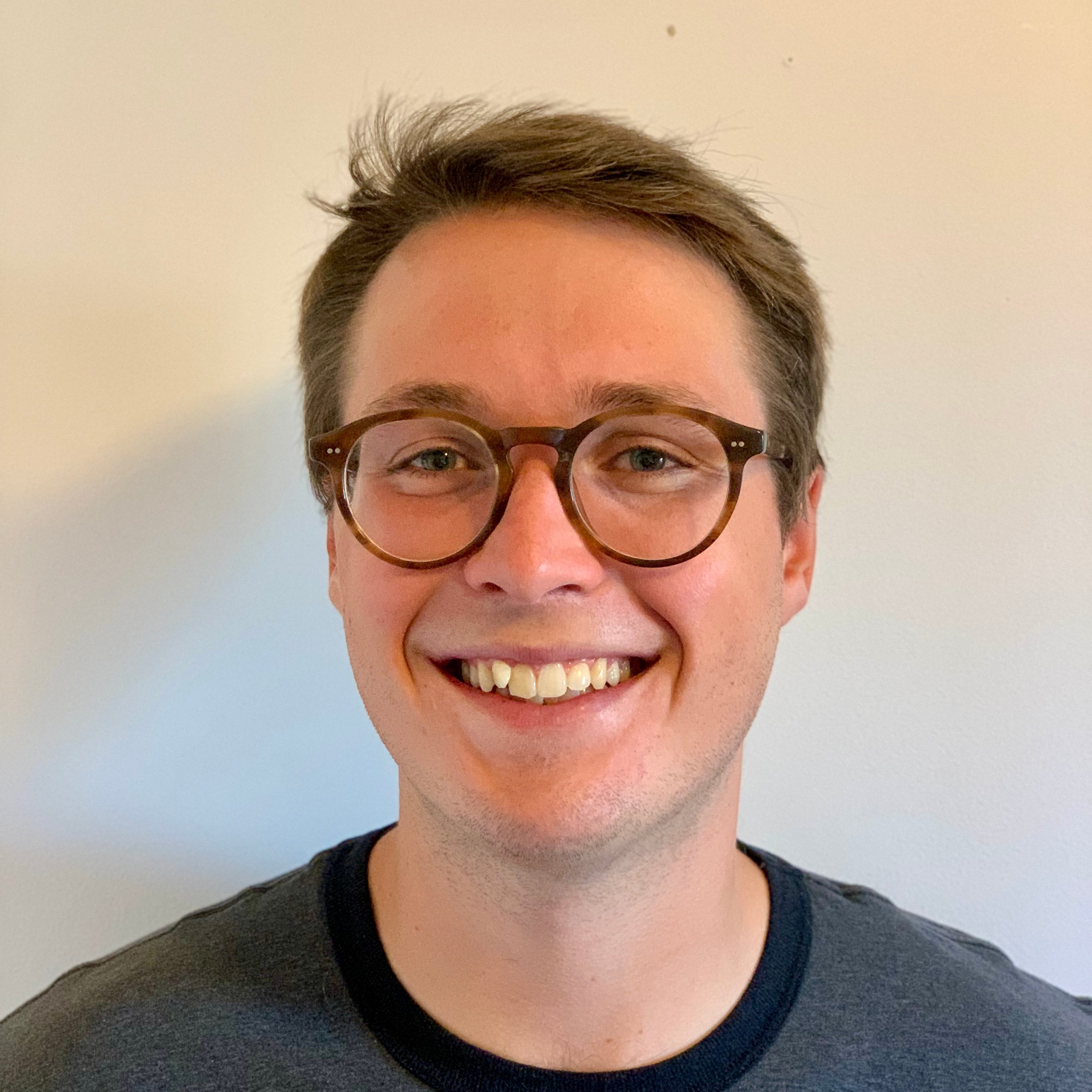}}]{Dominique Pich\'{e}-Meunier}
has completed his undergraduate degree in Engineering Physics at Universit\'{e} Laval in May of 2021. He is pursuing his M.S. in Electrical Engineering at the same institution since the Fall of 2021, working on camera parameters estimation from images. 
\vspace{-2cm}
\end{IEEEbiography}

\begin{IEEEbiography}[{\vspace{-0.6cm}\includegraphics[width=1in,height=1.15in,clip,keepaspectratio]{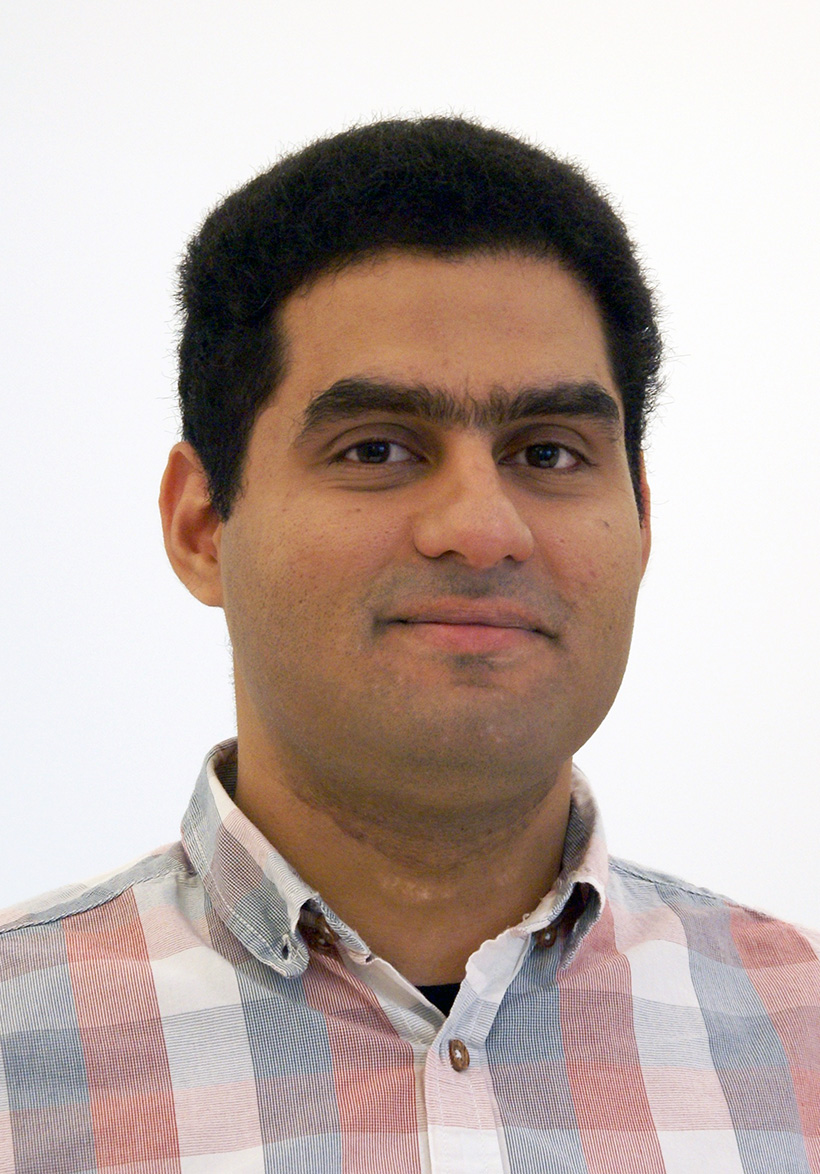}}]{Kalyan Sunkavalli}
is a Principal Scientist at Adobe Research. He received his Ph.D. from Harvard University in 2012, and his Masters in 2006 from Columbia University. He graduated from the Indian Institute of Technology, Delhi in 2003. His research interests lie at the intersection of computer vision and graphics; in particular, his work focuses on understanding visual appearance in images and videos. 
\vspace{-1.5cm}
\end{IEEEbiography}


\begin{IEEEbiography}[{\vspace{-0.2cm}\includegraphics[width=1in,height=1.15in]{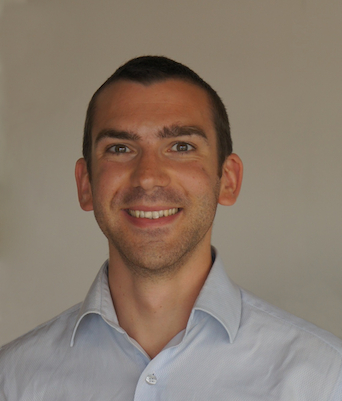}}]{Jean-Charles Bazin}
is now at Meta. He received his MS from the Universit\'{e} de Technologie de Compi\`{e}gne, France, in 2006, and his PhD from KAIST in 2011. He worked as Associate Research Scientist at Disney Research Zurich, Switzerland (2014-2016) and was a postdoc at ETH Zurich, Switzerland (2011-2014). He also worked as Postdoctoral Fellow at the Computer Vision Lab of Prof. Katsushi Ikeuchi, University of Tokyo, Japan (2010-2011). 
\vspace{-1.5cm}
\end{IEEEbiography}

\begin{IEEEbiography}[{\vspace{-0.6cm}\includegraphics[width=1in,height=1.15in,clip,keepaspectratio]{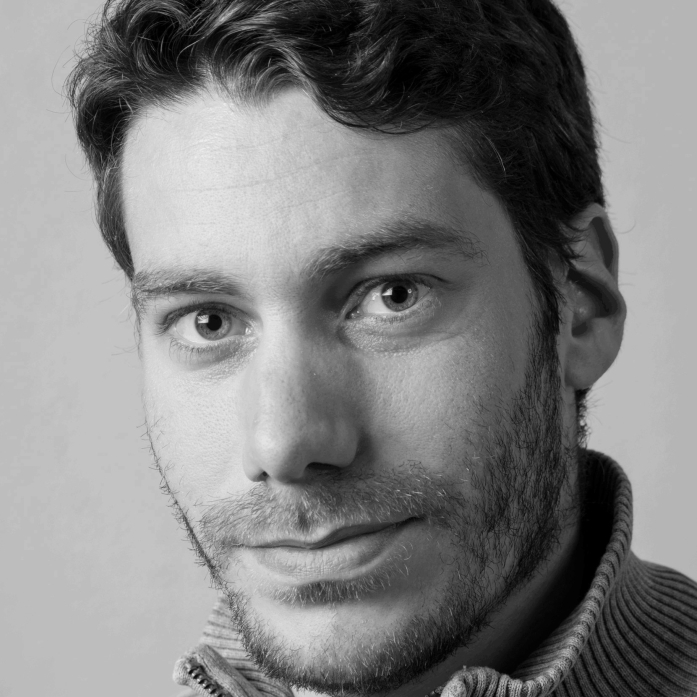}}]{Fran\c{c}ois Rameau}
is an Assistant Professor at the State University of New York (SUNY) in Korea.
He received his Ph.D. in Vision and Robotics from the University of Burgundy (France) in 2014. Later, he joined the Korea Advanced Institute of Science and Technology (KAIST, South Korea), first as a postdoctoral researcher and then as a Research Professor until 2023.
Dr. Rameau's research interests lie in 3D Computer Vision, Machine Learning, and Collaborative Robotics.
\vspace{-0.5cm}
\end{IEEEbiography}

\begin{IEEEbiography}[{\vspace{-0.6cm}\includegraphics[width=1in,height=1.15in,clip,keepaspectratio]{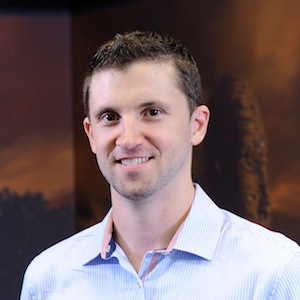}}]{Jean-Fran\c{c}ois Lalonde}
is a Professor at Université Laval. Previously, he was a Post-Doctoral Associate at Disney Research, Pittsburgh. He received a Ph.D. in Robotics from Carnegie Mellon University in 2011. His research interests focus computer vision, computer graphics, and machine learning to recreate the richness of our visual world.  
\vspace{-0.5cm}
\end{IEEEbiography}






%% file: main.bbl
\begin{thebibliography}{10}
\providecommand{\url}[1]{#1}
\csname url@samestyle\endcsname
\providecommand{\newblock}{\relax}
\providecommand{\bibinfo}[2]{#2}
\providecommand{\BIBentrySTDinterwordspacing}{\spaceskip=0pt\relax}
\providecommand{\BIBentryALTinterwordstretchfactor}{4}
\providecommand{\BIBentryALTinterwordspacing}{\spaceskip=\fontdimen2\font plus
\BIBentryALTinterwordstretchfactor\fontdimen3\font minus
  \fontdimen4\font\relax}
\providecommand{\BIBforeignlanguage}[2]{{%
\expandafter\ifx\csname l@#1\endcsname\relax
\typeout{** WARNING: IEEEtran.bst: No hyphenation pattern has been}%
\typeout{** loaded for the language `#1'. Using the pattern for}%
\typeout{** the default language instead.}%
\else
\language=\csname l@#1\endcsname
\fi
#2}}
\providecommand{\BIBdecl}{\relax}
\BIBdecl

\bibitem{Zhang:TPAMI:00}
Z.~Zhang, ``A flexible new technique for camera calibration,'' \emph{TPAMI},
  2000.

\bibitem{heikkila2000geometric}
J.~Heikkila, ``Geometric camera calibration using circular control points,''
  \emph{TPAMI}, vol.~22, no.~10, pp. 1066--1077, 2000.

\bibitem{Scaramuzza:IROS:06}
D.~Scaramuzza, A.~Martinelli, and R.~Siegwart, ``A toolbox for easily
  calibrating omnidirectional cameras,'' in \emph{IROS}, 2006.

\bibitem{boby2016single}
R.~A. Boby and S.~K. Saha, ``Single image based camera calibration and pose
  estimation of the end-effector of a robot,'' in \emph{ICRA}.\hskip 1em plus
  0.5em minus 0.4em\relax IEEE, 2016, pp. 2435--2440.

\bibitem{lalonde-siggraph-07}
J.-F. Lalonde, D.~Hoiem, A.~A. Efros, C.~Rother, J.~Winn, and A.~Criminisi,
  ``Photo clip art,'' \emph{SIGGRAPH}, vol.~26, no.~3, p.~3, 2007.

\bibitem{debevec-siggraph-98}
P.~Debevec, ``Rendering synthetic objects into real scenes : Bridging
  traditional and image-based graphics with global illumination and high
  dynamic range photography,'' in \emph{Proceedings of ACM SIGGRAPH}, 1998, pp.
  189--198.

\bibitem{Cavanagh2005}
P.~Cavanagh, ``The artist as neuroscientist,'' \emph{Nature}, vol. 434, 2005.

\bibitem{Farid2010}
H.~Farid and M.~J. Bravo, ``Image forensic analyses that elude the human visual
  system,'' in \emph{Media forensics and security II}, vol. 7541.\hskip 1em
  plus 0.5em minus 0.4em\relax International Society for Optics and Photonics,
  2010, p. 754106.

\bibitem{Bogdan:CVMP:2018}
O.~Bogdan, V.~Eckstein, F.~Rameau, and J.-C. Bazin, ``{DeepCalib}: a deep
  learning approach for automatic intrinsic calibration of wide field-of-view
  cameras,'' in \emph{CVMP}, 2018.

\bibitem{hold2017perceptual}
Y.~Hold-Geoffroy, K.~Sunkavalli, J.~Eisenmann, M.~Fisher, E.~Gambaretto,
  S.~Hadap, and J.-F. Lalonde, ``A perceptual measure for deep single image
  camera calibration,'' in \emph{CVPR}, 2018.

\bibitem{Criminisi2000}
A.~Criminisi, I.~Reid, and A.~Zisserman, ``Single view metrology,''
  \emph{IJCV}, vol.~40, no.~2, pp. 123--148, 2000.

\bibitem{Criminisi00}
A.~Criminisi and A.~Zisserman, ``Shape from texture: homogeneity revisited,''
  in \emph{BMVC}, 2000.

\bibitem{Fouhey2013}
D.~F. Fouhey, A.~Gupta, and M.~Hebert, ``Data-driven {3D} primitives for single
  image understanding,'' in \emph{ICCV}, 2013, pp. 3392--3399.

\bibitem{hedau-iccv-09}
V.~Hedau, D.~Hoiem, and D.~A. Forsyth, ``Recovering the spatial layout of
  cluttered rooms,'' in \emph{ICCV}, 2009, pp. 1849--1856.

\bibitem{izadinia-cvpr-17}
H.~Izadinia, Q.~Shan, and S.~M. Seitz, ``{IM2CAD},'' in \emph{CVPR}, 2017, pp.
  2422--2431.

\bibitem{hoiem-cvpr-06}
D.~Hoiem, A.~A. Efros, and M.~Hebert, ``Putting objects in perspective,'' in
  \emph{CVPR}, 2006, pp. 2137--2144.

\bibitem{Hartley2004}
R.~I. Hartley and A.~Zisserman, \emph{Multiple View Geometry in Computer
  Vision}, 2nd~ed.\hskip 1em plus 0.5em minus 0.4em\relax Cambridge University
  Press, 2004.

\bibitem{Sturm1999}
P.~Sturm and S.~Maybank, ``On plane-based camera calibration: {A} general
  algorithm, singularities, applications,'' in \emph{CVPR}, 1999, pp.
  1432--1437.

\bibitem{remondino2006digital}
F.~Remondino and C.~Fraser, ``Digital camera calibration methods:
  considerations and comparisons,'' \emph{International Archives of the
  Photogrammetry, Remote Sensing and Spatial Information Sciences}, vol.~36,
  no.~5, pp. 266--272, 2006.

\bibitem{Heikkila1997}
J.~Heikkil{\"{a}} and O.~Silv{\'{e}}n, ``A four-step camera calibration
  procedure with implicit image correction,'' in \emph{CVPR}, 1997, pp.
  1106--1112.

\bibitem{Chen2004}
Q.~Chen, H.~Wu, and T.~Wada, ``Camera calibration with two arbitrary coplanar
  circles,'' in \emph{ECCV}, 2004, pp. 521--532.

\bibitem{mei2007single}
C.~Mei and P.~Rives, ``Single view point omnidirectional camera calibration
  from planar grids,'' in \emph{ICRA}, 2007.

\bibitem{Gasparini:ICCV:09}
S.~Gasparini, P.~Sturm, and J.~P. Barreto, ``Plane-based calibration of central
  catadioptric cameras,'' in \emph{ICCV}, 2009.

\bibitem{Shah:ICRA:94}
S.~Shah and J.~K. Aggarwal, ``A simple calibration procedure for fish-eye
  (high-distortion) lens camera,'' in \emph{ICRA}, 1994.

\bibitem{Ying:IJCV:08}
X.~Ying and H.~Zha, ``Identical projective geometric properties of central
  catadioptric line images and sphere images with applications to
  calibration,'' \emph{IJCV}, 2008.

\bibitem{larsson_revisiting_2019}
V.~Larsson, T.~Sattler, Z.~Kukelova, and M.~Pollefeys, ``Revisiting {Radial}
  {Distortion} {Absolute} {Pose},'' in \emph{ICCV}, Oct. 2019, pp. 1062--1071.

\bibitem{zhang2015line}
M.~Zhang, J.~Yao, M.~Xia, K.~Li, Y.~Zhang, and Y.~Liu, ``Line-based multi-label
  energy optimization for fisheye image rectification and calibration,'' in
  \emph{CVPR}, 2015.

\bibitem{Workman2016}
S.~Workman, M.~Zhai, and N.~Jacobs, ``Horizon lines in the wild,'' in
  \emph{BMVC}, 2016.

\bibitem{Barreto:TPAMI:05}
J.~P. Barreto and H.~Ara{\'{u}}jo, ``Geometric properties of central
  catadioptric line images and their application in calibration,''
  \emph{TPAMI}, 2005.

\bibitem{Lee2014}
H.~Lee, E.~Shechtman, J.~Wang, and S.~Lee, ``Automatic upright adjustment of
  photographs with robust camera calibration,'' \emph{TPAMI}, vol.~36, no.~5,
  pp. 833--844, 2014.

\bibitem{hughes2010equidistant}
C.~Hughes, P.~Denny, M.~Glavin, and E.~Jones, ``{Equidistant Fish-Eye
  Calibration and Rectification by Vanishing Point Extraction},'' \emph{TPAMI},
  2010.

\bibitem{Antunes:CVPR:2017}
M.~Antunes, J.~P. Barreto, D.~Aouada, and B.~Ottersten, ``Unsupervised
  vanishing point detection and camera calibration from a single manhattan
  image with radial distortion,'' in \emph{CVPR}, 2017.

\bibitem{xian2019uprightnet}
W.~Xian, Z.~Li, M.~Fisher, J.~Eisenmann, E.~Shechtman, and N.~Snavely,
  ``Uprightnet: Geometry-aware camera orientation estimation from single
  images,'' in \emph{ICCV}, 2019, pp. 9974--9983.

\bibitem{Rother:BMVC:2000}
C.~Rother, ``A new approach for vanishing point detection in architectural
  environments,'' in \emph{BMVC}, 2000, pp. 1--10.

\bibitem{Melo:ICCV:13}
R.~Melo, M.~Antunes, J.~P. Barreto, G.~Falc{\~{a}}o, and N.~Gon{\c{c}}alves,
  ``Unsupervised intrinsic calibration from a single frame using a "plumb-line"
  approach,'' in \emph{ICCV}, 2013.

\bibitem{Lalonde2010}
J.-F. Lalonde, S.~G. Narasimhan, and A.~A. Efros, ``What do the sun and the sky
  tell us about the camera?'' \emph{IJCV}, vol.~88, no.~1, pp. 24--51, 2010.

\bibitem{Workman2014}
S.~Workman, R.~P. Mihail, and N.~Jacobs, ``A pot of gold: Rainbows as a
  calibration cue,'' in \emph{ECCV}, 2014.

\bibitem{mendoncca2002camera}
M.~Mendon{\c{c}}a, I.~N.~D. Silva, and J.~E. Castanho, ``Camera calibration
  using neural networks,'' \emph{WSCG}, 2002.

\bibitem{workman2015deepfocal}
S.~Workman, C.~Greenwell, M.~Zhai, R.~Baltenberger, and N.~Jacobs,
  ``{DeepFocal}: a method for direct focal length estimation,'' in \emph{ICIP},
  2015.

\bibitem{rong2016radial}
J.~Rong, S.~Huang, Z.~Shang, and X.~Ying, ``Radial lens distortion correction
  using convolutional neural networks trained with synthesized images,'' in
  \emph{ACCV}, 2016.

\bibitem{Jung:VR:upright:2019}
R.~Jung, A.~Lee, A.~Ashtari, and J.-C. Bazin, ``{Deep360Up:} deep
  learning-based approach for automatic {VR} image upright adjustment,'' in
  \emph{IEEE VR}, 2019.

\bibitem{jung2017robust}
J.~Jung, B.~Kim, J.-Y. Lee, B.~Kim, and S.~Lee, ``Robust upright adjustment of
  360 spherical panoramas,'' \emph{The Visual Computer}, vol.~33, no. 6-8, pp.
  737--747, 2017.

\bibitem{Lochman_minimal_2021_WACV}
Y.~Lochman, O.~Dobosevych, R.~Hryniv, and J.~Pritts, ``Minimal solvers for
  single-view lens-distorted camera auto-calibration,'' in \emph{WACV}, January
  2021, pp. 2887--2896.

\bibitem{wildenauer2013closed}
H.~Wildenauer and B.~Micusik, ``Closed form solution for radial distortion
  estimation from a single vanishing point,'' in \emph{BMVC}, vol.~1, 2013,
  p.~2.

\bibitem{pritts2018radially}
J.~Pritts, Z.~Kukelova, V.~Larsson, and O.~Chum, ``Radially-distorted conjugate
  translations,'' in \emph{Proceedings of the IEEE Conference on Computer
  Vision and Pattern Recognition}, 2018, pp. 1993--2001.

\bibitem{Lopez:CVPR:2019}
M.~L\'opez-Antequera, R.~Mar\'i, P.~Gargallo, Y.~Kuang, J.~Gonzalez-Jimenez,
  and G.~Haro, ``Deep single image camera calibration with radial distortion,''
  in \emph{CVPR}, 2019.

\bibitem{barreto2006unifying}
J.~P. Barreto, ``A unifying geometric representation for central projection
  systems,'' \emph{CVIU}, 2006.

\bibitem{duane1971close}
D.~C. Brown, ``Close-range camera calibration,'' \emph{Photogrammetric
  Engineering}, 1971.

\bibitem{sturm2011camera}
P.~Sturm, S.~Ramalingam, J.-P. Tardif, S.~Gasparini, and J.~P. Barreto,
  ``Camera models and fundamental concepts used in geometric computer vision,''
  \emph{Foundations and Trends in Computer Graphics and Vision}, 2011.

\bibitem{fitzgibbon2001simultaneous}
A.~Fitzgibbon, ``Simultaneous linear estimation of multiple view geometry and
  lens distortion,'' in \emph{CVPR}, 2001.

\bibitem{pizarro2003toward}
O.~Pizarro and H.~Singh, ``Toward large-area mosaicing for underwater
  scientific applications,'' \emph{IEEE journal of oceanic engineering},
  vol.~28, no.~4, pp. 651--672, 2003.

\bibitem{lin2020infrastructure}
Y.~Lin, V.~Larsson, M.~Geppert, Z.~Kukelova, M.~Pollefeys, and T.~Sattler,
  ``Infrastructure-based multi-camera calibration using radial projections,''
  in \emph{ECCV}.\hskip 1em plus 0.5em minus 0.4em\relax Springer, 2020, pp.
  327--344.

\bibitem{opencv_library}
G.~Bradski, ``{The OpenCV Library},'' \emph{Dr. Dobb's Journal of Software
  Tools}, 2000.

\bibitem{imagemagick}
\BIBentryALTinterwordspacing
{The ImageMagick Development Team}, ``Imagemagick.'' [Online]. Available:
  \url{https://epaperpress.com/ptlens/}
\BIBentrySTDinterwordspacing

\bibitem{ptlens}
\BIBentryALTinterwordspacing
{Tom Niemann and the PTLens Development Team}, ``Ptlens.'' [Online]. Available:
  \url{https://imagemagick.org}
\BIBentrySTDinterwordspacing

\bibitem{hugin}
\BIBentryALTinterwordspacing
{Pablo d'Angelo and the Hugin developers}, ``Hugin.'' [Online]. Available:
  \url{https://http://hugin.sourceforge.net/}
\BIBentrySTDinterwordspacing

\bibitem{FisheyeHemi}
Fisheye-Hemi, ``\url{https://imadio.com/products/prodpage_hemi.aspx},'' 2015.

\bibitem{CarrollAA09}
R.~Carroll, M.~Agrawala, and A.~Agarwala, ``Optimizing content-preserving
  projections for wide-angle images,'' \emph{TOG}, 2009.

\bibitem{pepperell2019fovo}
R.~Pepperell, A.~Burleigh, N.~Ruta, and T.~Langford, ``Fovo: A flexible
  real-time computer graphics rendering process,'' \emph{Proceedings of EVA
  London 2019}, pp. 237--238, 2019.

\bibitem{fairchild2013color}
M.~D. Fairchild, \emph{Color appearance models}.\hskip 1em plus 0.5em minus
  0.4em\relax John Wiley \& Sons, 2013.

\bibitem{Vangorp2013}
P.~Vangorp, C.~Richardt, E.~A. Cooper, G.~Chaurasia, M.~S. Banks, and
  G.~Drettakis, ``Perception of perspective distortions in image-based
  rendering,'' \emph{TOG}, vol.~32, no.~4, pp. 58:1--58:12, 2013.

\bibitem{ledig-cvpr-17}
C.~Ledig, L.~Theis, F.~Huszar, J.~Caballero, A.~Cunningham, A.~Acosta, A.~P.
  Aitken, A.~Tejani, J.~Totz, Z.~Wang, and W.~Shi, ``Photo-realistic single
  image super-resolution using a generative adversarial network,'' in
  \emph{CVPR}, 2017, pp. 105--114.

\bibitem{vinyals-cvpr-15}
O.~Vinyals, A.~Toshev, S.~Bengio, and D.~Erhan, ``Show and tell: {A} neural
  image caption generator,'' in \emph{CVPR}, 2015, pp. 3156--3164.

\bibitem{papazoglou-accv-16}
A.~Papazoglou, L.~D. Pero, and V.~Ferrari, ``Video temporal alignment for
  object viewpoint,'' in \emph{ACCV}, 2016, pp. 273--288.

\bibitem{YingH:ECCV:04}
X.~Ying and Z.~Hu, ``Can we consider central catadioptric cameras and fisheye
  cameras within a unified imaging model,'' in \emph{ECCV}, 2004.

\bibitem{Wilson2014}
K.~Wilson and N.~Snavely, ``Robust global translations with 1dsfm,'' in
  \emph{ECCV}, 2014.

\bibitem{Hold-Geoffroy2017}
Y.~Hold-Geoffroy, K.~Sunkavalli, S.~Hadap, E.~Gambaretto, and J.-F. Lalonde,
  ``Deep outdoor illumination estimation,'' in \emph{CVPR}, 2017.

\bibitem{Huang2016}
G.~Huang, Z.~Liu, K.~Q. Weinberger, and L.~van~der Maaten, ``Densely connected
  convolutional networks,'' in \emph{IEEE Conference on Computer Vision and
  Pattern Recognition}, 2016.

\bibitem{Russakovsky2015}
O.~Russakovsky, J.~Deng, H.~Su, J.~Krause, S.~Satheesh, S.~Ma, Z.~Huang,
  A.~Karpathy, A.~Khosla, M.~S. Bernstein, A.~C. Berg, and F.~Li, ``Imagenet
  large scale visual recognition challenge,'' \emph{IJCV}, vol. 115, no.~3, pp.
  211--252, 2015.

\bibitem{Kingma2015}
D.~Kingma and J.~Ba, ``{ADAM: A Method for Stochastic Optimization},'' in
  \emph{International Conference for Learning Representations}, 2015.

\bibitem{kannala2006generic}
J.~Kannala and S.~S. Brandt, ``A generic camera model and calibration method
  for conventional , wide-angle, and fish-eye lenses,'' \emph{TPAMI}, 2006.

\bibitem{Springenberg2015}
J.~T. Springenberg, A.~Dosovitskiy, T.~Brox, and M.~Riedmiller, ``Striving for
  simplicity: The all convolutional net,'' \emph{International Conference on
  Learning Representations}, 2015.

\bibitem{Smilkov2017}
D.~Smilkov, N.~Thorat, B.~Kim, F.~Vi\'{e}gas, and M.~Wattenberg, ``Smoothgrad:
  removing noise by adding noise,'' \emph{arXiv preprint arXiv:1706.03825},
  2017.

\bibitem{schoenberger2016sfm}
J.~L. Sch{\"{o}}nberger and J.~Frahm, ``Structure-from-motion revisited,'' in
  \emph{CVPR}, 2016.

\bibitem{schoenberger2016mvs}
J.~L. Sch\"{o}nberger, E.~Zheng, M.~Pollefeys, and J.-M. Frahm, ``Pixelwise
  view selection for unstructured multi-view stereo,'' in \emph{ECCV}, 2016.

\bibitem{Zhou2017}
B.~Zhou, {\`{A}}.~Lapedriza, A.~Khosla, A.~Oliva, and A.~Torralba, ``Places:
  {A} 10 million image database for scene recognition,'' \emph{TPAMI}, vol.~40,
  no.~6, pp. 1452--1464, 2018.

\bibitem{Gardner2017}
M.-A. Gardner, K.~Sunkavalli, E.~Yumer, X.~Shen, E.~Gambaretto, C.~Gagn\'{e},
  and J.-F. Lalonde, ``Learning to predict indoor illumination from a single
  image,'' \emph{ACM Transactions on Graphics (SIGGRAPH Asia)}, vol.~9, no.~4,
  2017.

\bibitem{blender}
\BIBentryALTinterwordspacing
{Blender Online Community}, \emph{Blender - a 3D modelling and rendering
  package}, Blender Foundation, Blender Institute, Amsterdam, 2021. [Online].
  Available: \url{http://www.blender.org}
\BIBentrySTDinterwordspacing

\bibitem{Gingold2012}
Y.~Gingold, A.~Shamir, and D.-O. Cohen, ``Micro perceptual human computation
  for visual tasks,'' \emph{ACM Transactions on Graphics}, vol.~31, no.~5,
  2012.

\bibitem{hartley2007parameter}
R.~Hartley and S.~B. Kang, ``Parameter-free radial distortion correction with
  center of distortion estimation,'' \emph{TPAMI}, 2007.

\bibitem{li2005non}
H.~Li and R.~Hartley, ``A non-iterative method for correcting lens distortion
  from nine point correspondences,'' \emph{OMNIVIS}, 2005.

\bibitem{cornelis2002lens}
K.~Cornelis, M.~Pollefeys, and L.~V. Gool, ``Lens distortion recovery for
  accurate sequential structure and motion recovery,'' in \emph{ECCV}, 2002.

\bibitem{zhang2018unreasonable}
R.~Zhang, P.~Isola, A.~A. Efros, E.~Shechtman, and O.~Wang, ``The unreasonable
  effectiveness of deep features as a perceptual metric,'' in \emph{Proceedings
  of the IEEE conference on computer vision and pattern recognition}, 2018, pp.
  586--595.

\bibitem{johnson2016perceptual}
J.~Johnson, A.~Alahi, and L.~Fei-Fei, ``Perceptual losses for real-time style
  transfer and super-resolution,'' in \emph{European conference on computer
  vision}.\hskip 1em plus 0.5em minus 0.4em\relax Springer, 2016, pp. 694--711.

\end{thebibliography}
